\documentclass{article}

 \usepackage[preprint]{neurips_2026}

\usepackage[utf8]{inputenc} 
\usepackage[T1]{fontenc}    
\usepackage{hyperref}       
\usepackage{url}            
\usepackage{booktabs}       
\usepackage{amsfonts}       
\usepackage{nicefrac}       
\usepackage{microtype}      
\usepackage{xcolor}  
\usepackage{microtype}
\usepackage{graphicx}
\usepackage{subcaption}
\usepackage{caption}
\usepackage{booktabs} 
\usepackage{algorithm}
\usepackage{wrapfig}
\usepackage{algpseudocode}
\usepackage{booktabs}
\usepackage{multirow}
\usepackage{amssymb}
\usepackage{pifont}
\usepackage{amsmath}
\usepackage{amssymb}
\usepackage{mathtools}
\usepackage{amsthm}

\usepackage{booktabs}
\usepackage{makecell}
\usepackage{pifont}

\usepackage{booktabs}       
\usepackage{amsfonts}       
\usepackage{nicefrac}       
\usepackage{microtype}      
\usepackage{xcolor}         
\usepackage{amsmath}
\usepackage{subcaption}
\usepackage{graphicx}
\usepackage{amsmath, amssymb, amsfonts, amsthm}
\usepackage{algpseudocode}
\usepackage{setspace}
\usepackage{multirow}
\usepackage{algorithm}

\usepackage{booktabs}
\usepackage{multirow}
\usepackage{amssymb}
\usepackage{pifont}

\newtheorem{theorem}{Theorem}[section]
\newtheorem{lemma}[theorem]{Lemma}

\definecolor{ForestGreen}{cmyk}{0.864, 0.0, 0.429, 0.396}
\definecolor{Green}{cmyk}{1.0, 0.0, 1.0, 0.5}
\definecolor{Brown}{rgb}{0.59,0.29,0.0}
\definecolor{Blue}{rgb}{0.0,0.0,1.0}
\definecolor{Orange}{rgb}{1.0,0.5,0.0}
\definecolor{Red}{rgb}{1.0,0.0,0.0}

\newif\ifcomments
\commentstrue    

\newcommand\shortsection[1]{\vspace{2pt}{\noindent\textbf{#1.}}}
\newcommand\shortersection[1]{\vspace{2pt}{\noindent\em #1.}}

\usepackage{amsmath,amsfonts,bm}

\def\eqref#1{equation~\ref{#1}}

\def\1{\bm{1}}

\DeclareMathAlphabet{\mathsfit}{\encodingdefault}{\sfdefault}{m}{sl}
\SetMathAlphabet{\mathsfit}{bold}{\encodingdefault}{\sfdefault}{bx}{n}

\DeclareMathOperator*{\argmin}{arg\,min}

\usepackage[textsize=tiny]{todonotes}

\usepackage{hyperref}

\title{ContextFlow: Context-Aware Flow Matching for Trajectory Inference from Spatial Omics Data}

\author{
  \textbf{Santanu Subhash Rathod}\textsuperscript{1} \quad
  \textbf{Francesco Ceccarelli}\textsuperscript{2} \quad
  \textbf{Sean B. Holden}\textsuperscript{2} \quad
  \textbf{Pietro Li\`{o}}\textsuperscript{2} \\
  \textbf{Xiao Zhang}\textsuperscript{1} \quad
  \textbf{Jovan Tanevski}\textsuperscript{3} \\[0.6em]
  \textsuperscript{1}CISPA Helmholtz Center for Information Security, Saarbrücken, Germany \\
  \textsuperscript{2}Department of Computer Science and Technology, University of Cambridge, Cambridge, UK \\
  \textsuperscript{3}Institute for Computational Biomedicine, Heidelberg University Hospital, Germany \\[0.3em]
  \texttt{\{santanu.rathod,xiao.zhang\}@cispa.de},\ \texttt{\{fc485,sbh11,pl219\}@cam.ac.uk} \\
  \texttt{jovan.tanevski@uni-heidelberg.de}
}

\begin{document}

\maketitle

\begin{abstract}
Inferring trajectories from longitudinal, spatially resolved omics data is fundamental to understanding the dynamics of structural and functional tissue changes in development, regeneration and repair, disease progression, and response to treatment. We propose ContextFlow, a context-aware flow matching framework that leverages prior knowledge to guide the inference of structural tissue dynamics from spatially resolved omics data. Existing approaches rely on optimal-transport-derived couplings that are determined solely by transcriptomic similarity, which can lead to biologically implausible transitions and, in turn, inconsistent learned dynamics. Specifically, ContextFlow integrates local tissue organization and ligand-receptor communication patterns into a transition plausibility matrix that regularizes the optimal transport objective. By embedding these contextual constraints, ContextFlow generates trajectories that are not only statistically consistent but also biologically meaningful, making it a generalizable framework for modeling cellular dynamics from longitudinal, spatially resolved omics data. Comprehensive experiments show that ContextFlow consistently improves state-of-the-art flow matching methods across multiple metrics of inference accuracy and biological coherence. Our code is available at: \url{https://github.com/santanurathod/ContextFlow}.
\end{abstract}

\section{Introduction}
Inferring the underlying dynamics from sparse and noisy observations is a central challenge in many applications~\citep{gontis2010long,brunton2016discovering,pandarinath2018inferring,9_UIUC_FM_survey}, 
where continuous trajectories are rarely captured; instead, cross-sectional snapshots, collected at discrete time points, are available. In single-cell RNA sequencing, this challenge becomes especially critical, as the destructive nature of profiling technologies yields only unpaired population-level snapshots over time.  
Uncovering temporal dynamics from such snapshot data is essential for understanding developmental processes, disease progression, treatment, and perturbation responses~\citep{wagner2020lineage}.
Traditional methods~\citep{trapnell2017monocle,3_neural_ODE,rubanova2019latent,bergen2020generalizing} rely on heuristics or computationally intensive likelihood-based generative models, which struggle with flexibility and scalability in handling high-dimensional data. In contrast, flow matching~\citep{lipman2023flow}, an emerging generative paradigm, overcomes these issues by directly learning continuous latent dynamics constrained to match population-level distributions at observational time points, while enabling parametric flexibility to model data distributions.

Cellular dynamics are affected by the tissue microenvironment, in which cells engage in reciprocal communication with their neighbors~\citep{19_Dimitrov_cc_interaction_2,17_Tanevski_kasumi}.
State-of-the-art flow matching frameworks~\citep{tong2024improving,kapusniak2024metric,atanackovic2024meta,11_rohbeck} use optimal transport (OT) to derive couplings between consecutive observational time points and design conditional paths for regression over the velocity fields.
Although OT couplings improve training stability and inference efficiency over vanilla independent couplings~\citep{lipman2023flow}, they are determined solely by transcriptomic similarity and do not account for the contextual richness of spatial transcriptomics~\citep{armingol2021deciphering, 12_Shimrit_cell_cell_communication, 2_Sole_cancer_ecosystems,zhang2025modeling}, thus yielding couplings that are statistically optimal but biologically implausible. Since OT-based flow matching algorithms directly regress on these couplings, these errors propagate into the learned velocity field, leading to globally inconsistent inferred dynamics.

To address these limitations, we introduce ContextFlow, a generative framework that incorporates explicit spatial priors into OT-coupled flow matching to model temporal tissue dynamics (see Figure \ref{fig:graphical} for an illustration). In contrast to existing approaches, ContextFlow integrates prior knowledge directly into the construction of OT couplings to avoid biologically implausible transitions as the regression target. By encoding local tissue organization and ligand–receptor-derived spatial communication patterns into prior-regularized optimal transport formulations, ContextFlow exploits the contextual richness of spatial omics data and embeds both structural and functional aspects of tissue organization into the training objective, thereby generating more biologically informed trajectories.

In summary, our contributions are as follows:
\begin{itemize}
\setlength{\itemsep}{0pt}
\setlength{\topsep}{0.1pt}
\vspace{-0.05in}
\item We define transitional plausibility to capture local tissue organization and ligand–receptor communication patterns, which provides a flexible and interpretable mechanism for encoding spatial and functional priors into couplings used to train the velocity field (Section \ref{sec: spatial prior}). Our definition allows priors to be explicitly and flexibly adapted in a domain-specific manner, enabling the incorporation of external biological knowledge and improving interpretability.
\item We introduce principled methods for integrating explicit biological priors into the state-of-the-art OT-based flow matching framework, instantiated via two novel schemes—cost-based and entropy-based—both amenable to efficient Sinkhorn optimization (Section \ref{sec: contextflow}). 
\item Comprehensive experiments on regeneration and developmental datasets show that ContextFlow consistently improves existing methods for both interpolation and extrapolation tasks across metrics that capture biological plausibility and statistical fidelity (Section \ref{sec: experiments}).
\end{itemize}
\section{Related Work}
\label{apex: related work}
\subsection{Flow Matching and its Applications in Biology}
Flow matching (FM)~\citep{lipman2023flow, 5_Vanden_FM, 6_Xingchao_FM} is a generative modeling framework that trains the velocity field via a tractable regression objective for learning continuous normalizing flows~\citep{1_JMLR_normalizing_flows}, which is more efficient and scalable when dealing with high-dimensional data than prior approaches, such as neural ODE~\citep{3_neural_ODE,rubanova2019latent}. Compared with score-based diffusion models~\citep{5_song_scorebaseddiffusion} and Schrödinger Bridges (SB)~\citep{34_SB_Bunne}, FM is faster and more stable to train and sample from during inference, as it characterizes the latent continuous dynamics via an ODE. Therefore, a growing body of literature has adapted and improved the vanilla FM framework to model dynamic scientific processes, among others, especially in biology and the life sciences~\citep{9_UIUC_FM_survey}. 

Instead of independent couplings, \citep{pooladian2023multisample,tong2024improving} used optimal transport (OT) maps to define the conditional coupling between the source and the target, and approximated it using minibatch samples, leading to more stable training and faster inference via straightening the flows. 
Based on this framework, ~\citep{kapusniak2024metric} introduced metric flow matching to learn the flows over a non-Euclidean manifold to make a case against straighter flows in the context of transcriptomics. \citep{atanackovic2024meta} proposed meta flow matching, which incorporates implicit cell-cell interaction prior via GNN conditioning. Apart from these trajectory inference-centric studies, FM has also been utilized to generate imaging-based cell morphology changes~\citep{12_cellflux_Zhang}, simulate spatial transcriptomics data from histology images~\citep{13_Spatial_Huang}, and approximate OT maps for drug response modeling and cross-modal translation tasks~\citet{10_Klien_GENOT}. Despite these advances, existing work does not address how to meaningfully incorporate \textit{explicit biological prior knowledge} to constrain the velocity field, thereby limiting the biological plausibility of inferred trajectories, interpretability, and control over the learned dynamics. 


\subsection{Advancement in Spatial Transcriptomics}
The state and function of cells within a tissue are influenced by interactions with neighboring cells, extracellular matrix components, and local signaling gradients~\citep{rao2021exploring}. Recent advances in spatial omics technologies, particularly spatial transcriptomics (ST), allow gene expression profiling without tissue dissociation, thereby preserving spatial context and providing a complementary view of cellular organization.
The dynamics of cellular processes are regulated by the tissue microenvironment and by cell communications with their neighbours~\citep{19_Dimitrov_cc_interaction_2,17_Tanevski_kasumi}. A growing body of literature highlights the critical role of spatial cell–cell communication patterns in shaping cellular phenotypes~\citep{armingol2021deciphering}. In particular, location-specific communication circuits between distinct cell types dynamically interact to reprogram cellular states and influence tissue-level behavior~\citep{12_Shimrit_cell_cell_communication, 2_Sole_cancer_ecosystems, 3_Zheng_genai_cancer}. These insights, made possible by the spatiotemporal resolution of transcriptomics data, pave the way for understanding the mechanisms by which cellular interactions drive tissue organization and function in organogenesis~\citep{2_dataset_mosta}, regeneration~\citep{3_dataset_liver, 1_dataset_steroseq}, disease progression~\citep{4_dataset_mouse_MS}, and treatment response~\citep{1_Liu_Cellreview_Spatial}.

Optimal transport~\citep{5_Bunne_OT_methods, 8_Klien_moscot} has been increasingly used to analyze omics data, as they often exhibit uncoupled measurements across modalities, conditions, or time points.
In ST, several OT formulations have been introduced depending on context. For instance, \citet{zeira2022alignment} and \citet{liu2023partial} proposed PASTE and PASTE2 to align ST data from adjacent tissue slices, while DeST-OT~\citep{halmos2025dest} integrates spatio-temporal slices by modeling cell growth and differentiation. In addition, \citet{rahimi2024dot} developed DOT, a multi-objective OT framework for mapping features across scRNA-seq and spatially resolved assays, and \citet{ceccarelli2026topography} introduced TOAST, a spatially regularized OT framework for slice alignment and annotation transfer. 
While these OT techniques can successfully leverage spatial or structural information to improve alignment quality across space, time, and modalities, they do not learn a velocity field governing temporal evolution, as in generative modeling frameworks like flow matching. Consequently, while these techniques can model temporal relationships and align distributions across time, they do not learn a continuous-time generative model and therefore cannot directly simulate intermediate or future cellular states.
Moreover, although some of the aforementioned approaches incorporate spatial or structural constraints, these are typically introduced in a task-specific or implicit manner, rather than as general mechanisms to encode explicit biological priors directly into the transport plan.

\section{Problem Formulation}
 \label{sec: problem}

We aim to infer the temporal dynamics of cell states from spatially resolved gene expression data. 
Let $0 = t_1 < t_2 < \ldots < t_{m+1} = 1$ be a sequence of normalized time points and $q_i$ be the underlying distribution over $\mathbb{R}^d$ at time $t_i$. 
Given $\{ \mathbf{X}_{t_i} \}_{i\in\{1, 2,\ldots, m+1\}}$, where $\mathbf{X}_{t_i} = \{\bm{x}_{i}(k)\}_{k=1}^{n_i}$
corresponds to the gene expressions at time $t_i$ consisting of $n_i$ snapshot data sampled from $q_i$, our goal is to learn a neural velocity vector field $u_{\theta}:[0,1] \times \mathbb{R}^d \rightarrow \mathbb{R}^d$ to faithfully characterize the temporal evolution of spatially resolved tissues over time, such that the induced probability path $p_{t}$ can describe the state of each cell at any time $t\in[0,1]$. 
A standard approach is to apply \textit{conditional flow matching with minibatch OT couplings} (MOTFM)~\citep{tong2024improving} for each pair of consecutive time points. 
Specifically, for any $t\in [0,1]$ satisfying $t\in[t_i, t_{i+1})$, the following linear conditional velocity field $u_t(\bm{x} | \bm{z})$ and its associated probability path $p_t(\bm{x} | \bm{z})$ are defined as the regression target in MOTFM:
\begin{align}
\label{eq: multi-time sampling path}
    u_t(\bm{x} | \bm{z}_i) = \frac{\bm{x}_{i+1} - \bm{x}_i}{t_{i+1} - t_i} \quad \text{and} \quad p_t(\bm{x} | \bm{z}_i) = \mathcal{N}\left( \frac{t_{i+1} - t}{t_{i+1} - t_i} \bm{x}_i + \frac{t - t_i}{t_{i+1} - t_i}  \bm{x}_{i+1},  \sigma^2 \mathbf{I}  \right),
\end{align}
where $\bm{z}_i = (\bm{x}_{i}, \bm{x}_{i+1})$. Let $q(\bm{z}_i)$ be the joint probability measure with marginals $q_i$ and $q_{i+1}$, corresponding to the \textit{entropic optimal transport} (EOT) coupling, defined as:
\begin{equation}
\label{eq: EOT formulation}
\begin{aligned}
    q(\bm{z}_i) = \argmin_{\pi \in \Pi (q_i, q_{i+1})} \int_{\mathbb{R}^d \times \mathbb{R}^d} \|\bm{x}_0 - \bm{x}_1\|_2^2 \: d\pi(\bm{x}_0, \bm{x}_1) + \epsilon H(\pi \: | \: q_i \otimes q_{i+1}),
\end{aligned}
\end{equation}
where $\epsilon > 0$ is the regularization parameter, and $H(\pi \: | \: q_i \otimes q_{i+1})$ is the KL divergence with respect to the transport plan $\pi$ and the product measure $q_i \otimes q_{i+1}$.
To train the velocity vector field $u_\theta$, MOTFM randomly samples a mini-batch of data at each time, uses the Sinkhorn algorithm~\citep{1_Cuturi_Sinkhorn_fast} to obtain the EOT couplings, and optimizes the following regression loss using SGD:
\begin{align}
\label{eq: conditional flow matching loss}
    \min_{\theta} \: \mathbb{E}_{t \sim \mathcal{U}(0,1), \bm{z} \sim q(\bm{z}_{i}), \bm{x} \sim p_t(\bm{x}|\bm{z}_i)} \big\| u_\theta(t, \bm{x}) - u_t(\bm{x} | \bm{z}_i) \big\|^2,
\end{align}
where $\mathcal{U}(0,1)$ is the uniform distribution over $[0,1]$, and $i = i(t)$ denotes the integer that satisfies $t_{i} \leq t < t_{i+1}$ for each sampled $t$.
Full preliminaries of flow matching are provided in Appendix~\ref{sec:preliminaries}.



\begin{figure}[t]
    \centering
    \includegraphics[width=0.98\linewidth]{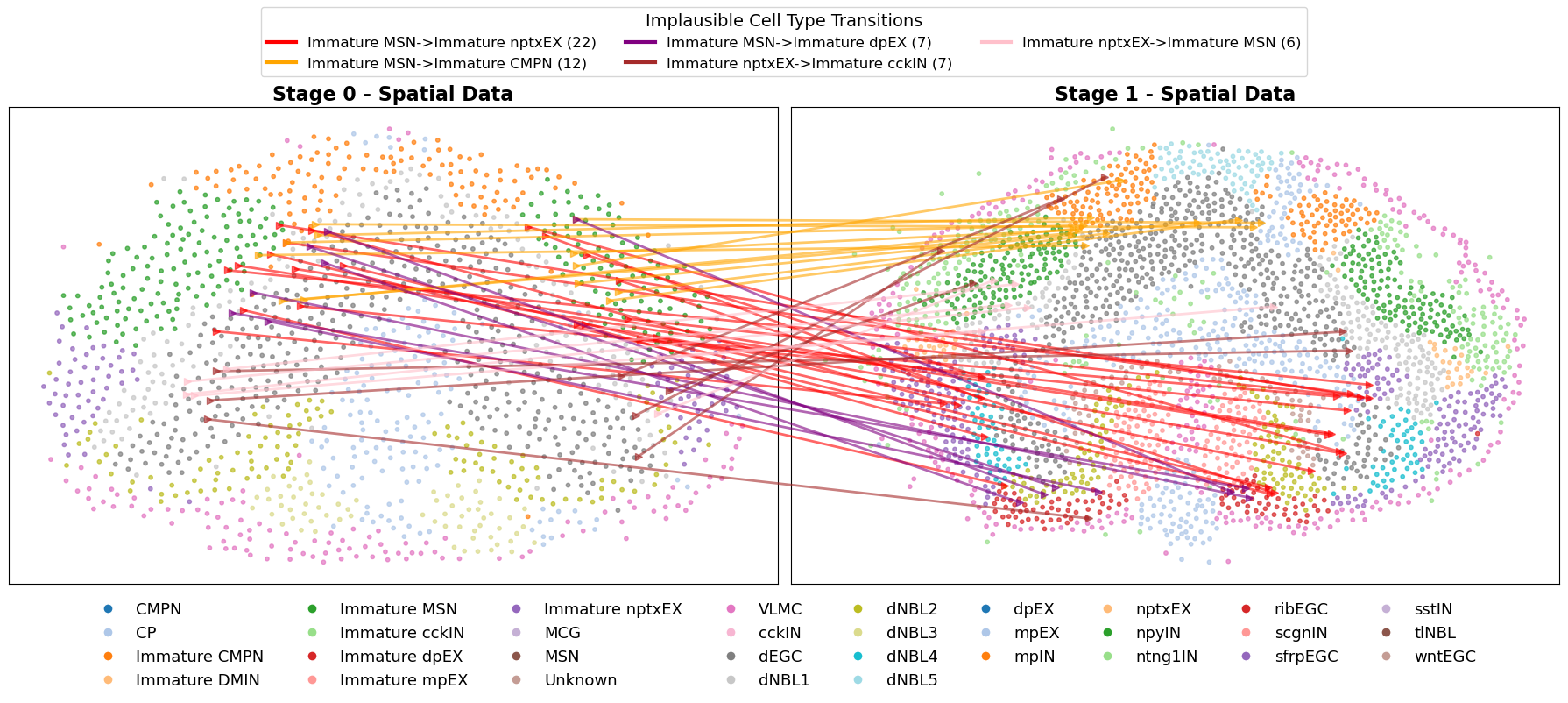}
    \vspace{-0.05in}
    \caption{Visualization of inhibitory-exitatory coupled cells sampled from the EOT plan used in MOTFM on temporally consecutive spatial omics slides from the Brain Regeneration dataset~\citep{1_dataset_steroseq}. Here, different colors correspond to different cell types. Since inhibitory/excitatory cells serve distinct biological functions, their coupling and thus the implied state transition are infeasible.
}
    \vspace{-0.1in}
    \label{fig:EOT_implausible_couplings}
\end{figure}

Despite their enhanced ability to model system dynamics, state-of-the-art OT-based flow matching frameworks lack provisions to fully exploit the contextual richness and integrate biological prior knowledge inferred from other associated data modalities. Existing approaches can generate statistically optimal trajectories by targeting probability paths induced by OT couplings along the temporal dimension. However, because they overlook important functional or structural prior information, their velocity fields can be trained on couplings that violate domain principles, leading to biologically implausible trajectories. Figure~\ref{fig:EOT_implausible_couplings} presents an example of \textit{statistically optimal yet biologically implausible} couplings between inhibitory-excitatory neuron cells coming from the EOT plan used to train MOTFM~\citep{tong2024improving} (see Appendix~\ref{apdx:vizualization_implausibility} for detailed discussions). This motivates our work to develop a mechanism to integrate domain prior knowledge into the OT-based flow matching framework, thereby encouraging domain-aligned training of the velocity field. 

\section{Regularizing the Flow with Spatial Priors}
\label{sec: regularizing the flow}

\subsection{Spatial Priors and Transitional Plausibility}
\label{sec: spatial prior}

To address the limitations, we propose modeling the context and cellular organization of spatial omics data using two types of spatial priors (spatial smoothness and cell-cell communication patterns), strong indicators of transitional plausibility between locations and cell states across time points.

\shortsection{Spatial Smoothness}
Tissues are well-organized systems. Within a microenvironment, neighboring cells respond to the same set of external mechanical stimuli and intercellular communication, which affect their states similarly and result in local smoothness in cell-type-specific expression. Due to tissue heterogeneity, we cannot assume a common reference coordinate frame across tissue samples or even slices at $t_i$ and $t_j$ at a larger scale.  However, the same heterogeneity allows us to consider spatial coherence and neighborhood consistency \citep{Greenwald2024Integrative, ceccarelli2026topography} as proxies for relative cell localization, which cannot change significantly over short time intervals. Therefore, the aggregate expression within each cell's microenvironment can be used to quantify the plausibility of transition across consecutive time points.

Specifically, let $c_i = (\bm{x}_i, \bm{s}_i)$ and $c_j = (\bm{x}_j, \bm{s}_j)$ be cells at time points $t_i$ and $t_j$, respectively, where $\bm{x}_i, \bm{x}_j \in \mathbb{R}^d$ denote their gene expression profiles, and $\bm{s}_i, \bm{s}_j \in \mathbb{R}^2$ denote their spatial coordinates in the relative tissue reference frame. Let $\mathrm{TP}(c_i, c_j)$ denote the \textit{transitional plausibility}, i.e., the likelihood that $c_i$ evolves to $c_j$ between $t_i$ and $t_j$. Spatial smoothness suggests that $\mathrm{TP}(c_i, c_j)$ is inversely related to the difference between the average expression profiles of their local neighborhoods:
\begin{equation}\label{eq: spatial smoothness prior}
\begin{aligned}
\mathrm{SS}(c_i, c_j) &= \bigg\| \frac{1}{|\mathcal{N}_r(c_i)|}\sum_{c\in\mathcal{N}_r(c_i)}\bm{x}(c) -\frac{1}{|\mathcal{N}_r(c_{j})|}\sum_{c\in \mathcal{N}_r(c_{j})} \bm{x}(c)\bigg\|_2^2, 
\end{aligned}
\end{equation}
where $\mathcal{N}_r(c_i) = \{c: \| \bm{s}(c) - \bm{s}(c_i) \|_2 \leq r \}$ denotes the set of neighboring cells of $c_i$ in the same tissue slice, $|\mathcal{N}_r(c_i)|$ is the cardinality of $\mathcal{N}_r(c_i)$, and $\bm{x}(c)$ is the gene expression profile of cell $c$. 
 
\shortsection{Cell-Cell Communication Patterns}
Cell–cell communication plays a critical role in regulating biological processes, including development, apoptosis, and homeostasis in health and disease~\citep{armingol2024diversification}. A major type of cell-cell communication is ligand–receptor (LR) signaling, in which ligands expressed by one cell bind to cognate receptors on another, initiating intracellular cascades that ultimately affect the expression profiles of the cell~\citep{armingol2021deciphering}. Numerous databases of prior knowledge of LR binding exist, and computational methods use them to systematically link gene expression with the activity of ligand-receptor-mediated communication.

Specifically, we represent each cell $c_i$ by a feature vector $f_{\mathrm{LR}} \in \mathbb{R}^{p}$, where each entry corresponds to one of $p$ possible ligand–receptor pairs and encodes the extent of $c_i$’s participation in communication through that pair. The transitional plausibility $\mathrm{TP}(c_i, c_j)$ between cells in different tissue slices is higher when they exhibit similar ligand-receptor communication patterns $f_{\mathrm{LR}}$ (see Figure~\ref{fig:LR_viz} in Appendix \ref{apdx:LR_patterns} for an illustration). Formally, we define $\mathrm{LR}(c_i, c_j)$, the dissimilarity between the ligand–receptor communication patterns in the microenvironments of cells $c_i$ and $c_j$, as:
\begin{align}
\label{eq: LR prior}
    \mathrm{LR}(c_i, c_j) = \left\| \: f_{\mathrm{LR}}(\mathcal{N}_r(c_i)) - f_{\mathrm{LR}}(\mathcal{N}_r(c_j)) \right\|_2^2.
\end{align}

\begin{figure}[t]
    \centering
    \includegraphics[width=0.98\linewidth]{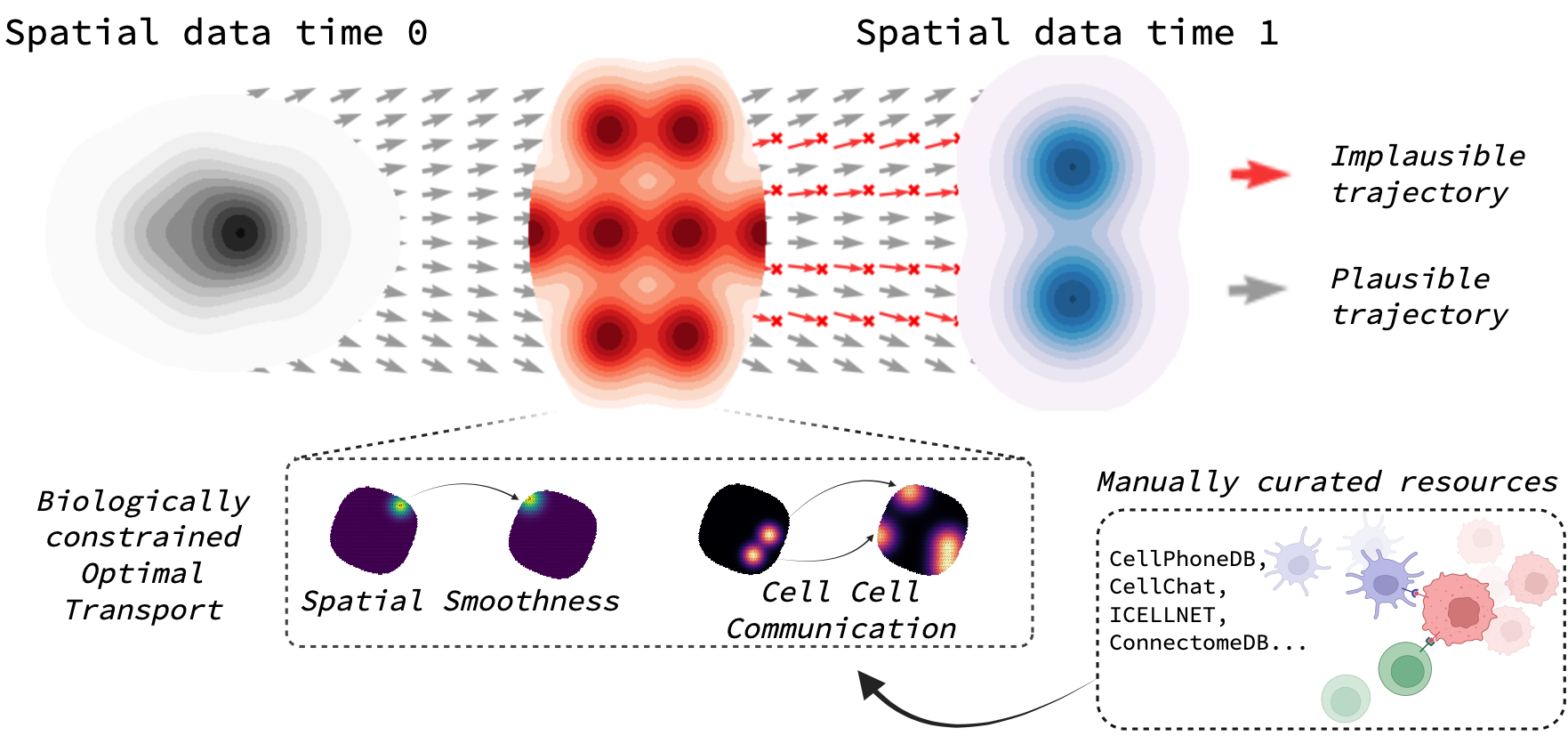}
    \vspace{-0.05in}
    \caption{ContextFlow integrates local tissue organization and ligand-receptor communications to learn biologically meaningful trajectories from spatial omics data. Prior knowledge acts as a soft filter, discouraging implausible transitions while preserving flexibility in trajectory inference.}
    \vspace{-0.1in}
    \label{fig:graphical}
\end{figure}

\subsection{Flow Matching with Context-Aware OT Couplings}
\label{sec: contextflow}

Our proposed framework, graphically depicted in Figure~\ref{fig:graphical}, consists of the following three main steps:

\shortsection{Transitional Plausibility Matrix}
To begin with, we create a sequence of \emph{transitional plausibility matrices} (TPMs) to encode the biological priors for each pair of consecutive time points.
Specifically, denote by $\mathbf{M}_i \in \mathbb{R}^{n_i \times n_{i+1}}$ the TPM with respect to the set of cells measured at time $t_i$ and at time $t_{i+1}$, with size $n_i$ and $n_{i+1}$ respectively, where the $(k,l)$-th entry of $\mathbf{M}_i$ indicates how plausibly the $k$-th cell measured at $t_i$ will evolve to the $l$-th cell measured at $t_{i+1}$, defined as follows:
\begin{equation}\label{eq: TPM}
\begin{aligned}
\big[ \mathbf{M}_i \big]_{kl} = \lambda \cdot \mathrm{SS}&\left( c_i (k), c_{i+1}(l) \right) + 
(1-\lambda) \cdot \mathrm{LR}\left( c_i(k), c_{i+1}(l) \right).
\end{aligned}
\end{equation}
Here, $\lambda \in [0,1]$ denotes a trade-off hyperparameter that balances the contribution of the spatial smoothness prior ($\mathrm{SS}$) and the ligand–receptor communication prior ($\mathrm{LR}$).


\shortsection{Prior-Regularized OT Couplings}
The transitional plausibility matrix encodes our spatially informed prior over cell-cell transitions between consecutive time points, which can naturally be incorporated into the EOT formulation (Equation \ref{eq: EOT formulation}) to promote couplings that maintain the structural and functional properties of tissue organization. In particular, we propose two techniques for prior integration:

\shortersection{Prior-Aware Cost Matrix (PACM)}
\label{sec: pacm}
Consider $(t_i, t_{i+1})$ and the empirical counterpart of Equation \ref{eq: EOT formulation}.
Our first approach incorporates the transitional plausibility matrix directly into the transport cost:
\begin{align}
\label{eq: PACM}
    \min_{\mathbf{\Pi} \in \mathbb{R}^{n_i \times n_{i+1}}}
    \;
    \langle \mathbf{\Pi}, \mathbf{C}_{i,i+1} \rangle
    - \epsilon \sum_{k,l} \Pi_{kl}(\log \Pi_{kl}-1),
\end{align}
where the $(k,l)$-th entry $[\mathbf{C}_{i,i+1}]_{kl}
= \alpha \|\bm{x}_i(k)-\bm{x}_{i+1}(l)\|_2^2 + (1-\alpha)[\mathbf{M}_{i,i+1}]_{kl}$, the transport plan $\mathbf{\Pi}$ needs to satisfy the boundary conditions: $\sum_{l} \Pi_{kl} = 1 / n_i$ for any $k\in [n_i]$, and $\sum_{k} \Pi_{kl} = 1 / n_{i+1}$ for any $l\in [n_{i+1}]$, and $\alpha\in[0,1]$ controls the trade-off between the original Euclidean cost and the prior-aware cost derived from the transitional plausibility.
If $[\mathbf{M}_i]_{kl}$ is high, Equation \ref{eq: PACM} will impose a higher transport cost between the $k$-cell at time $i$ to the $j$-cell at time $i+1$. This observation aligns with our assumption that such transitions are less biologically plausible.

\shortersection{Prior-Aware Entropy Regularization (PAER)}
\label{sec: paer}
While PACM penalizes couplings according to our spatial priors, it defines a different OT problem with a modified cost function. Thus, the standard interpretation of OT as minimizing the transport energy between two transcriptomic distributions no longer holds. Since the scales of the pairwise distances often differ, normalization of the cost terms is required to enable meaningful comparison. This normalization, however, may result in couplings that deviate from their original counterparts (see Theorem \ref{thm:relative_change_norm_cost} in Appendix~\ref{apdx:pacm} for a formal statement). Besides, selecting an appropriate $\alpha$ in Equation~\ref{eq: PACM} introduces an additional layer of tuning, increasing computational overhead. To address the above issues, we propose the following PAER approach:
\begin{align}
\label{eq: PAER}
    \min_{\mathbf{\Pi} \in \mathbb{R}^{n_i \times n_{i+1}}} \sum_{k, l} \Pi_{kl} \| \bm{x}_i (k)- \bm{x}_{i+1} (l) \|_2^2 - \epsilon 
    \sum_{k,l} \Pi_{kl}  \big(\log \big(\Pi_{kl} / [\widehat{\mathbf{M}}_i]_{kl}\big)-1 \big),
\end{align}
where $[\widehat{\mathbf{M}}_i]_{kl} = \exp(-[\mathbf{M}_i]_{kl}) / {\sum_{l} \exp(-[\mathbf{M}_i]_{kl})}$ denotes the $(k, l)$-th entry of the prior joint probability matrix induced by $\mathbf{M}_i$.
Note that the lower the cost $[\mathbf{M}_i]_{kl}$, the higher plausibility of the transition from cell $k$ at $t_i$ to cell $l$ at $t_{i+1}$. Consequently, the proposed entropy regularization term in Equation~\ref{eq: PAER} biases the learned transport plan toward the prior $\widehat{\mathbf{M}}_i$ rather than a uniform baseline, providing a soft mechanism for incorporating the intended biological prior knowledge.

\shortsection{ContextFlow}
Finally, we apply the Sinkhorn algorithm~\citep{1_Cuturi_Sinkhorn_fast} to obtain the spatial context-aware EOT couplings by solving optimization problems \ref{eq: PACM} and \ref{eq: PAER} for minibatch samples, and train the neural velocity vector field $u_\theta$ by minimizing the regression loss defined by Equation \ref{eq: conditional flow matching loss} with respect to conditionals $p_t(\bm{x} | \bm{z}_i)$ and $u_t(\bm{x} | \bm{z}_i)$ defined according to Equation \ref{eq: multi-time sampling path}. 
The pseudocode of ContextFlow 
is presented in Algorithm \ref{alg: contextflow} in Appendix \ref{apdx: alg pseudocode}, and we analyze its time complexity in Appendix \ref{apdx:complexity_analysis}.
In particular, to apply the Sinkhorn algorithm to solve our PAER problem, we make use of the following theorem, a generalization of \citet{peyre2019computational}.

\begin{theorem}
\label{thm: PAER Sinkhorn}
Let $\mathbf{C}\in \mathbb{R}^{n_0 \times n_1}$ be a cost matrix and $\mathbf{M}\in\mathbb{R}^{n_0 \times n_1}$ be a prior transition probability matrix. Suppose $\mathbf{\Pi}^*_{\mathrm{CTF-H}}$ is the solution to the following prior-aware optimal transport problem:
\begin{align*}
    \min_{\mathbf{\Pi} \in \mathbb{R}^{n_0 \times n_1}} \sum_{k, l} \Pi_{kl} C_{kl} - \epsilon \sum_{k,l} \Pi_{kl}  \big(\log (\Pi_{kl} / {M}_{kl})-1 \big),
\end{align*}
where $\epsilon>0$ is the regularization parameter.
Then, we can show that $\mathbf{\Pi}^*_{\mathrm{CTF-H}}$ can be computed by Sinkhorn and takes the form $\mathrm{diag}(\bm{u}) \cdot \mathbf{M} \odot \exp( -\mathbf{C} / \epsilon) \cdot \mathrm{diag}(\bm{v})$, where $\odot$ denotes element-wise multiplication, and $\bm{u}\in\mathbb{R}^{n_0}, \bm{v}\in\mathbb{R}^{n_1}$ are vectors satisfying the marginalization constraints.
\end{theorem}

Theorem \ref{thm: PAER Sinkhorn}, proven in Appendix \ref{apdx:proof of theorem PAER sinkhorn}, suggests a new Gibbs kernel $\mathbf{K} = \mathbf{M} \odot \exp(-\mathbf{C} / \epsilon)$, which combines both the transport cost matrix $\mathbf{C}$ and the prior joint probability matrix $\mathbf{M}$. 
When $\epsilon \rightarrow 0$, $\mathbf{\Pi}^*_{\mathrm{CTF-H}} \rightarrow \mathbf{\Pi}_{\mathrm{OT}}^*$, suggesting it recovers the standard OT couplings (Equation \ref{eq: Kantorovich's formulation}). 
When $\epsilon \rightarrow \infty$, $\mathbf{\Pi}^*_{\mathrm{CTF-H}} \rightarrow \mathrm{diag}(\bm{u}) \cdot \mathbf{M} \cdot \mathrm{diag}(\bm{v})$, suggesting a plan that aligns with the prior defined by $\mathbf{M}$ rather than the independent couplings obtained with EOT defined by Equation \ref{eq: EOT formulation}. This has the same effect as constraining our transport plan through the proposed prior and, by extension, the flow. By varying the parameter $\epsilon$, we can thus efficiently optimize for a desirable coupling via the Sinkhorn algorithm.



\section{Experiments}
\label{sec: experiments}

\shortsection{Datasets}
We evaluate ContextFlow on three longitudinal spatial transcriptomics datasets: Axolotl Brain Regeneration~\citep{1_dataset_steroseq}, Mouse Embryo Organogenesis~\citep{2_dataset_mosta}, and Liver Regeneration~\citep{3_dataset_liver}. For all the evaluated datasets, the gene expression values are log-normalized, and we extract the top $50$ principal components (PCs) as feature vectors. The strength of ligand–receptor interactions in the microenvironment was inferred using spatially informed bivariate statistics implemented in LIANA+~\citep{14_Dimitrov_Liana}, where we applied the cosine similarity metric to gene expression profiles. Interaction evidence was aggregated using the consensus of multiple curated ligand–receptor resources, ensuring robustness of the inferred signals.


\shortsection{Baselines \& Metrics} 
We evaluate ContextFlow with PACM and PAER (CTF-C and CTF-H) against several state-of-the-art FM baselines, including conditional flow matching (CFM)~\citep{lipman2023flow}, minibatch-OT flow matching (MOTFM)~\citep{tong2024improving}, metric flow matching (Metric-FM)~\citep{kapusniak2024metric}, and meta flow matching (MFM)~\citep{atanackovic2024meta}.
To evaluate statistical fidelity, we measure $2$-Wasserstein distance ($\mathcal{W}_{2}$), maximum mean discrepancy (MMD), and energy distance (Energy). To assess the biological plausibility of inferred dynamics, we employ a cell-type-weighted $2$-Wasserstein distance (Weighted $\mathcal{W}_{2}$), where the weights correspond to the relative frequency of each cell type in the dataset. Formal definitions of these evaluation metrics are detailed in Appendix~\ref{apdx:metrics}. To account for randomness, all reported metrics are averaged across $10$ runs.

\begin{table*}[t]
\caption{Comparisons under interpolation at the middle holdout time point on Brain Regeneration.}
\vspace{-0.05in}
\label{tab:stereoseq_interpolation_table}
\centering
\small
\setlength{\tabcolsep}{5pt} 
\resizebox{1.0\textwidth}{!}{
\begin{tabular}{l l | cccc}
\toprule
\textbf{Sampling} & \textbf{Method} & \textbf{Weighted} $\bm{\mathcal{W}_{2}}$ & $\bm{\mathcal{W}_{2}}$ & \textbf{MMD} & \textbf{Energy} \\
\midrule
\multirow{6}{*}{Next Step} 
& CFM & $2.618 \pm 0.142$ & $2.579 \pm 0.197$ & $0.043 \pm 0.003$ & $12.505 \pm 1.271$ \\
& MOTFM & $2.567 \pm 0.088$ & $2.476 \pm 0.161$ & $0.040 \pm 0.003$ & $11.269 \pm 1.388$ \\
& Metric-FM & $2.697 \pm 0.044$ & $2.688 \pm 0.024$ & $\bm{0.029 \pm 0.001}$ & $9.262 \pm 0.311$ \\
& MFM & $5.809 \pm 0.140$ & $5.851 \pm 0.156$ & $0.124 \pm 0.002$ & $49.770 \pm 0.479$ \\
\cmidrule{2-6}
& CTF-C ($\lambda=0,\alpha=0.2$) & $2.396 \pm 0.028$ & $2.100 \pm 0.102$ & $0.033 \pm 0.003$ & $8.577 \pm 0.976$ \\
& CTF-H ($\lambda=1$) & $\bm{2.316 \pm 0.141}$ & $\bm{1.969 \pm 0.221}$ & $0.030 \pm 0.004$ & $\bm{6.359 \pm 1.336}$ \\
\midrule
\multirow{6}{*}{IVP} 
& CFM & $4.216 \pm 0.463$ & $4.266 \pm 0.308$ & $0.170 \pm 0.029$ & $32.413 \pm 5.122$ \\
& MOTFM & $4.198 \pm 0.319$ & $4.452 \pm 0.243$ & $0.173 \pm 0.017$ & $33.149 \pm 3.321$ \\
& Metric-FM & $4.319 \pm 0.299$ & $4.501 \pm 0.409$ & $0.226 \pm 0.007$ & $34.941 \pm 5.063$ \\
& MFM & $5.976 \pm 0.133$ & $5.699 \pm 0.311$ & $0.177 \pm 0.010$ & $51.024 \pm 0.859$ \\
\cmidrule{2-6}
& CTF-C ($\lambda=0,\alpha=0.2$) 
& $\bm{3.465 \pm 0.232}$ & $\bm{3.641 \pm 0.320}$ 
& $0.119 \pm 0.025$ & $23.055 \pm 5.939$ \\
& CTF-H ($\lambda=1$) 
& $3.905 \pm 0.395$ & $4.188 \pm 0.685$ 
& $\bm{0.074 \pm 0.014}$ & $\bm{18.728 \pm 2.689}$ \\
\bottomrule
\end{tabular}
}
\vspace{-0.1in}
\end{table*}

\shortsection{Sampling} 
A learned velocity field can be evaluated through the samples it generates. 
We consider two configurations: \emph{initial-value-problem} (IVP) and \textit{next-step} (Next Step) sampling.
In particular, IVP integrates the gradients learned from the first observed batch of cells and evolves them to a later time point, providing the most comprehensive evaluation of flow quality, as errors can accumulate across steps. Next Step integrates the gradient only from the most recent batch of cells, thus limiting error propagation but providing a less stringent test of long-term trajectory fidelity.

\subsection{Main Results on Axolotl Brain Regeneration}

First, we evaluate our ContextFlow method on longitudinal Stereo-seq spatial transcriptomic data from a post-traumatic brain regeneration study of the Salamander (axolotl telencephalon) species~\citep{1_dataset_steroseq}. The dataset contains samples from $5$ developmental stages, with replicates collected from different individual organisms at each stage. For our CTF-C method, we present the best ablated $\alpha$, with full ablation results across different $\alpha$ values in Appendix~\ref{apdx:Ablations}.

\shortsection{Interpolation}
For interpolation, we hold out the middle time point during training and evaluate it using samples generated by the trained velocity field $u_{\theta}$ via both IVP and Next-Step sampling. Table~\ref{tab:stereoseq_interpolation_table} presents the results. 
Across multiple evaluation metrics, ContextFlow with entropy regularization (CTF-H) produces trajectories that most closely match the ground truth. CTF-H consistently achieves the best or comparable performance relative to baselines and CTF-C, despite the latter being explicitly tuned across multiple $\alpha$ values. This highlights the computational efficiency and superior generalization of CTF-H, which avoids the need for additional hyperparameter tuning while maintaining strong performance.


\begin{table*}[t]
\caption{Comparisons under extrapolation on the last holdout time point on Brain Regeneration.}
\vspace{-0.05in}
\label{tab:stereoseq_extrapolation_table}
\centering
\small
\setlength{\tabcolsep}{5pt} 
\resizebox{1.0\textwidth}{!}{
\begin{tabular}{l l | cccc}
\toprule
\textbf{Sampling} & \textbf{Method} & \textbf{Weighted} $\bm{\mathcal{W}_{2}}$ & $\bm{\mathcal{W}_{2}}$ & \textbf{MMD} & \textbf{Energy} \\
\midrule
\multirow{6}{*}{Next Step} 
& CFM & $7.124 \pm 0.443$ & $7.133 \pm 0.533$ & $0.275 \pm 0.011$ & $76.947 \pm 5.661$ \\
& MOTFM & $7.619 \pm 0.611$ & $7.769 \pm 0.763$ & $0.272 \pm 0.007$ & $85.352 \pm 8.140$ \\
& Metric-FM & $10.125 \pm 1.863$ & $10.391 \pm 1.968$ & $0.181 \pm 0.022$ & $112.807 \pm 24.325$ \\
& MFM & $\bm{6.868 \pm 0.348}$ & $6.968 \pm 0.051$ & $\bm{0.175 \pm 0.060}$ & $81.158 \pm 3.677$ \\
\cmidrule{2-6}
& CTF-C ($\lambda=0.5,\alpha=0.5$) 
& $7.188 \pm 0.391$ & $\bm{6.931 \pm 0.260}$ 
& $0.267 \pm 0.005$ & $78.992 \pm 6.195$ \\
& CTF-H ($\lambda=0$) 
& $6.914 \pm 0.471$ & $7.198 \pm 0.726$ 
& $0.266 \pm 0.009$ & $\bm{76.149 \pm 8.436}$ \\
\midrule
\multirow{6}{*}{IVP} 
& CFM & $6.633 \pm 1.312$ & $7.116 \pm 1.084$ & $0.143 \pm 0.037$ & $60.573 \pm 21.756$ \\
& MOTFM & $6.503 \pm 0.720$ & $6.352 \pm 0.592$ & $0.162 \pm 0.038$ & $56.452 \pm 15.932$ \\
& Metric-FM & $8.213 \pm 4.959$ & $6.972 \pm 2.557$ & $0.124 \pm 0.034$ & $29.155 \pm 13.973$ \\
& MFM & $13.267 \pm 1.802$ & $13.513 \pm 1.908$ & $0.080 \pm 0.001$ & $109.714 \pm 26.407$ \\
\cmidrule{2-6}
& CTF-C ($\lambda=0.5,\alpha=0.5$) 
& $6.696 \pm 0.427$ & $6.481 \pm 0.387$ 
& $0.195 \pm 0.024$ & $66.212 \pm 3.542$ \\
& CTF-H ($\lambda=1$) 
& $\bm{5.277 \pm 0.936}$ & $\bm{6.021 \pm 1.192}$ 
& $\bm{0.099 \pm 0.007}$ & $\bm{27.777 \pm 8.621}$ \\
\bottomrule
\end{tabular}
}
\vspace{-0.05in}
\end{table*}

\shortsection{Extrapolation}
For extrapolation, we evaluate the data generation at the last holdout time point, which represents the most challenging test of generalizability for the velocity fields $u_{\theta}$, as it lies outside the training time horizon. As shown in Table~\ref{tab:stereoseq_extrapolation_table}, CTF-H again achieves the best overall performance, particularly under IVP sampling, where errors are most likely to accumulate. This result further reinforces the robustness and reliability of CTF-H across the entire sampling horizon.
Figure~\ref{fig:implausible_transitions_comparison} in Appendix \ref{apdx:vizualization_implausibility} further illustrates that incorporating spatial priors enables ContextFlow to produce substantially fewer biologically implausible couplings compared to its context-free counterpart.

\begin{table*}
\caption{Comparisons under interpolation (time $5$) and extrapolation (time $8$) on Organogenesis.}
\vspace{-0.05in}
\label{tab:mosta_combined_table}
\centering
\small
\renewcommand{\arraystretch}{1.05} 
\resizebox{1.0\textwidth}{!}{
\begin{tabular}{l | cc | cc | cc}
\toprule
\textbf{Method} 
& \multicolumn{2}{c|}{\textbf{Next Step (Interpolation)}} 
& \multicolumn{2}{c|}{\textbf{IVP (Interpolation)}} 
& \multicolumn{2}{c}{\textbf{Next Step (Extrapolation)}} \\
\cmidrule(lr){2-3} \cmidrule(lr){4-5} \cmidrule(lr){6-7}
& \textbf{Weighted} $\bm{\mathcal{W}_{2}}$ & $\bm{\mathcal{W}_{2}}$ 
& \textbf{Weighted} $\bm{\mathcal{W}_{2}}$ & $\bm{\mathcal{W}_{2}}$
& \textbf{Weighted} $\bm{\mathcal{W}_{2}}$ & $\bm{\mathcal{W}_{2}}$ \\
\midrule

MOTFM 
& $1.892 \pm 0.028$ & $1.873 \pm 0.086$ 
& $3.251 \pm 0.676$ & $3.418 \pm 0.727$ 
& $1.626 \pm 0.066$ & $1.682 \pm 0.096$ \\

Metric-FM 
& $1.897 \pm 0.036$ & $1.864 \pm 0.084$ 
& $4.030 \pm 1.552$ & $3.826 \pm 0.767$ 
& $1.595 \pm 0.114$ & $1.692 \pm 0.272$ \\

MFM 
& $\bm{1.839 \pm 0.065}$ & $1.901 \pm 0.099$ 
& $3.476 \pm 0.371$ & $3.319 \pm 0.281$ 
& $1.607 \pm 0.049$ & $1.407 \pm 0.072$ \\

\cmidrule{1-7}

CTF-C ($\lambda=1,\alpha=0.5$) 
& $1.865 \pm 0.030$ & $\bm{1.852 \pm 0.093}$ 
& $3.137 \pm 0.407$ & $4.093 \pm 1.187$ 
& $1.685 \pm 0.096$ & $1.714 \pm 0.160$ \\

CTF-H ($\lambda=0.5$) 
& $1.871 \pm 0.030$ & $1.919 \pm 0.067$ 
& $\bm{2.814 \pm 0.414}$ & $\bm{3.233 \pm 0.567}$ 
& $1.636 \pm 0.060$ & $1.684 \pm 0.099$ \\

CTF-H ($\lambda=0$) 
& $1.884 \pm 0.027$ & $1.862 \pm 0.123$ 
& $3.244 \pm 0.713$ & $3.946 \pm 1.671$ 
& $\bm{1.505 \pm 0.057}$ & $\bm{1.397 \pm 0.088}$ \\

\bottomrule
\end{tabular}
}
\vspace{-0.1in}
\end{table*}

\begin{table*}[t]
\caption{Comparisons under interpolation on the middle holdout time point on Liver Regeneration.}
\vspace{-0.05in}
\label{tab:liver_interpolation_table}
\centering
\small
\resizebox{1.0\textwidth}{!}{
\begin{tabular}{l | c |ccc|ccc}
\toprule
& \textbf{MOTFM} & \multicolumn{3}{c|}{\textbf{CTF-C}} & \multicolumn{3}{c}{\textbf{CTF-H}} \\
\cmidrule(lr){2-2} \cmidrule(lr){3-5} \cmidrule(lr){6-8}
\textbf{($\lambda$, $\alpha$)} & -- & ($1$, $0.5$) & ($0$, $0.5$) & ($0.5$, $0.8$) & ($0$, --) & ($1$, --) & ($0.5$, --) \\
\midrule
$\bm{\mathcal{W}_{2}}$ & $34.303 \pm 1.448$ & $33.506 \pm 1.148$ & $32.741 \pm 1.864$ & $33.045 \pm 1.644$ & $\bm{32.682 \pm 1.472}$ & $33.481 \pm 1.001$ & $33.414 \pm 0.995$ \\
\bottomrule
\end{tabular}
}
\vspace{-0.05in}
\end{table*}

\subsection{Additional Evaluations on Organogenesis and Liver Regeneration}

\shortsection{Organogenesis}
We further evaluate ContextFlow on the larger Mouse Organogenesis Spatiotemporal Atlas (MOSTA) Stereo-seq dataset~\citep{2_dataset_mosta} spanning measurements from $8$ developmental time points. For the interpolation study of this dataset, we hold out the time point $5$ during training and evaluate its generation during testing. Table~\ref{tab:mosta_combined_table} reports the evaluation results. We observe that ContextFlow, with both integration strategies, performs best overall relative to baselines across all metrics, showcasing the effectiveness of the contextual information. While CTF-C shows stronger performance under next-step sampling—albeit only after fine-tuning the trade-off parameter $\alpha$—CTF-H consistently outperforms it in the more challenging IVP sampling setting. 
On the extrapolation task, integrating to the final time point, CTF-H again achieves the strongest performance, underscoring that the entropy-regularized formulation not only removes the need for additional parameter tuning but also offers more robust generalization to unseen temporal horizons. 

\shortsection{Liver Regeneration}
Lastly, we evaluate ContextFlow on a Visium spatial transcriptomics dataset profiling the temporal dynamics of mouse liver regeneration following acetaminophen-induced injury~\citep{3_dataset_liver}, which is collected across $3$ distinct regeneration stages. Unlike previous datasets resolved at single-cell resolution, Visium data is captured at the level of $55$-micron-diameter spots, revealing joint expression across multiple cells. Since direct cell-type information is unavailable, we restrict evaluation to the $2$-Wasserstein distance. Moreover, since evaluation is performed midway through the $3$ time points, the IVP and next-step predictions coincide. 
Table \ref{tab:liver_interpolation_table} summarizes the comparisons between MOTFM and our methods. Consistent with our previous findings, CTF-H achieves the lowest reconstruction error, indicating that incorporating contextual information improves trajectory estimation even in aggregated spot-level measurements.

\begin{figure}[t]
\centering
    \begin{subfigure}{0.235\textwidth}
        \centering
        \includegraphics[width=\linewidth, height=0.6\linewidth]{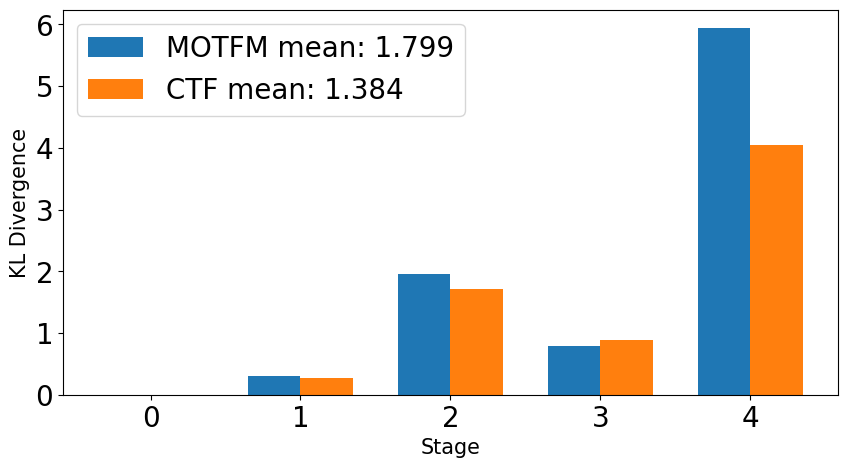}
        \caption{Brain Regeneration}
        \label{fig:IVP_celltype_kldiv_brain}
    \end{subfigure}
    \hspace{0.02in}
    \begin{subfigure}{0.235\textwidth}
        \centering
        \includegraphics[width=\linewidth, height=0.6\linewidth]{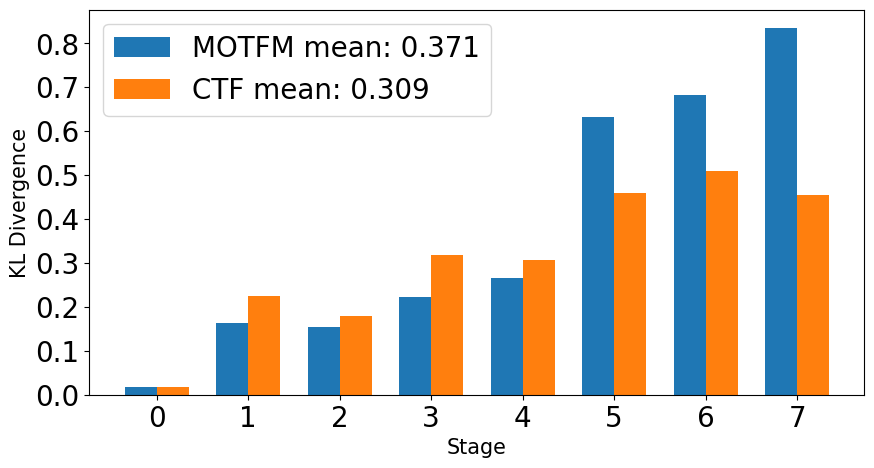}
        \caption{Organogenesis}
        \label{fig:IVP_celltype_kldiv_organogenesis}
    \end{subfigure}
    \hspace{0.02in}
    \begin{subfigure}{0.235\textwidth}
        \centering
        \includegraphics[width=\linewidth, height=0.6\linewidth]{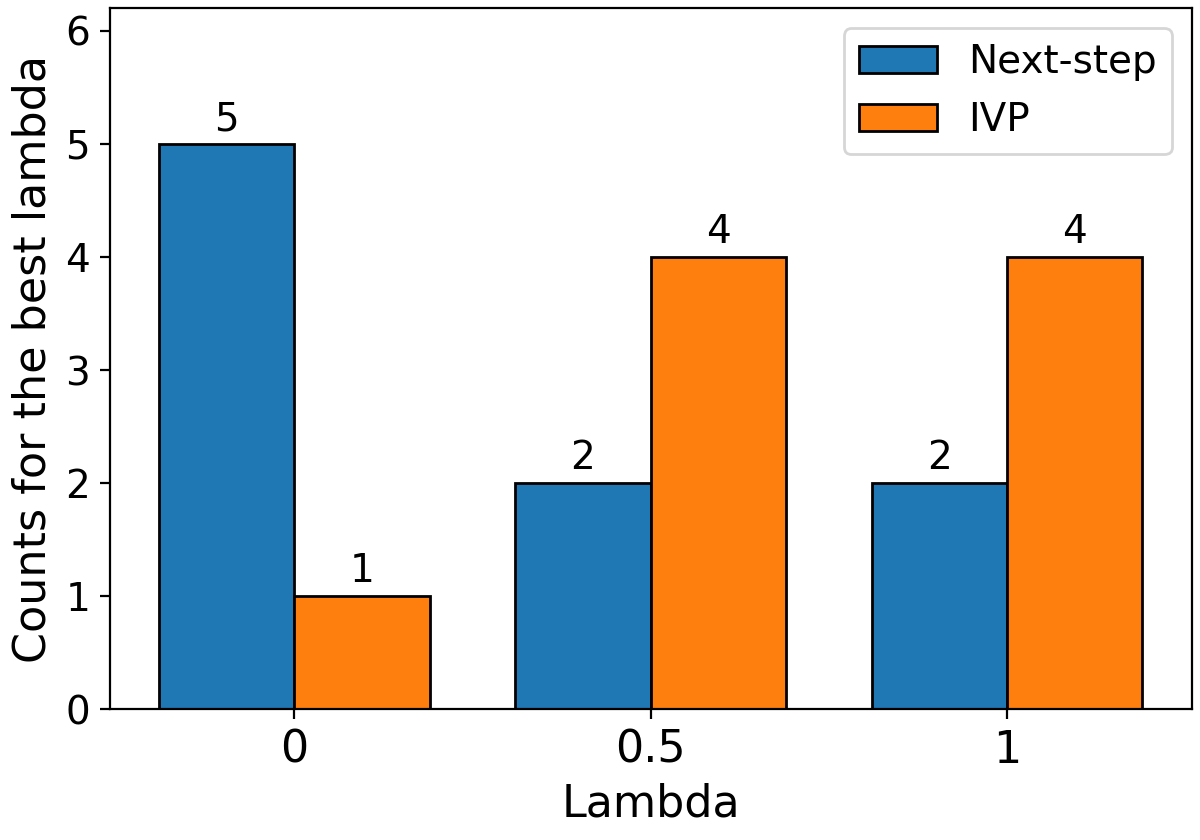}
        \caption{Ablations on $\lambda$}
        \label{fig:lambda_counts}
    \end{subfigure}
    \hspace{0.02in}
    \begin{subfigure}{0.235\textwidth}
        \centering
        \includegraphics[width=\linewidth, height=0.6\linewidth]{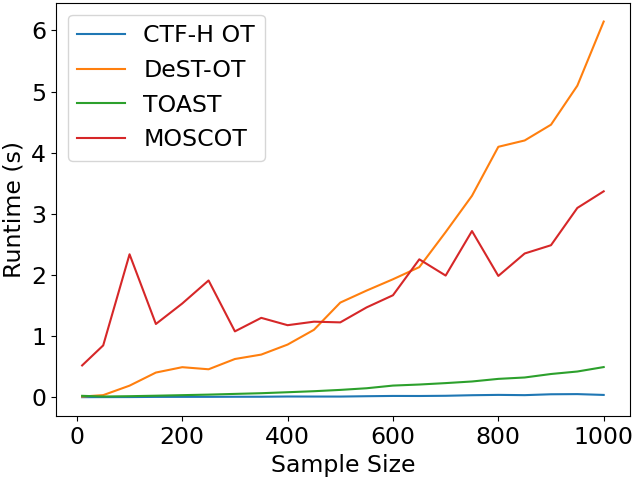}
        \caption{OT Runtime}
        \label{fig:ot_runtime}
    \end{subfigure}
\vspace{-0.05in}
\caption{(a) and (b) compare ContextFlow with MOTFM in terms of the KL divergence between predicted and true cell type distributions under IVP sampling on two datasets, (c) summarizes the ablations on $\lambda$ in terms of the optimal choice counts, and (d) visualizes the runtime curves with respect to varying sample sizes across different OT-based coupling schemes.}
\vspace{-0.1in}
\end{figure}



\subsection{Further Analysis}



\shortsection{Cell Type Distribution}
Figures~\ref{fig:IVP_celltype_kldiv_brain} and \ref{fig:IVP_celltype_kldiv_organogenesis} demonstrate the KL divergence between normalized histograms of predicted and ground-truth cell types. Compared with MOTFM, ContextFlow exhibits lower average divergence across time points, indicating that the inferred trajectories better preserve the biological composition of cell types. 
The cell type progression is further visualized in Figure~\ref{fig:sankey} in Appendix \ref{apdx:IVP cell type progression}, showing the temporal trajectories of cell types evolve smoothly and consistently across consecutive developmental stages. Early progenitor populations, such as neural crest and mesenchyme, progressively diminish as development progresses, while terminal fates, including muscle, the cartilage primordium, and the liver, emerge at later stages. Major lineages such as brain, heart, and connective tissue remain continuous throughout, confirming that ContextFlow captures biologically coherent and temporally consistent developmental dynamics. 

\shortsection{Relative Importance of Priors}
We conduct extensive ablation studies on CTF-H and report them in Appendix~\ref{apdx:Hyperparameter_ablations}. To summarize trends across datasets and settings, we report the counts of the optimal $\lambda$ values in Figure~\ref{fig:lambda_counts}. For IVP sampling, $\lambda=0.5/1$ (both useful, with a bias towards SS) are more frequently optimal settings, while for next-step sampling, $\lambda=0$ (bias towards LR) is the most frequent. This behavior can be explained by the different roles of the priors. The SS prior ($\lambda=1$) serves as a proxy for relative spatial consistency across tissue microenvironments, providing a global structural constraint that remains informative over long integration horizons and is therefore beneficial for IVP sampling. In contrast, the LR prior is informative depending on how distinct the cell-cell communication patterns are over a given time interval, making it more localized and time-dependent, particularly informative for short-range transitions, and thus more beneficial for next-step sampling.

\shortsection{Computational Advantage} 
We compare the performance of CTF-H OT, the coupling scheme used in ContextFlow, against state-of-the-art spatiotemporal OT methods, including DeST-OT~\citep{halmos2025dest}, MOSCOT~\citep{8_Klien_moscot}, and
TOAST~\citep{ceccarelli2026topography}. 
Figure~\ref{fig:ot_runtime} reports the runtime of different OT methods on the Brain Regeneration dataset with respect to varying sample sizes, where CTF-H OT is significantly faster than prior methods ($270$x for DeST-OT, $257$x for MOSCOT, and $24$x for TOAST), while achieving comparable OT alignment performance (see Tables~\ref{tab:growth_distortion}-\ref{tab:ctpi_metric} in Appendix~\ref{apdx:static_ot} for comparisons in growth distortion, migration, and coupled transcription distance metrics). These results suggest that the proposed prior integration strategy is more suited for flow matching frameworks,
where thousands of OT problems are involved during training. 

\section{Conclusion}

We introduced ContextFlow, a contextually aware flow matching framework that leverages spatial priors and biologically motivated constraints to learn more plausible trajectories from snapshot spatial transcriptomic data. 
Our experiments showed that the entropic variant of ContextFlow (CTF-H) outperforms existing baselines, even in challenging IVP sampling settings.
In particular, our method effectively reduces the number of biologically implausible couplings and generates more temporally consistent developmental trajectories, while maintaining strong quantitative performance across $2$-Wasserstein, MMD, and Energy metrics. These results highlight the value of embedding proper biological context into flow-based generative models. Future work can adapt our methods to reconstruct tissues and learn spatial latent dynamics by formulating the flow in space (rather than time), or by leveraging multi-marginal OT formulations to optimize temporal flows.
Looking forward, our ContextFlow framework offers a principled foundation for modeling perturbations and disease progression, potentially bridging generative power with biological interpretability.

\bibliography{flow_matching.bib}
\bibliographystyle{plainnat}
\newpage
\section*{NeurIPS Paper Checklist}

\begin{enumerate}

\item {\bf Claims}
    \item[] Question: Do the main claims made in the abstract and introduction accurately reflect the paper's contributions and scope?
    \item[] Answer: \answerYes{}
    \item[] Justification: Our main claims focus on demonstrating the limitations of prior flow matching techniques and on our proposed solution to mitigate them. Section~\ref{sec: problem} discusses the limitations of prior techniques, Section~\ref{sec: regularizing the flow} presents the proposed methodology, and Section~\ref{sec: experiments} reports comprehensive experimental results against baselines to validate our claims. Additionally, we defer more analysis to Appendix~\ref{apdx:pacm} (effect of normalization in OT cost), Appendix~\ref{apdx:vizualization_implausibility} (biological plausibility analysis), and Appendices~\ref{apdx:Hyperparameter_ablations} and~\ref{apdx:Ablations} (full ablation studies).
    
\item {\bf Limitations}
    \item[] Question: Does the paper discuss the limitations of the work performed by the authors?
    \item[] Answer: \answerYes{} 
    \item[] Justification: Main limitations of our work include poorly specified or noisy priors, which may bias the learned dynamics. Furthermore, while our evaluation demonstrates strong statistical and biological coherence, validation using experimental or longitudinal ground truth data would strengthen confidence in the inferred trajectories.

\item {\bf Theory assumptions and proofs}
    \item[] Question: For each theoretical result, does the paper provide the full set of assumptions and a complete (and correct) proof?
    \item[] Answer: \answerYes{}
    \item[] Justification: Yes, the formal statements of our theorems list all the necessary assumptions. Appendix~\ref{apdx: proofs} provides their detailed proofs.

    \item {\bf Experimental result reproducibility}
    \item[] Question: Does the paper fully disclose all the information needed to reproduce the main experimental results of the paper to the extent that it affects the main claims and/or conclusions of the paper (regardless of whether the code and data are provided or not)?
    \item[] Answer: \answerYes{} 
    \item[] Justification: We've included the information to fully describe the algorithm in Appendix~\ref{apdx: contextflow}, as well as the data processing, splits, and the settings of the packages used in our studies in Section~\ref{sec: experiments} and Appendix~\ref{apdx: training_details}.

\item {\bf Open access to data and code}
    \item[] Question: Does the paper provide open access to the data and code, with sufficient instructions to faithfully reproduce the main experimental results, as described in supplemental material?
    \item[] Answer: \answerNo{} 
    \item[] Justification: The refactored source code will be released after the review process.

\item {\bf Experimental setting/details}
    \item[] Question: Does the paper specify all the training and test details (e.g., data splits, hyperparameters, how they were chosen, type of optimizer) necessary to understand the results?
    \item[] Answer: \answerYes{} 
    \item[] Justification: We've included the information to fully describe the algorithm in Appendix~\ref{apdx: contextflow}, as well as the data processing, splits, and the settings of the packages used in our studies in Section~\ref{sec: experiments} and Appendix~\ref{apdx: training_details}.

\item {\bf Experiment statistical significance}
    \item[] Question: Does the paper report error bars suitably and correctly defined or other appropriate information about the statistical significance of the experiments?
    \item[] Answer: \answerYes{} 
    \item[] Justification: We've provided both the mean and standard deviation statistics across random seeds for all the reported metrics, and extensive ablations in Appendices~\ref{apdx:Hyperparameter_ablations} and~\ref{apdx:Ablations}.

\item {\bf Experiments compute resources}
    \item[] Question: For each experiment, does the paper provide sufficient information on the computer resources (type of compute workers, memory, time of execution) needed to reproduce the experiments?
    \item[] Answer: \answerYes{} 
    \item[] Justification: We've included information in terms of data processing, splits, settings of packages used in our studies, runtime, etc. Importantly, all our experiments were run on a single Nvidia DGX A100.
    
\item {\bf Code of ethics}
    \item[] Question: Does the research conducted in the paper conform, in every respect, with the NeurIPS Code of Ethics \url{https://neurips.cc/public/EthicsGuidelines}?
    \item[] Answer: \answerYes{} 
    \item[] Justification: We've read the guidelines and confirm that we're not in violation of any code of ethics.

\item {\bf Broader impacts}
    \item[] Question: Does the paper discuss both potential positive societal impacts and negative societal impacts of the work performed?
    \item[] Answer: \answerNo{}
    \item[] Justification: The paper presents a general principled methodology to propose a flow matching model for modeling spatial omics in general, rather than specific applications with direct societal implications.
    
\item {\bf Safeguards}
    \item[] Question: Does the paper describe safeguards that have been put in place for responsible release of data or models that have a high risk for misuse (e.g., pre-trained language models, image generators, or scraped datasets)?
    \item[] Answer: \answerNA{} 
    \item[] Justification: \answerNA{}

\item {\bf Licenses for existing assets}
    \item[] Question: Are the creators or original owners of assets (e.g., code, data, models), used in the paper, properly credited and are the license and terms of use explicitly mentioned and properly respected?
    \item[] Answer: \answerNA{}
    \item[] Justification: \answerNA{}

\item {\bf New assets}
    \item[] Question: Are new assets introduced in the paper well documented and is the documentation provided alongside the assets?
    \item[] Answer: \answerNA{} 
    \item[] Justification: \answerNA{}

\item {\bf Crowdsourcing and research with human subjects}
    \item[] Question: For crowdsourcing experiments and research with human subjects, does the paper include the full text of instructions given to participants and screenshots, if applicable, as well as details about compensation (if any)? 
    \item[] Answer: \answerNA{}
    \item[] Justification: \answerNA{}

\item {\bf Institutional review board (IRB) approvals or equivalent for research with human subjects}
    \item[] Question: Does the paper describe potential risks incurred by study participants, whether such risks were disclosed to the subjects, and whether Institutional Review Board (IRB) approvals (or an equivalent approval/review based on the requirements of your country or institution) were obtained?
    \item[] Answer: \answerNA{}
    \item[] Justification: \answerNA{}

\item {\bf Declaration of LLM usage}
    \item[] Question: Does the paper describe the usage of LLMs if it is an important, original, or non-standard component of the core methods in this research? Note that if the LLM is used only for writing, editing, or formatting purposes and does \emph{not} impact the core methodology, scientific rigor, or originality of the research, declaration is not required.
    \item[] Answer: \answerNA{} 
    \item[] Justification: \answerNA{}

\end{enumerate}

\appendix
\onecolumn

\appendix




\section{Formal Definitions of Evaluation Metrics}
\label{apdx:metrics}

\shortsection{2-Wasserstein} 
The \textit{2-Wasserstein distance} ($\mathcal{W}_2$ between empirical distributions $\mu,\nu$ is defined as:
\[
\mathcal{W}_2(\mu,\nu) = \inf_{\gamma \in \Pi(\mu,\nu)} 
\bigg( \sum_{(\bm{x}, \bm{y})} \gamma(\bm{x}, \bm{y}) \cdot \| \bm{x} - \bm{y}\|_2^2 \bigg)^{1/2},
\]
where $\Pi(\mu,\nu)$ denotes the set of couplings between $\mu$ and $\nu$.

\shortsection{Weighted 2-Wasserstein}
Implausible velocity fields can steer a cell's transcriptional trajectory in unrealistic directions, potentially leading to entirely different terminal cell types. We thus employ the weighted 2-Wasserstein metric, which ensures the evaluation accounts for both transcriptional similarity and the distributional balance of cell types.
We define the \textit{weighted 2-Wasserstein distance} (Weighted $\mathcal{W}_2$) between true and predicted distributions as:  
\begin{align*}
\text{Weighted-$\mathcal{W}_{2}$}(\mu, \nu) = \sum_{i=1}^{C} \frac{n_i^{\text{true}}}{N} \cdot 
\mathcal{W}_{2}\!\bigg(\frac{1}{n_i^{\text{true}}} \!\!\sum_{j: y_j = i} \delta_{\bm{x}_j}, 
\frac{1}{n_i^{\text{pred}}} \!\!\sum_{j: \hat{y}_j = i} \delta_{\bm{x}_j}\bigg),
\end{align*}
where $n_i^{\text{true}}, n_i^{\text{pred}}$ are the number of true and predicted cells of type $i$, and $N$ is the total number of samples. To determine the cell type of generated trajectories, we employ a multi-class classifier $M_{\phi}$, implemented as an XGBoost model~\citep{14_Chen_xgboost} trained for each dataset.

\shortsection{Energy Distance} 
Let \(\mu\) and \(\nu\) be probability distributions with samples 
\(X = \{\bm{x}_i\}_{i=1}^m \sim \mu\) and \(Y = \{\bm{y}_j\}_{j=1}^n \sim \nu\). 
The squared empirical \textit{energy distance} (Energy) is defined as:
\begin{align*}
\mathrm{ED}(\mu, \nu) 
= \frac{2}{mn} \sum_{i=1}^m \sum_{j=1}^n \|\bm{x}_i - \bm{y}_j\| 
- \frac{1}{m^2} \sum_{i=1}^m \sum_{i'=1}^m \|\bm{x}_i - \bm{x}_{i'}\| 
- \frac{1}{n^2} \sum_{j=1}^n \sum_{j'=1}^n \|\bm{y}_j - \bm{y}_{j'}\|,
\end{align*}
where \(\|\cdot\|\) is the Euclidean norm. 
The distance is non-negative and equals zero if and only if \(\mu = \nu\). 

\shortsection{Maximum Mean Discrepancy} 
For the same samples, the unbiased empirical estimate of the squared \textit{maximum mean discrepancy} (MMD)
with kernel \(\kappa\) is defined as:
\begin{align*}
\mathrm{MMD}(\mu, \nu; \kappa) 
= \frac{1}{m(m-1)} \sum_{i \neq i'} \kappa(\bm{x}_i, \bm{x}_{i'}) 
+ \frac{1}{n(n-1)} \sum_{j \neq j'} \kappa(\bm{y}_j, \bm{y}_{j'}) 
- \frac{2}{mn} \sum_{i=1}^m \sum_{j=1}^n \kappa(\bm{x}_i, \bm{y}_j).
\end{align*}
In our evaluations, we use a multi-kernel variant with radial basis function (RBF) kernels 
\( \kappa_\gamma(\bm{x},y) = \exp(-\gamma \|\bm{x} - \bm{y}\|^2) \), 
and average over $\gamma \in [2, 1, 0.5, 0.1, 0.01, 0.005]$.



\section{Detailed Preliminaries on Flow Matching}
\label{sec:preliminaries}


\shortsection{Flow Matching Basics} 
Flow matching~\citep{lipman2023flow} is a simulation-free and sample-efficient generative framework for training continuous normalizing flows~\citep{3_neural_ODE}.
Given a pair of source and target distributions over $\mathbb{R}^d$ with probability densities $q_0=q(\bm{x}_0)$ and $q_1=q(\bm{x}_1)$, the problem task is to learn a time-varying velocity vector field $u_\theta: [0,1] \times \mathbb{R}^d \rightarrow \mathbb{R}^d$,  whose continuous evolution is captured by a function in the form of a neural-net-based model with weights $\theta$, that can transform $q_0$ to $q_1$ through integration via an ordinary differential equation (ODE).

To be more specific, \emph{flow matching} (FM) seeks to optimize $\theta$ by minimizing a simple regression loss between $u_\theta$ and a target time-varying velocity vector field $u_t: [0,1] \times \mathbb{R}^d \rightarrow \mathbb{R}^d$ as follows:
\begin{align}
\label{eq: flow matching regression loss apdx}
    \min_{\theta} \: \mathbb{E}_{t \sim \mathcal{U}(0,1), \bm{x} \sim p_t(\bm{x})} \big\| u_\theta(t, \bm{x}) - u_t(\bm{x}) \big\|^2.
\end{align}
Here, $\mathcal{U}(0,1)$ is the uniform distribution over $[0, 1]$, and $p_t: [0,1] \times \mathbb{R}^d \rightarrow \mathbb{R}_{+}$ denotes a time-varying probability path induced by $u_t$ such that (i) $p_t$ is a probability density function for any $t\in[0,1]$, (ii) $p_t$ satisfies the two boundary conditions: $p_{t=0} = q_0$ and $p_{t=1} = q_1$, and (iii) the connection between $p_t$ and $u_t$ can be characterized by the \emph{transport equation}~\citep{villani2008optimal}:
\begin{align}
\label{eq: continuity equation}
    \frac{\partial p_t(\bm{x})}{\partial t} = - \nabla \cdot (u_t(\bm{x}) p_t(\bm{x})),
\end{align}
where $\nabla$ is the divergence operator.
From a dynamical system's view, $u_t$ defines an ODE system $d\bm{x} = u_t(\bm{x}) dt$. The corresponding solution to the ODE, usually termed as the probability flow, can then transport any $\bm{x}_0\sim q_0$ to a point $\bm{x}_1\sim q_1$ along $u_t$ from $t=0$ to $t=1$.
While the FM objective in Equation~\ref{eq: flow matching regression loss apdx} is simple and intuitive, it is generally intractable in practice: the closed-form velocity vector field $u_t$ is unknown for arbitrary source and target distributions ($q_0$ and $q_1$), and multiple valid probability paths $p_t$ may exist between them.


\shortsection{Conditional Flow Matching} 
The central idea is to express the target probability path via a mixture of more manageable \emph{conditional probability paths}~\citep{lipman2023flow}.
By marginalizing over a conditioning variable $z$, both $p_t$ and $u_t$ can be constructed using their conditional counterparts: 
\begin{equation}
\label{eq: conditional probability path velocity vector field}
\begin{aligned}
p_t(\bm{x}) &= \int p_t(\bm{x} | \bm{z}) q(\bm{z}) d\bm{z} \quad \text{and} \quad u_t(\bm{x}) &= \int u_t(\bm{x} | \bm{z}) \frac{p_t(\bm{x} | \bm{z}) q(\bm{z})}{p_t(\bm{x})} d\bm{z},
\end{aligned}
\end{equation}
where $q(\bm{z})$ is the distribution of the conditioning variable $\bm{z}$, and the conditional probability path $p_t(\bm{x}|\bm{z})$ is constructed such that the boundary conditions are satisfied: $\int p_{t=0}(\bm{x} | \bm{z}) q(\bm{z}) = q_0$ and $\int p_{t=1}(\bm{x} | \bm{z}) q(\bm{z}) = q_1$. Theorem 1 of \citet{lipman2023flow} proves that $p_t$ and $u_t$ defined by Equation \ref{eq: conditional probability path velocity vector field} satisfy the transport equation, suggesting that $p_t$ is a valid probability path generated by $u_t$. 

To avoid the intractable integrals, \cite{lipman2023flow} proposed the following training objective of \emph{conditional flow matching}, and proved its equivalence to Equation \ref{eq: flow matching regression loss apdx} in terms of gradient computation:
\begin{align}
\label{eq: conditional flow matching loss apdx}
    \min_{\theta} \: \mathbb{E}_{t \sim \mathcal{U}(0,1), z \sim q(\bm{z}), \bm{x} \sim p_t(\bm{x}|\bm{z})} \big\| u_\theta(t, \bm{x}) - u_t(\bm{x} | \bm{z}) \big\|^2.
\end{align}
By choosing an appropriate conditional velocity vector field $u_t(\bm{x}|\bm{z})$, we can train the neural network using Equation \ref{eq: conditional flow matching loss apdx} without requiring a closed-form solution of the conditional probability path $p_t(\bm{x}|\bm{z})$, thereby avoiding the intractable integration operation.
Therefore, the remaining task is to define the conditional probability path and velocity vector field properly such that we can sample from $p_t(\bm{x}|\bm{z})$ and compute $u_t(\bm{x}|\bm{z})$ efficiently for solving the optimization problem in Equation \ref{eq: conditional flow matching loss apdx}.

\shortsection{Gaussian Conditional Probability Paths}
A specific choice proposed in \citet{lipman2023flow} is Gaussian conditional paths and their corresponding velocity vector fields, which are defined as:
\begin{align}
\label{eq: general Gaussian construction apdx}
    p_t(\bm{x} | \bm{z}) = \mathcal{N}(\bm{x} \: | \: \mu_t(\bm{z}), \sigma_t(\bm{z})^2 \mathbf{I}) \quad \text{and} \quad
    u_t(\bm{x} | \bm{z}) = \frac{\sigma_t'(\bm{z})}{\sigma_t(\bm{z})} (\bm{x} - \mu_t(\bm{z})) + \mu_t'(\bm{z}),
\end{align}
where $\mu_t: [0,1] \times \mathbb{R}^d \rightarrow \mathbb{R}^d$ stands for the time-varying mean of the Gaussian distribution, $\mu_t'$ is its derivative with respect to time, $\sigma_t: [0,1] \times \mathbb{R}^d \rightarrow \mathbb{R}_{+}$ is the time-varying scalar standard deviation, and $\sigma_t'$ denotes its derivative.
We refer to Theorem 3 of \citet{lipman2023flow} for a formal argument that $u_t(\bm{x} | \bm{z})$ uniquely generates the Gaussian conditional probability path $p_t(\bm{x} | \bm{z})$ for any differentiable $\mu_t(\bm{z})$ and $\sigma_t(\bm{z})$ in Equation \ref{eq: general Gaussian construction apdx} satisfying the boundary conditions.

In particular, \citet{lipman2023flow} selected $q(\bm{z}) = q(\bm{x}_1)$, $\mu_t(\bm{z}) = t \bm{x}_1$, and $\sigma_t(\bm{z}) = 1 - (1 - \sigma) t$. Then, we can see that $u_t(\bm{x} | \bm{z})$ transports the standard Gaussian distribution $p_{t=0}(\bm{x}|\bm{z}) = \mathcal{N}(\bm{x}; \bm{0}, \mathbf{I})$ to a Gaussian distribution with mean $\bm{x}_1$ and covariance $\sigma^2 \mathbf{I}$, namely $p_{t=1}(\bm{x}|\bm{z}) = \mathcal{N}(\bm{x}; \bm{x}_1, \sigma^2 \mathbf{I})$ for any target point $\bm{x}_1$. By letting $\sigma \rightarrow 0$, the marginal boundary conditions can easily be validated.
\citet{tong2024improving} further generalized the application scope to arbitrary source distributions by setting: 
\begin{align}
\label{eq: random coupling apdx}
    q(\bm{z}) = q(\bm{x}_0) q(\bm{x}_1), \quad \mu_t(\bm{z}) = (1-t) \bm{x}_0 + t \bm{x}_1, \quad \sigma_t(\bm{z}) = \sigma.    
\end{align}
The above choice satisfies the boundary conditions $p_{t=0}(\bm{x}) = q_0$ and $p_{t=1}(\bm{x}) = q_1$ when $\sigma \rightarrow 0$. 
Based on Equation \ref{eq: general Gaussian construction apdx}, the conditional velocity vector field has an analytical form $u_t(\bm{x} | \bm{z}) = \bm{x}_1 - \bm{x}_0$. 


\shortsection{Conditional Flow Matching with OT-Couplings}
The above construction in Equation \ref{eq: random coupling apdx} corresponds to the simplest choice of \emph{independent coupling}, where $\bm{z}=(\bm{x}_0, \bm{x}_1)$ with source $\bm{x}_0$ and target $\bm{x}_1$ are independently sampled from $q(\bm{z}) = q(\bm{x}_0) q(\bm{x}_1)$. The use of couplings to construct sampling paths in the CFM framework naturally connects to optimal transport theory~\citep{villani2008optimal}. Choosing OT-based couplings has several advantages over independent coupling, including smaller training variance and more efficient sampling~\citep{pooladian2023multisample, tong2024improving}.

The most classical definition of the \emph{optimal transport} (OT) problem seeks a joint coupling to move a probability measure to another that minimizes the Euclidean distance cost, corresponding to the following minimization problem with respect to the Wasserstein-2 distance:
\begin{align}
\label{eq: Kantorovich's formulation}
     \Pi^*_{\mathrm{OT}} = {\arg\!\min}_{\pi \in \Pi (q_0, q_1)} \int_{\mathbb{R}^d \times \mathbb{R}^d} \|\bm{x}_0 - \bm{x}_1\|_2^2 \: d\pi(\bm{x}_0, \bm{x}_1),
\end{align}
where $\Pi(q_0, q_1)$ denotes the set of joint probability measures such that the left and right marginals are $q_0$ and $q_1$. The above formulation is also known as Kantorovich's formulation~\citep{peyre2019computational}.
Equation \ref{eq: Kantorovich's formulation} can be solved in a mini-batch fashion using standard solvers such as POT~\citep{flamary2021pot}; however, the computational complexity is cubic in batch size.

A more efficient alternative is \emph{entropic optimal transport} (EOT), which approximately solves the OT problem using entropic regularization, reducing the computational costs from cubic to quadratic:
\begin{equation}
\label{eq: EOT formulation apdx}
\begin{aligned}
    \Pi^*_{\mathrm{EOT}}(\epsilon) = \argmin_{\pi \in \Pi (q_0, q_1)} \int_{\mathbb{R}^d \times \mathbb{R}^d} \|\bm{x}_0 - \bm{x}_1\|_2^2 \: d\pi(\bm{x}_0, \bm{x}_1) + \epsilon H(\pi \: | \: q_0 \otimes q_1),
\end{aligned}
\end{equation}
where $\epsilon > 0$ is the regularization parameter, and $H(\pi \: | \: q_0 \otimes q_1)$ denotes the relative entropy (or KL divergence) with respect to $\pi$ and the product measure $q_0 \otimes q_1$.
Note that this optimization problem can also be viewed as a special case of the static Schrödinger bridge problem~\citep{bernton2022entropic},
which can be efficiently solved in a mini-batch fashion via the Sinkhorn algorithm~\citep{1_Cuturi_Sinkhorn_fast}. 
Theoretically, one can prove that $\Pi^*_{\mathrm{EOT}}(\epsilon)$ recovers the Kantorovich's OT coupling $\Pi^*_{\mathrm{OT}}$ when $\epsilon \rightarrow 0$ and $\Pi^*_{\mathrm{EOT}}(\epsilon)$ corresponds to the independent coupling $q_0 \otimes q_1$ when $\epsilon \rightarrow \infty$.

\section{Effect of Normalization in PACM Optimization}
\label{apdx:pacm}

We elaborate on why normalizing the cost matrix in the EOT optimization objective results in suboptimal couplings, as discussed in Section \ref{sec: contextflow} when motivating our PAER design.
According to \cite{peyre2019computational}, we know that optimal EOT coupling takes the form $\Pi^{*}_{\text{EOT}}= \text{diag}(u) \cdot K \cdot \text{diag}(v)$, where $K$ is the kernel matrix such that $[K]_{ij}= \exp(-c_{ij} / \epsilon)$, and $u,v$ satisfies marginalization constraints $u \odot Kv= a$ and $K^{T}u \odot v= b$. Thus, the Sinkhorn updates are given by:
\begin{align*}
    u^{l+1}= \frac{a}{Kv^{l}} \quad \text{and} \quad v^{l+1}= \frac{b}{K^{T}u^{l+1}}.
\end{align*}
In cases where the OT cost function involves information from different modalities, the distances are usually normalized to a similar scale. Normalizing the cost results $\tilde{c}_{ij} = c_{ij} / {\epsilon}$ such that the new kernel matrix $[K_{\text{norm}}]_{ij}= \exp(\frac{-c_{ij}}{C_{\text{max}}\epsilon})$ can cause numerical issues if $C_{\text{max}} \gg 1$. Therefore, cost normalization should be performed carefully when considering different pairwise distances. Scaling the cost has the same effect as that of increasing $\epsilon$, making solutions more diffuse.
The following theorem, proven in Appendix \ref{apdx:proof of thm relative change norm cost}, formalizes how normalization affects the optimal transport plan.


\begin{theorem}
\label{thm:relative_change_norm_cost}
Let $\mathbf{C}\in \mathbb{R}^{n_0 \times n_1}$ be a cost matrix and $\mathbf{M}\in\mathbb{R}^{n_0 \times n_1}$ a prior transition matrix with positive entries.  
Consider the following optimization problem with respect to entropy-regularized OT:
\begin{align*}
    \Pi^* &= {\arg\!\min}_{\Pi}
        \sum_{k,l} \Pi_{kl} C_{kl}
        + \epsilon \sum_{k,l} \Pi_{kl} (\log\!\left(\Pi_{kl}\right)-1).
\end{align*}
Let $\tilde{\Pi}^{*}$ be the EOT coupling where the cost is scaled by a normalization constant $c$ (i.e., $\tilde{C}_{ij} = \frac{C_{ij}}{c}$) with the same regularization parameter $\epsilon>0$. Then, for any $(i,j)$ and $(k,l)$, we have
\[
\frac{\tilde{\Pi}^*_{ij}}{\tilde{\Pi}^*_{kl}}
\leq \gamma\,
\left(\frac{\Pi^*_{ij}}{\Pi^*_{kl}}\right)^{\frac{1}{c}} \quad \text{and} \quad H\big( \tilde{\Pi}_{ij}^* \big) \geq m \cdot H(\Pi_{ij}^*) - s,
\]
where $\gamma$, $m$ and $s$ are constants depending on $\Pi^{*}_{ij}$, the normalization scaler $c$, and the parameters $a, b$ in the marginalization constraints, $H(\cdot)$ stands for the entropy function.
\end{theorem}

Suppose $\Pi^{*}_{ij} = m \cdot \Pi^{*}_{kl}$ with $m>1$. Then, from Theorem \ref{thm:relative_change_norm_cost} we know $\tilde{\Pi}^{*}_{ij} < m \cdot \tilde{\Pi}^{*}_{kl}$ for any $c>1$.  This suggests that normalization brings probabilities that are far apart under the optimal transport plan closer together, leading to more diffuse couplings and increased entropy.

\section{Technical Proofs of Our Theoretical Results}
\label{apdx: proofs}


\subsection{Proof of Theorem \ref{thm: PAER Sinkhorn}}
\label{apdx:proof of theorem PAER sinkhorn}

\begin{proof}
We have that:
\begin{align*}
    \Pi^*_{\mathrm{CTF-H}} = \argmin_{\Pi \in \mathbb{R}^{n_0 \times n_1}} &\sum_{k, l} \Pi_{kl} C_{kl} + \epsilon \sum_{k,l} \Pi_{kl}  (\log (\Pi_{kl} / {M}_{kl})-1),
\end{align*}
subject to:
\[
\Pi\mathbf{1} = a, \quad \Pi^\top \mathbf{1} = b.
\]
This formulation is a standard convex optimization setting with constraints. The Lagrangian of this setting is:
\begin{align*}
\mathcal{L}(\Pi, f, g) = \sum_{k,l} C_{kl} \Pi_{kl} &+ \epsilon \sum_{k,l} \Pi_{kl} \left(\log\left(\frac{\Pi_{kl}}{M_{kl}}\right) - 1\right)
\\
&\qquad - \sum_k f_k \left(\sum_l \Pi_{kl} - a_k\right)
- \sum_l g_l \left(\sum_k \Pi_{kl} - b_l\right).
\end{align*}
Differentiating with respect to \(\Pi_{kl}, f_k, g_l\), we get:
\[
\frac{\partial \mathcal{L}}{\partial \Pi_{kl}} = C_{kl} + \epsilon \log\left(\frac{\Pi_{kl}}{M_{kl}}\right) - f_k - g_l.
\]
Setting the derivative to zero:
\[
\epsilon \log\left(\frac{\Pi^{*}_{kl}}{M_{kl}}\right) = f_k - C_{kl} + g_l
\]

\[
\implies \frac{\Pi^{*}_{kl}}{M_{kl}} = e^{\frac{f_k}{\epsilon}} e^{-\frac{C_{kl}}{\epsilon}} e^{\frac{g_l}{\epsilon}}
\]

\[
\implies \Pi^{*}_{kl} = e^{\frac{f_k}{\epsilon}} M_{kl} e^{-\frac{C_{kl}}{\epsilon}} e^{\frac{g_l}{\epsilon}}
\]
Let \(u \in \mathbb{R}^n\) and \(v \in \mathbb{R}^m\) such that
\[
u_k = e^{\frac{f_k}{\epsilon}}, \quad v_l = e^{\frac{g_l}{\epsilon}}.
\]
Let \(K_{kl}\) be the kernel \(M_{kl} e^{-C_{kl}/\epsilon}\). Then, we have:
\[
\Pi^{*}_{kl} = u_k K_{kl} v_l,
\]
suggesting that
\begin{equation}
\label{eq: prop_3_eq2}
\Pi^{*} = \operatorname{diag}(u) \cdot K \cdot \operatorname{diag}(v).
\end{equation}
Differentiating the Lagrangian with respect to \(f_k\) and \(g_l\), we get:
\[
\frac{\partial \mathcal{L}}{\partial f_k} = 1 \cdot \left(\sum_l \Pi^{*}_{kl} - a_k\right) = 0
\]

\begin{equation}
\label{eq: prop_3_eq3}
    \implies \Pi^{*} \mathbf{1} = a
\end{equation}

\[
\frac{\partial \mathcal{L}}{\partial g_l} = 1 \cdot \left(\sum_i \Pi^{*}_{kl} - b_l\right) = 0
\]

\begin{equation}
\label{eq: prop_3_eq4}
\implies \Pi^{*\top} \mathbf{1} = b
\end{equation}
From Equations~\ref{eq: prop_3_eq2}, \ref{eq: prop_3_eq3} and \ref{eq: prop_3_eq4} above, we get:
\begin{align*}
\operatorname{diag}(u) \cdot K \cdot \operatorname{diag}(v) \cdot \mathbf{1} &= a, \\
(\operatorname{diag}(u) \cdot K \cdot \operatorname{diag}(v))^\top \mathbf{1} &= b,
\end{align*}
which can be rewritten as: 
\begin{align*}
u \odot (Kv) &= a \\
K^\top u \odot v & = b.
\end{align*}
This is the usual matrix scaling formulation for which the Iterative Proportional Fitting (IPF) updates are:
\[
u_k^{t+1} = \frac{a_k}{(Kv^t)_k}, \quad v_l^{t+1} = \frac{b_l}{(K^\top u^{t+1})_l}.
\]
The Sinkhorn Algorithm uses these updates iteratively, and these updates are shown to converge in \cite{Franklin_sinkhorn_convergence}. Thus, the Sinkhorn Algorithm can be used for the ContextFlow's Prior Aware Entropy Regularized (\textit{PAER}) (CTF-H) formulation.
From Equation \ref{eq: prop_3_eq2}, we get:
\[
\Pi^{*}_{kl} = e^{f_k/\epsilon} \cdot M_{kl} \cdot e^{-C_{kl}/\epsilon} e^{g_l/\epsilon}.
\]
When \(\epsilon \to \infty\), we have \(C_{kl}/\epsilon \to 0\).
\[
 \quad e^{-C_{kl}/\epsilon} \to 1
\]
\begin{align*}
\implies & \Pi^{*}_{kl} \to u_{k}M_{kl} v_{l} \\
\implies & \Pi^{*}_{\text{CTF-H}} \to \mathrm{diag}(\bm{u}) \cdot \mathbf{M} \cdot \mathrm{diag}(\bm{v})
\end{align*}
Therefore, both marginal constraints, $\Pi^{*}_{\text{CTF-H}}\mathbf{1} = a$ and $\Pi^{*\top}_{\text{CTF-H}} \mathbf{1} = b$, are satisfied.
\end{proof}

\subsection{Proof of Theorem \ref{thm:relative_change_norm_cost}}
\label{apdx:proof of thm relative change norm cost}

To derive Theorem \ref{thm:relative_change_norm_cost}, we make use of the following lemma, which is proven in Appendix \ref{apdx:proof lemma D.1 for proving theorem c.1}.

\begin{lemma}
\label{lemma:auxiliary lemma for proving theorem c.1}
Let $\mathbf{C}\in \mathbb{R}^{n_0 \times n_1}$ be a cost matrix and $\mathbf{M}\in\mathbb{R}^{n_0 \times n_1}$ a prior transition matrix with positive entries.  
Consider the entropy-regularized OT formulation:
\begin{align*}
    \Pi^* &= {\arg\!\min}_{\Pi \ge 0}\;
        \sum_{k,l} \Pi_{kl} C_{kl}
        + \epsilon \sum_{k,l} \Pi_{kl} (\log\!\left(\Pi_{kl}\right)-1).
\end{align*}
Let $\tilde{\Pi}^{*}$ be the EOT-coupling in the case when cost is scaled by a normalization constant $c$ or $\tilde{C}_{ij} = \frac{C_{ij}}{c}$. Let the regularization parameter $\epsilon>0$ be the same in both cases. Consider the scaling factors \(\alpha, \beta\) such that: $\tilde{u}_i = \alpha_i u_i^{1/c}$, $\tilde{v}_j = \beta_j v_j^{1/c}$ where \(u, v\) are the Sinkhorn algorithm converged vectors for the original setting and \(\tilde{u}, \tilde{v}\) are for the cost-scaled version. Then, we have
\[
\max\{\left\|\phi\right\|_{\infty}, \left\|\psi\right\|_{\infty}\}
\leq \|M^{-1}\|_{\infty} \cdot \left\|
\begin{pmatrix}
\Delta_{a} \\
\Delta_{b}
\end{pmatrix}
\right\| _{\infty},
\]
where $\phi= \log(\alpha)$ and $\psi= \log(\beta)$. We also have that, 
\[
\max_i |\alpha_i-1|, \: \max_i |\beta_i-1| \leq \|M^{-1}\|_\infty \cdot \max(\|\Delta_a\|_\infty, \|\Delta_b\|_\infty),
\]
where \(M, \Delta_a, \Delta_b\) depend on $\Pi^{*}$, marginalization constants $a, b$  and normalization constant $c$.
\end{lemma}




Now, we are ready to prove Theorem \ref{thm:relative_change_norm_cost}.

\begin{proof}[Proof of Theorem \ref{thm:relative_change_norm_cost}]
For the original optimal transport (OT) formulation, we note:
\[
\Pi^{*}_{ij} = u_i K_{ij} v_j, \quad K_{ij} = e^{-C_{ij}/\epsilon},
\]
with the constraints \(\Pi^{*}\mathbf{1} = a\) and \(\Pi^{*\top} \mathbf{1} = b\).

Let
\[
\Pi_{ij}^{*1/c} = u_i^{1/c} K_{ij}^{1/c} v_j^{1/c},
\]
where:
\[
\tilde{K}_{ij} = K_{ij}^{1/c} = \exp\left(-C_{ij}/(c\epsilon)\right)
\]
is the kernel for the scaled/normalized OT formulation. Let \(\tilde{\Pi}^{*}_{ij}\) be the coupling for the scaled version, then:
\[
\tilde{\Pi}^{*}_{ij} = \tilde{u}_i \tilde{K}_{ij} \tilde{v}_j.
\]

Thus, there exist scaling factors \(\alpha_{i}, \beta_{j} \in \mathbb{R}\) such that:
\begin{align*}
\tilde{u}_{i} = \alpha_{i}u^{\frac{1}{c}}_{i}, \quad \tilde{v}_{j} = \beta_{j}v^{\frac{1}{c}}_{j}.
\end{align*}
This implies:
\[
\tilde{\Pi}^{*}_{ij} = (\alpha_i u_i^{1/c}) \tilde{K}_{ij} (\beta_j v_j^{1/c}),
\]
\begin{align}
\label{eq:g1}
\implies \tilde{\Pi}^{*} = \operatorname{diag}(\alpha u^{1/c}) \tilde{K} \operatorname{diag}(\beta v^{1/c}), 
\end{align}
\[
\implies \tilde{\Pi}^{*} = \operatorname{diag}(\alpha) \Pi^{1/c} \operatorname{diag}(\beta),
\]
subject to the constraints:
\[
\sum_i \alpha_i \beta_j \Pi_{ij}^{*1/c} = a_i, \quad \sum_i \alpha_i \beta_j \Pi_{ij}^{*1/c} = b_j.
\]
For any pair \((i, j) \& (k, l)\), we can express:
\[
\frac{\tilde{\Pi}^{*}_{ij}}{\tilde{\Pi}^{*}_{kl}} = \frac{\alpha_i}{\alpha_k} \frac{\beta_j}{\beta_l} \left(\frac{\Pi^{*}_{ij}}{\Pi^{*}_{kl}}\right)^{1/c}.
\]
Taking logarithms on both sides, we have:
\[
\log\left(\frac{\tilde{\Pi}^{*}_{ij}}{\tilde{\Pi}^{*}_{kl}}\right) = \log(\alpha_i) - \log(\alpha_k) + \log(\beta_j) - \log(\beta_l) + \frac{1}{c} \log\left(\frac{\Pi^{*}_{ij}}{\Pi^{*}_{kl}}\right).
\]

Let \(\log(\alpha) = \phi\) and \(\log(\beta) = \psi\), then:
\[
\log\left(\frac{\tilde{\Pi}^{*}_{ij}}{\tilde{\Pi}^{*}_{kl}}\right) = (\phi_i - \phi_k) + (\psi_j - \psi_l) + \frac{1}{c} \log\left(\frac{\Pi^{*}_{ij}}{\Pi^{*}_{kl}}\right).
\]
This implies:
\[
\left| \log\left(\frac{\tilde{\Pi}^{*}_{ij}}{\tilde{\Pi}^{*}_{kl}}\right) - \frac{1}{c} \log\left(\frac{\Pi^{*}_{ij}}{\Pi^{*}_{kl}}\right) \right| \leq |\phi_i| + |\phi_k| + |\psi_j| + |\psi_l|.
\]
According to Lemma \ref{lemma:auxiliary lemma for proving theorem c.1}, we have:
\[
\max_i \phi_i \leq E, \quad \max_i \psi_i \leq E.
\]

Thus:
\[
\left| \log\left(\frac{\tilde{\Pi}^{*}_{ij}}{\tilde{\Pi}^{*}_{kl}}\right) - \frac{1}{c} \log\left(\frac{\Pi^{*}_{ij}}{\Pi^{*}_{kl}}\right) \right| \leq 4E.
\]

Therefore:
\[
-4E + \frac{1}{c} \log\left(\frac{\Pi^{*}_{ij}}{\Pi^{*}_{kl}}\right) \leq \log\left(\frac{\tilde{\Pi}^{*}_{ij}}{\tilde{\Pi}^{*}_{kl}}\right) \leq 4E + \frac{1}{c} \log\left(\frac{\Pi^{*}_{ij}}{\Pi^{*}_{kl}}\right).
\]

This implies:
\[
\frac{\tilde{\Pi}^{*}_{ij}}{\tilde{\Pi}^{*}_{kl}} \leq \exp(4E) \left(\frac{\Pi^{*}_{ij}}{\Pi^{*}_{kl}}\right)^{1/c}.
\]

Let \(\gamma = \exp(4E)\), then:
\[
\frac{\tilde{\Pi}^{*}_{ij}}{\tilde{\Pi}^{*}_{kl}} \leq \gamma \left(\frac{\Pi^{*}_{ij}}{\Pi^{*}_{kl}}\right)^{1/c}.
\]
Now the remaining task is to prove the second inequality.
From Equation \ref{eq:g1}, we know that: 
\[
\tilde{\Pi}^{*}_{ij} = (\Pi^{*}_{ij})^{1/c} \cdot \exp(\phi_i, \psi_j),
\]
and from Lemma \ref{lemma:auxiliary lemma for proving theorem c.1}, we have
\[
\tilde{\Pi}^{*}_{ij} \leq (\Pi^{*}_{ij})^{1/c} \cdot e^{2E}
\]

\[
\Rightarrow \log(\tilde{\Pi}^{*}_{ij}) \leq \frac{1}{c} \log(\Pi^{*}_{ij}) + 2E
\]

\[
\Rightarrow -\tilde{\Pi}^{*}_{ij} \log(\tilde{\Pi}^{*}_{ij}) \geq -\frac{1}{c} (\Pi^{*}_{ij})^{1/c - 1} \cdot \Pi^{*}_{ij} \log(\Pi^{*}_{ij}) \cdot e^{2E} - 2E \cdot e^{2E} \cdot (\Pi^{*}_{ij})^{1/c}
\]

For \(c \gg 1\), \(\frac{1}{c} \to 0\):
\begin{align*}
\Rightarrow -\tilde{\Pi}^{*}_{ij} \log(\tilde{\Pi}^{*}_{ij}) & \geq -\frac{1}{c\Pi^{*}_{ij}} \cdot \Pi^{*}_{ij} \log(\Pi^{*}_{ij}) \cdot e^{2E} - 2E \cdot e^{2E} \cdot (\Pi^{*}_{ij})^{1/c} \\
\Rightarrow -\tilde{\Pi}^{*}_{ij} \log(\tilde{\Pi}^{*}_{ij}) & \geq -\frac{1}{c\Pi^{*}_{\min}} \cdot \Pi^{*}_{ij} \log(\Pi^{*}_{ij}) \cdot e^{2E} - 2E \cdot e^{2E} \cdot (\Pi^{*}_{ij})^{1/c}
\end{align*}

Summing for all $(i,j)$ we get, 

\[
H(\tilde{\Pi^{*}}) \geq m H(\Pi^{*}) - s,
\]
where $m= \frac{e^{2E}}{c\Pi^{*}_{\min}}$ and $s= 2E \cdot e^{2E}$. Thus, we complete the proof.
\end{proof}

\subsection{Proof of Lemma \ref{lemma:auxiliary lemma for proving theorem c.1}}
\label{apdx:proof lemma D.1 for proving theorem c.1}

\begin{proof}
Let \(X_{ij} = \Pi_{ij}^{*1/c}\) and \(X = \Pi^{*1/c}\). Consider the exponentiated versions of \(\alpha\) and \(\beta\):
\[
\phi = \log(\alpha) \in \mathbb{R}^n, \quad \psi = \log(\beta) \in \mathbb{R}^m.
\]

From the marginal constraints, we have:
\[
\sum_j X_{ij} e^{\phi_i + \psi_j} = a_i, \quad \sum_i X_{ij} e^{\phi_i + \psi_j} = b_j.
\]

Applying a first-order Taylor expansion gives:
\[
\sum_j X_{ij} (1 + \phi_i + \psi_j) = a_i \quad \implies \quad \sum_j X_{ij} (\phi_i + \psi_j) = a_i - \sum_j X_{ij},
\]
\[
\sum_i X_{ij} (1 + \phi_i + \psi_j) = b_j \quad \implies \quad \sum_i X_{ij} (\phi_i + \psi_j) = b_j - \sum_i X_{ij}.
\]

Define:
\[
\Delta_{a_i} = a_i - \sum_j X_{ij}, \quad \Delta_{b_j} = b_j - \sum_i X_{ij}.
\]

Thus, we have:
\[
\sum_j X_{ij} (\phi_i + \psi_j) = \Delta a_i, \quad \sum_i X_{ij} (\phi_i + \psi_j) = \Delta b_j.
\]

This implies:
\[
\phi_i \left(\sum_j X_{ij}\right) + \sum_j X_{ij} \psi_j = \Delta a_i,
\]
\[
\sum_i X_{ij} \phi_i + \psi_j \left(\sum_i X_{ij}\right) = \Delta b_j.
\]

Let:
\[
D_r = \operatorname{diag}(X \mathbf{1}) \in \mathbb{R}^{n \times n}, \quad D_c = \operatorname{diag}(X^T \mathbf{1}) \in \mathbb{R}^{m \times m}.
\]

Then we can express the system as:
\[
\begin{pmatrix}
D_r & X \\
X^T & D_c
\end{pmatrix}
\begin{pmatrix}
\phi \\
\psi
\end{pmatrix}
=
\begin{pmatrix}
\Delta a \\
\Delta b
\end{pmatrix}.
\]

Let:
\[
M = 
\begin{pmatrix}
D_r & X \\
X^T & D_c
\end{pmatrix}.
\]

Thus:
\[
\begin{pmatrix}
\phi \\
\psi
\end{pmatrix}
= M^{-1}
\begin{pmatrix}
\Delta_{a} \\
\Delta_{b}
\end{pmatrix}.
\]

This implies:
\[
\left\|
\begin{pmatrix}
\phi \\
\psi
\end{pmatrix}
\right\|
\leq \|M^{-1}\| \cdot \left\|
\begin{pmatrix}
\Delta_{a} \\
\Delta_{b}
\end{pmatrix}
\right\|.
\]

Since \(\alpha = \exp(\phi)\) and \(\beta = \exp(\psi)\), by assumption:
\[
|\alpha_i - 1| \approx |\exp(\phi_i) - 1| \approx \phi_i,
\]
\[
|\beta_j - 1| \approx |\exp(\psi_j) - 1| \approx \psi_j.
\]

Therefore, we obtain
\[
\max_i |\alpha_i - 1|, \max_j |\beta_j - 1| \leq \|M^{-1}\|_{\infty} \cdot \max(\|\Delta a\|_{\infty}, \|\Delta b\|_{\infty}),
\]
which completes the proof.
\end{proof}





\section{Detailed Analysis of ContextFlow}
\label{apdx: contextflow}

\subsection{Algorithm Pseudocode}
\label{apdx: alg pseudocode}

\begin{algorithm}[H]
\caption{ContextFlow: Flow Matching with Spatial-Context-Aware OT Couplings}
\setstretch{1.1}
\begin{algorithmic}[1]
\State \textbf{Input:} gene data $\{\mathbf{X}_{t_1}, \cdots, \mathbf{X}_{t_{m+1}}\}$, spatial data $\{\mathbf{S}_{t_1}, \ldots, \mathbf{S}_{t_{m+1}}\}$, parameters $\lambda$, $\alpha$, $\epsilon$, $\sigma$, $r$, $\eta$ 
\State \textbf{Data-Preprocessing:} Compute local neighborhood means using Nearest Neighbor Algorithm and Ligand-Receptor features $f_{\text{LR}}$ using LIANA+ \Comment{As defined in Equation \ref{eq: spatial smoothness prior} and \ref{eq: LR prior} }
\State \textbf{Output:} neural velocity vector field $u_\theta$
\State Initialize $\theta$
\While{training}
    \For{$i=1, 2, \ldots, m$}
        \State Sample a batch $\mathcal{B} = \{(\bm{x}_i, \bm{x}_{i+1}): (\bm{x}_i, \bm{x}_{i+1}) \sim  (\mathbf{X}_{t_i}, \mathbf{X}_{t_{i+1}})\}$
        \State Construct TPM: $\mathbf{M}_{i, i+1}(\mathcal{B})$\Comment{$\mathbf{M}_{i, i+1}$ is defined in Equation \ref{eq: TPM}}
        \If{``prior-aware cost matrix''}
            \State $C_{kl} \leftarrow \alpha \cdot \| \bm{x}_{i}(k) - \bm{x}_{i+1}(l)\|_2^2 + (1-\alpha) \cdot [\mathbf{M}_{i, i+1}]_{kl}$ for any pair $(k,l)$
            \State $\mathbf{K} \leftarrow \exp(-\mathbf{C} / \epsilon)$
        \ElsIf{``prior-aware entropy regularization''}
            \State $C_{kl} \leftarrow \| \bm{x}_{i}(k) - \bm{x}_{i+1}(l)\|_2^2$ for any pair $(k,l)$
            \State $\mathbf{K} \leftarrow \widehat{\mathbf{M}}_{i, i+1} \odot \exp(-\mathbf{C} / \epsilon)$ \Comment{$\widehat{\mathbf{M}}_{i, i+1}$ is defined in Equation \ref{eq: PAER}}
        \EndIf
        \State Initialize $\bm{a} \gets \frac{1}{n_i}\mathbf{1}_{n_i}$, $\bm{b} \gets \frac{1}{n_{i+1}}\mathbf{1}_{n_{i+1}}$, $\bm{u} \gets \mathbf{1}_{n_i}$, $\bm{v} \gets \mathbf{1}_{n_{i+1}}$
        \While{not converged}
            \State $\bm{u} \gets \bm{a} \oslash (\mathbf{K} \bm{v}), \: \bm{v} \gets \bm{b} \oslash (\mathbf{K}^{\top} \bm{u})$ \Comment{Run Sinkhorn algorithm}
            \EndWhile
            \State Obtain spatial-prior-aware OT couplings $\mathbf{\Pi}^{\mathrm{CTF}}_{i, i+1} \leftarrow \operatorname{diag}(\bm{u}) \mathbf{K} \operatorname{diag}(\bm{v})$
    
            \State Sample $t \sim \mathcal{U}(t_i, t_{i+1})$ and $\{(\bm{x}_i, \bm{x}_{i+1}): (\bm{x}_i, \bm{x}_{i+1}) \sim \mathbf{\Pi}^{\mathrm{CTF}}_{i, i+1}\}$ 
            \State  Sample $\bm{x}_t \sim \mathcal{N}\left( \frac{t_{i+1} - t}{t_{i+1} - t_i} \bm{x}_i + \frac{t - t_i}{t_{i+1} - t_i} \bm{x}_{i+1}, \sigma^2 \mathbf{I}  \right)$ 
            \State $L_{\mathrm{CFM}} \leftarrow \frac{1}{|\mathcal{B}|}\sum_{t, (\bm{x}_i, \bm{x}_{i+1})}\left \| u_\theta ( \bm{x}_t, t) - \frac{\bm{x}_{i+1} - \bm{x}_{i}}{t_{i+1} - t_i} \right\|_2^2$ 
        \EndFor
        \State $\theta \leftarrow \theta - \eta \cdot \nabla_{\theta} L_{\mathrm{CFM}}$
\EndWhile
\end{algorithmic}
\label{alg: contextflow}
\end{algorithm}

\subsection{Training Details and Hyperparameters}
 \label{apdx: training_details}
Similar neural network architectures and training settings
are used for all the evaluated baselines and ContextFlow.
Specifically, we list the training hyperparameters used in our experiments: 
\begin{itemize}
    \item Learning rate: $0.0005$
    \item Learning rate scheduler: Cosine
    \item Epochs: $10000$ 
    \item NN: MLP
    \item Depth: $4$
    \item Width: $100$
    \item Radius of SS prior: $50$
    \item LIANA+~\citep{14_Dimitrov_Liana} settings for LR prior:
    \begin{itemize}
        \item Resource: Consensus
        \item nzprop: $0.01$
        \item Similarity Metric: Cosine
    \end{itemize}
    \item Hyperparameter $\epsilon$ used in EOT formulation (Equation \ref{eq: EOT formulation}):
    \begin{itemize}
        \item $0.1$ for Brain Regeneration experiments
        \item $10$ for Organogenesis and Liver regeneration
    \end{itemize}
\end{itemize}
 
\subsection{Time Complexity Analysis}
\label{apdx:complexity_analysis}

\begin{figure}[t]
\centering
\includegraphics[width=0.8\linewidth]{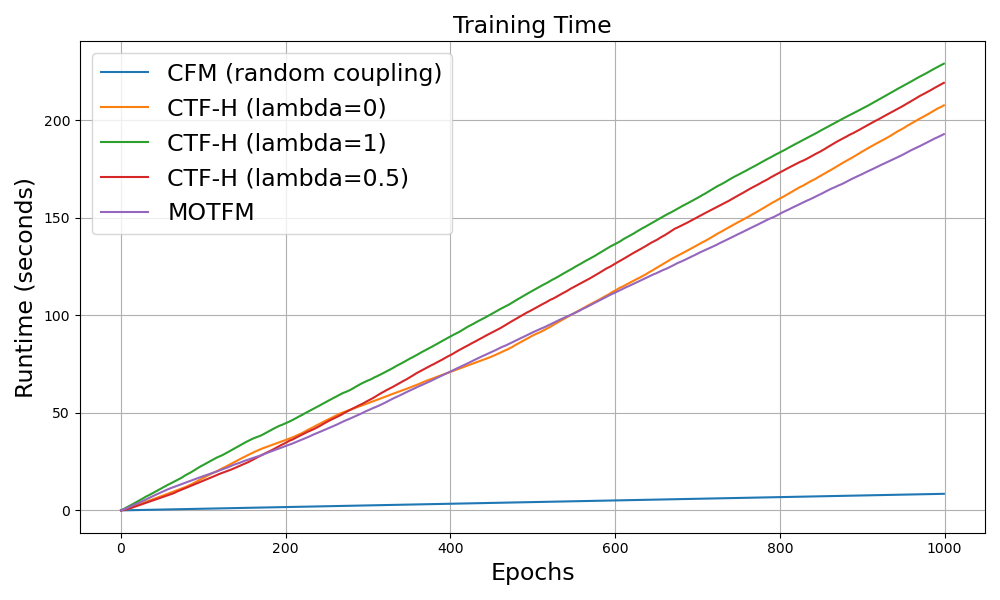}
\vspace{-0.1in}
\caption{Runtime comparisons across various algorithms on Brain Regeneration with batch size $256$.}
\label{fig:runtime_FM}
\end{figure}

The training time of ContextFlow is comparable to that of Minibatch-OT FM \citep{tong2024improving} (Figure \ref{fig:runtime_FM}), as both methods solve an entropic variant of optimal transport using the GPU-optimized Sinkhorn algorithm, alongside forward and backward propagation steps that are also GPU-accelerated. Although ContextFlow incorporates prior knowledge, such as spatial smoothness (Equation \ref{eq: spatial smoothness prior}) and cell-cell communication patterns (Equation \ref{eq: LR prior}), their corresponding features are computed once during preprocessing, resulting in a one-time cost. The precomputed features can be reused across multiple hyperparameter settings and model variants, making ContextFlow highly scalable and efficient.

\shortsection{Data Preprocessing}
The preprocessing steps generate additional biologically informed features that complement the original transcriptomic profiles. These features incur a one-time computational cost and can be reused across different experiments and model configurations.

For the SS prior, we use a nearest-neighbor (NN) algorithm to compute the mean of local transcriptomic features for each cell. The computational complexity of the NN search is known to be $O(N^{2}d)$, where $N$ denotes the total number of points considered and $d$ denotes the data dimension. 
For the LR prior, we employ spatially informed bivariate statistics implemented in LIANA+~\citep{14_Dimitrov_Liana}, using the cosine similarity metric on gene expression profiles and the recommended hyperparameters. The exact runtime complexity for LIANA+ is unknown. Table \ref{tab:runtime_LR} summarizes the total time taken for the Brain Regeneration and Mouse Organogenesis datasets.

\begin{table}[t]
\centering
\caption{Runtime for computing cell-cell communication patterns.}
\label{tab:runtime_LR}
\small
\begin{tabular}{lcc}
\toprule
\textbf{Dataset} & \textbf{Total Number of Cells} & \textbf{Runtime (seconds)} \\
\midrule
Brain Regeneration \citep{1_dataset_steroseq} & $28{,}780$ & $23.35$ \\
Mouse Organogenesis \citep{2_dataset_mosta} & $399{,}248$ & $200.40$ \\
\bottomrule
\end{tabular}
\end{table}

\shortsection{Training of ContextFlow}
The time complexity depends on the total number of training epochs ($E$), mini-batch size ($B$), time per forward and backward pass ($P$), transcriptomic feature dimension ($d$), and the total number of LR pairs ($l$). Below, we compute the time complexity for each step.

The construction of the transition plausibility matrix involves a calculation of pairwise distances for each mini-batch, resulting in the runtime of $O(B^{2}(l+d))$.
According to Theorem \ref{thm: PAER Sinkhorn}, we know that Sinkhorn iterations can be adapted to solve the prior-aware entropy regularization problem (Equation \ref{eq: PAER}). Since the Sinkhorn algorithm has a well-known quadratic time complexity \citep{1_Cuturi_Sinkhorn_fast}, the runtime for computing minibatch OT couplings in ContextFlow is $O(B^{2})$.
Putting pieces together, across all the training epochs, the total runtime complexity of ContextFlow turns out to be $O(E \times (B^{2}(d+l)+ P))$. As shown in Figure \ref{fig:runtime_FM}, the runtime is linearly dependent on the total epochs $E$, with different linear rates for various configurations. CFM is the fastest because it bypasses the optimal transport coupling step required by the other methods.

\section{Spatiotemporal Optimal Transport}
\label{apdx:static_ot}

In this section, we compare state-of-the-art spatiotemporal alignment methods, including DeST-OT~\citep{halmos2025dest}, MOSCOT~\citep{8_Klien_moscot} and TOAST~\citep{ceccarelli2026topography}, with our PAER-OT objective used in the entropic variant of ContextFlow (CTF-H). It is important to note that these OT methods are not generative models and are only used for pairwise alignment tasks. Our method, on the other hand, is a generative model that learns a dynamic flow across the time horizon and utilizes OT couplings to design better conditional paths for regression. 
In particular, we compute the metrics described in DeST-OT on the Axolotl Brain Regeneration dataset, using the same setup as in flow matching. Specifically, for each time step, we randomly sample a batch of $1000$ cells and compute the corresponding coupling matrix $\mathbf{\Pi}$, which is then used to derive the metrics. We use the CTF-H ($\lambda=0.8$) version of ContextFlow for comparison.

\begin{table}[t]
\centering
    \centering
    \caption{Comparisons in terms of growth distortion metric.}
    \label{tab:growth_distortion}
    \small
    \renewcommand{\arraystretch}{1.05}
    \begin{tabular}{c cccc}
    \toprule
    \textbf{Static Pair} & \textbf{CTF-H OT} & \textbf{DeST-OT} & \textbf{TOAST} & \textbf{MOSCOT} \\
    \midrule
    $(1,2)$ & $0.0007$ & $0.0000$ & $0.0000$ & $0.0000$ \\
    $(2,3)$ & $0.0042$ & $0.0000$ & $0.0000$ & $0.0000$ \\
    $(3,4)$ & $0.0027$ & $0.0000$ & $0.0000$ & $0.0000$ \\
    $(4,5)$ & $0.0009$ & $0.0000$ & $0.0000$ & $0.0000$ \\
    \bottomrule
    \end{tabular}
\end{table}

\begin{table}[t]
\centering
    \centering
    \caption{Comparisons in terms of migration metric.}
    \label{tab:migration_metric}
    \small
    \renewcommand{\arraystretch}{1.05}
    \begin{tabular}{c cccc}
    \toprule
    \textbf{Static Pair} & \textbf{CTF-H OT} & \textbf{DeST-OT} & \textbf{TOAST} & \textbf{MOSCOT} \\
    \midrule
    $(1,2)$ & $231.39$ & $308.96$ & $793.70$ & $232.83$ \\
    $(2,3)$ & $289.54$ & $551.52$ & $1103.28$ & $395.15$ \\
    $(3,4)$ & $2094.11$ & $1015.72$ & $2052.65$ & $950.87$ \\
    $(4,5)$ & $1763.51$ & $777.13$ & $2257.01$ & $536.65$ \\
    \bottomrule
    \end{tabular}
\end{table}


\shortsection{Growth Distortion} 
DeST-OT introduces an OT objective for aligning spatial transcriptomic tissue slices from different developmental timesteps, with an emphasis on modeling cell growth and tissue expansion/contraction. The growth distortion metric assesses whether the inferred growth pattern aligns with changes in cell-type abundance across timesteps. Table \ref{tab:growth_distortion} shows that our CTF-H OT method is competitive, despite DeST-OT being specifically designed with cell growth in mind.

\shortsection{Migration}
The migration metric introduced in DeST-OT assesses whether the coupling produces realistic cell movements between timesteps. As seen in Table \ref{tab:migration_metric}, DeST-OT and MOSCOT achieve the best performance overall, highlighting the advantage of its growth-aware objective compared to TOAST and CTF-H OT, which do not explicitly model tissue expansion or contraction. Nevertheless, CTF-H OT achieves superior performance, especially for the first two developmental stage pairs.

\shortsection{Coupled Transcriptomic Distance}
Finally, coupled transcriptomic distance computes how similar the transcriptomic values of coupled cells are, which is defined as:
$$
\sum_{k=1}^{N}\sum_{l=1}^{M}\big\|X_{t_{i}}[k,:]-X_{t_{i+1}}[l,:]\big\|^{2}\times\Pi_{i,j},
$$
where $X_{t_{i}}[k,:]$ represents the transcriptomic feature of cell $k$ from timestep $t_{i}$ and $X_{t_{i+1}}[l,:]$ represents the transcriptomic feature of cell $l$ from timestep $t_{i+1}$, and $\Pi$ is the OT coupling matrix. From Table \ref{tab:ctpi_metric}, we can observe that CTF-H OT is competitive with DeST, MOSCOT, and TOAST.

\begin{table}[t]
    \centering
    \caption{Comparisons in terms of coupled transcriptomic distance}
    \label{tab:ctpi_metric}
    \small
    \renewcommand{\arraystretch}{1.05}
    \begin{tabular}{c cccc}
    \toprule
    \textbf{Static Pair} & \textbf{CTF-H OT} & \textbf{DeST-OT} & \textbf{TOAST} & \textbf{MOSCOT} \\
    \midrule
    $(1,2)$ & $33.58$ & $32.64$ & $34.13$ & $33.07$ \\
    $(2,3)$ & $18.01$ & $14.22$ & $21.87$ & $15.63$ \\
    $(3,4)$ & $43.26$ & $42.26$ & $43.89$ & $43.19$ \\
    $(4,5)$ & $20.14$ & $18.47$ & $20.54$ & $18.40$ \\
    \bottomrule
    \end{tabular}
\end{table}

\begin{table}[t]
\centering
    \caption{Comparisons in terms of runtime (s) with varying batch size.}
    \label{tab:sample_time_complexity_ot}
    \small
    \renewcommand{\arraystretch}{1.05}
    \begin{tabular}{c cccc}
    \toprule
    \textbf{Batch Size} & \textbf{CTF-H OT} & \textbf{DeST-OT} & \textbf{MOSCOT} & \textbf{TOAST} \\
    \midrule
    $10$   & $0.0124$ & $0.0114$ & $0.5199$ & $0.0227$ \\
    $50$   & $0.0017$ & $0.0351$ & $0.8479$ & $0.0114$ \\
    $100$  & $0.0011$ & $0.1016$ & $2.3381$ & $0.0167$ \\
    $150$  & $0.0018$ & $0.1778$ & $1.1979$ & $0.0247$ \\
    $200$  & $0.0034$ & $0.3521$ & $1.5328$ & $0.0337$ \\
    $250$  & $0.0025$ & $0.3836$ & $1.9100$ & $0.0426$ \\
    $300$  & $0.0035$ & $0.5323$ & $1.0772$ & $0.0543$ \\
    $350$  & $0.0044$ & $0.5935$ & $1.2983$ & $0.0656$ \\
    $400$  & $0.0047$ & $0.6217$ & $1.1782$ & $0.0817$ \\
    $450$  & $0.0051$ & $0.7806$ & $1.2357$ & $0.0983$ \\
    $500$  & $0.0048$ & $0.9984$ & $1.2243$ & $0.1197$ \\
    $550$  & $0.0054$ & $1.3473$ & $1.4695$ & $0.1468$ \\
    $600$  & $0.0073$ & $1.6192$ & $1.6680$ & $0.1905$ \\
    $650$  & $0.0070$ & $1.7771$ & $2.2559$ & $0.2077$ \\
    $700$  & $0.0090$ & $2.2104$ & $1.9901$ & $0.2309$ \\
    $750$  & $0.0089$ & $2.9324$ & $2.7181$ & $0.2574$ \\
    $800$  & $0.0106$ & $3.2879$ & $1.9840$ & $0.3001$ \\
    $850$  & $0.0117$ & $5.0941$ & $2.3511$ & $0.3229$ \\
    $900$  & $0.0128$ & $5.6585$ & $2.4864$ & $0.3798$ \\
    $950$  & $0.0150$ & $5.3771$ & $3.0964$ & $0.4206$ \\
    $1000$ & $0.0146$ & $5.9733$ & $3.3683$ & $0.4931$ \\
    \bottomrule
    \end{tabular}
\end{table}

\shortsection{Runtime}
To complement Figure \ref{fig:ot_runtime}, we report the runtime complexity of the above-mentioned OT methods in Table \ref{tab:sample_time_complexity_ot}. CTF-H OT is the fastest of the three, followed by TOAST, MOSCOT, and DeST-OT, and is competitive in the metrics above. We also observe that DeST-OT is the slowest, as expected, since its OT objective includes a Gromov-Wasserstein term, which has $O(n^{3})$ runtime, along with other growth- and tissue-distortion-specific terms. To summarize, our evaluations above demonstrate that ContextFlow's design choices enable it to be highly scalable compared to state-of-the-art methods, while remaining competitive across several spatiotemporal OT alignment metrics.

\begin{figure}[t]
\centering
    \includegraphics[width=0.98\textwidth]{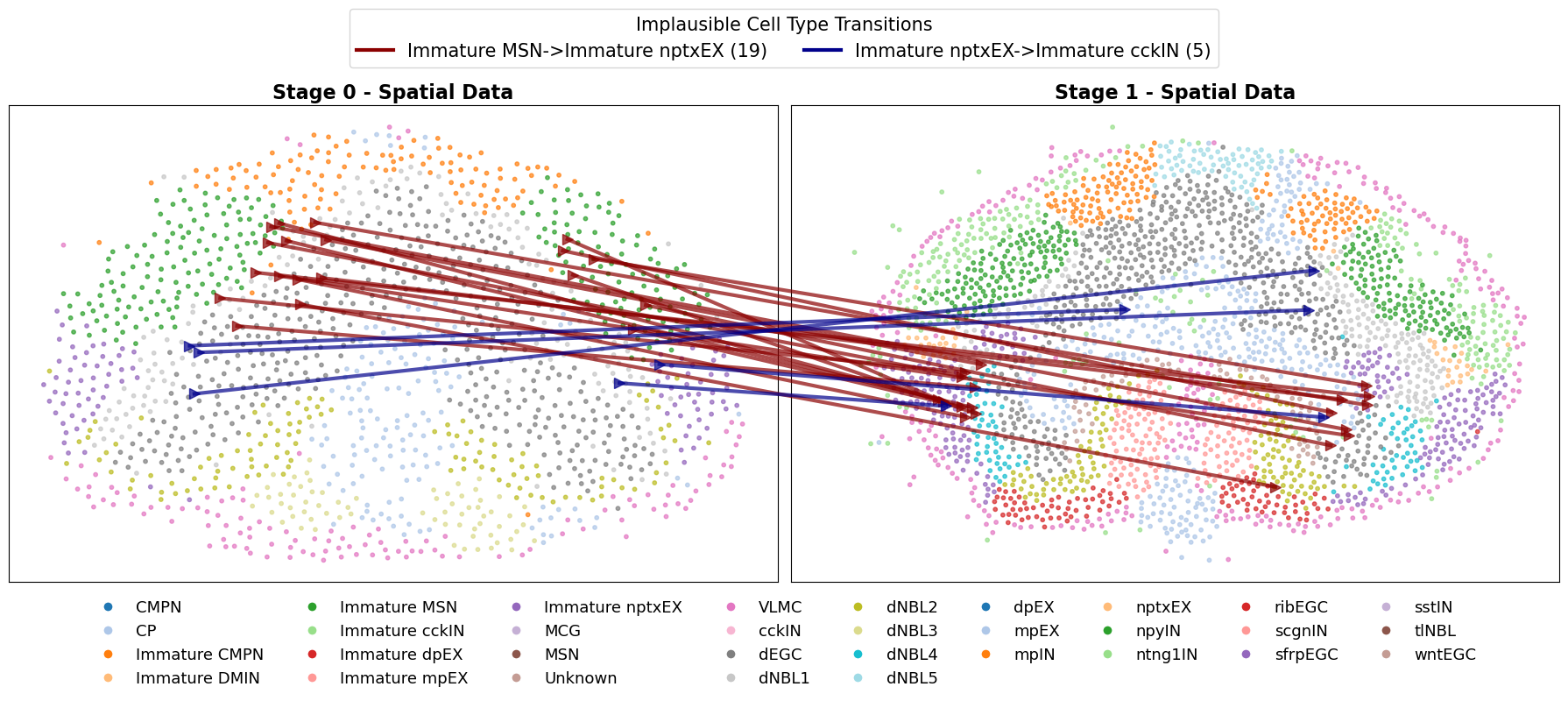}
    \label{fig:prot_implausible}
\caption{Visualization of biologically implausible cell type couplings resulted from the PAER-OT plan used in ContextFlow on the Brain Regeneration dataset, where implausibility refers to transitions involving excitatory-inhibitory lineage switches. This result is in sharp contrast to Figure \ref{fig:EOT_implausible_couplings} with \textbf{55} percent decrease in implausible couplings in this case.}
\label{fig:implausible_transitions_comparison}
\end{figure}

\section{Visualizations}
\label{apdx:vizualization}

\subsection{(Im-)Plausibility of OT-Couplings}
\label{apdx:vizualization_implausibility}

To demonstrate the need for integrating biological priors within a generative framework, we compute the EOT plan (Equation \ref{eq: EOT formulation}) for the MOTFM framework and the PAER-OT plan (Section~\ref{sec: paer}) for the ContextFlow framework. From these transport plans, we sample couplings corresponding to the first two stages of the Brain Regeneration dataset~\citep{1_dataset_steroseq} with their associated cell types. Figures~\ref{fig:EOT_implausible_couplings} and \ref{fig:implausible_transitions_comparison} illustrate the excitatory–inhibitory lineage switches present in these couplings. Since excitatory and inhibitory neurons have mutually exclusive neurotransmitter functions and originate from distinct progenitor populations with different transcription factor profiles, a transition from excitatory to inhibitory identity is considered biologically implausible.

In our transport plan couplings, we observed the following cell type lineage switches: (i) Immature MSN $\rightarrow$ Immature nptxEX, (ii) Immature MSN $\rightarrow$ Immature dpEX, (iii) Immature MSN $\rightarrow$ Immature CMPN, (iv) Immature nptxEX $\rightarrow$ Immature cckIN, and (v) Immature nptxEX $\rightarrow$ Immature MSN.
Of these, $54$ implausible transitions arise from the Entropic-OT plan, compared to $24$ from PAER-OT, with the specific transitions detailed in the figure legends. We also observed that the EOT formulation produced implausible transitions across brain hemispheres, for example, coupling cells from the left hemisphere with those from the right. In contrast, the PAER-OT formulation typically restricted transitions to within the same hemisphere, reflecting its integration of spatially aware contextual information. These observations provide strong motivation for incorporating biological priors through ContextFlow as a principled approach to learning biologically consistent developmental trajectories.




\subsection{Cell Type Distributions across Datasets}
\label{apdx:spatio_temporal_celltypes}

\begin{figure}[H]
  \centering
    \includegraphics[width=\linewidth]{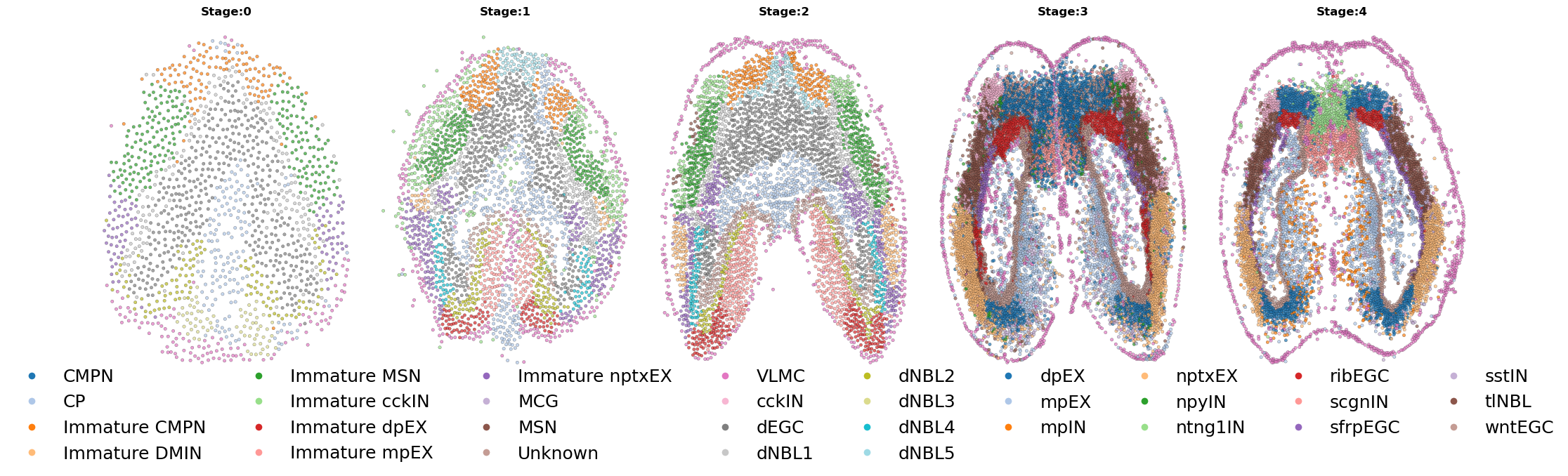} 
  \caption{Temporal progression of spatial distribution of different cell types for Brain Regeneration.}
  \label{fig:temporal_progression_stereoseq}
\end{figure}

\begin{figure}[H]
  \centering
    \includegraphics[width=\linewidth]{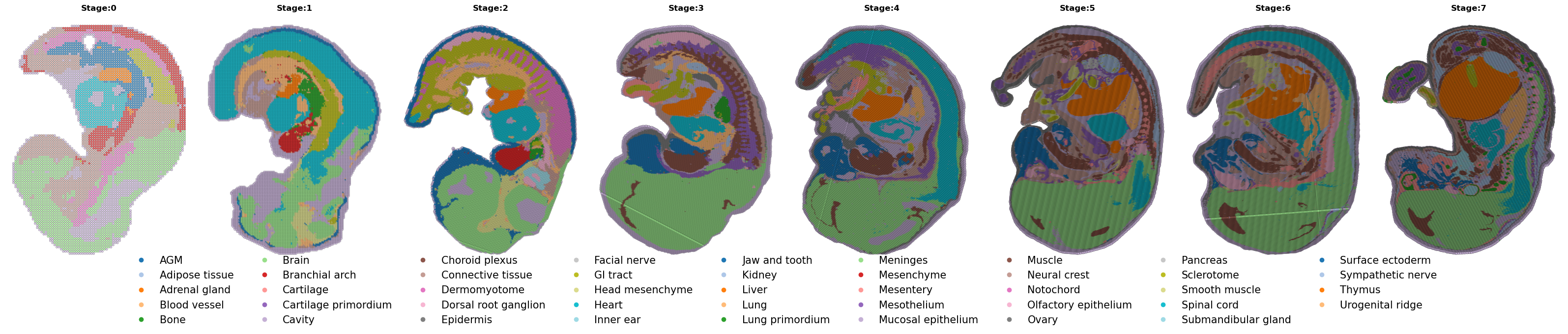} 
  \caption{Temporal progression of spatial distribution of different cell types for Mouse Organogenesis.}
  \label{fig:temporal_progression_mosta}
\end{figure}

\begin{figure}[H]
  \centering
    \includegraphics[width=0.9\linewidth]{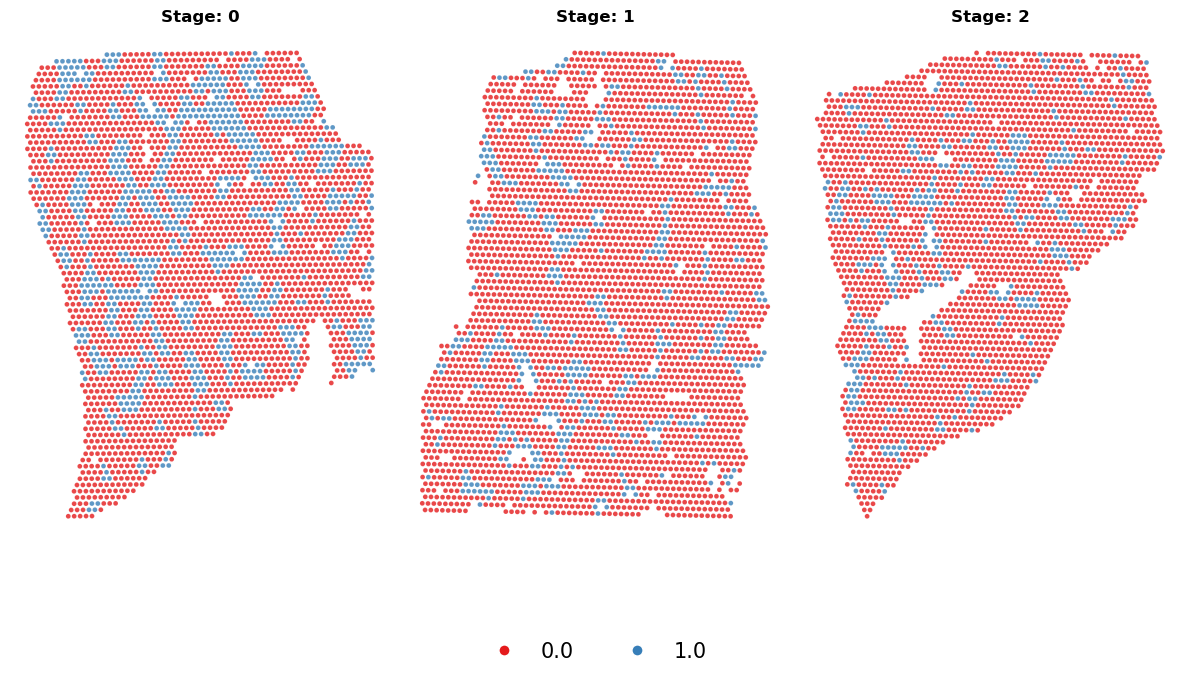} 
  \caption{Temporal progression of spatial distribution of fibrogenic states for Liver Regeneration. Here, $0/1$ refers to the absence or presence of fibrogenic spots.}
  \label{fig:temporal_progression_liver}
\end{figure}

Figures~\ref{fig:temporal_progression_stereoseq}–\ref{fig:temporal_progression_liver} present spatial maps of transcriptomic datasets across different time points, illustrating how tissue organization and cell-type distributions evolve during development and regeneration. These maps highlight not only changes in cellular composition but also the preservation of spatial neighborhoods and geometrical arrangements of specific cell types over time. Such contextual information, specific to spatial transcriptomics, remains inaccessible to standard flow-matching frameworks. By contrast, ContextFlow is designed to exploit these spatial features, enabling the inference of trajectories that are both temporally smooth and spatially coherent.

\subsubsection{Brain Regeneration}

\subsubsection{Mouse Embryo Organogenesis}

\subsubsection{Liver Regeneration}


\subsection{Ligand-Receptor Interactions}
\label{apdx:LR_patterns}

\begin{figure}[t]
  \centering
  \begin{subfigure}{0.48\textwidth}
    \centering
    \includegraphics[width=\linewidth]{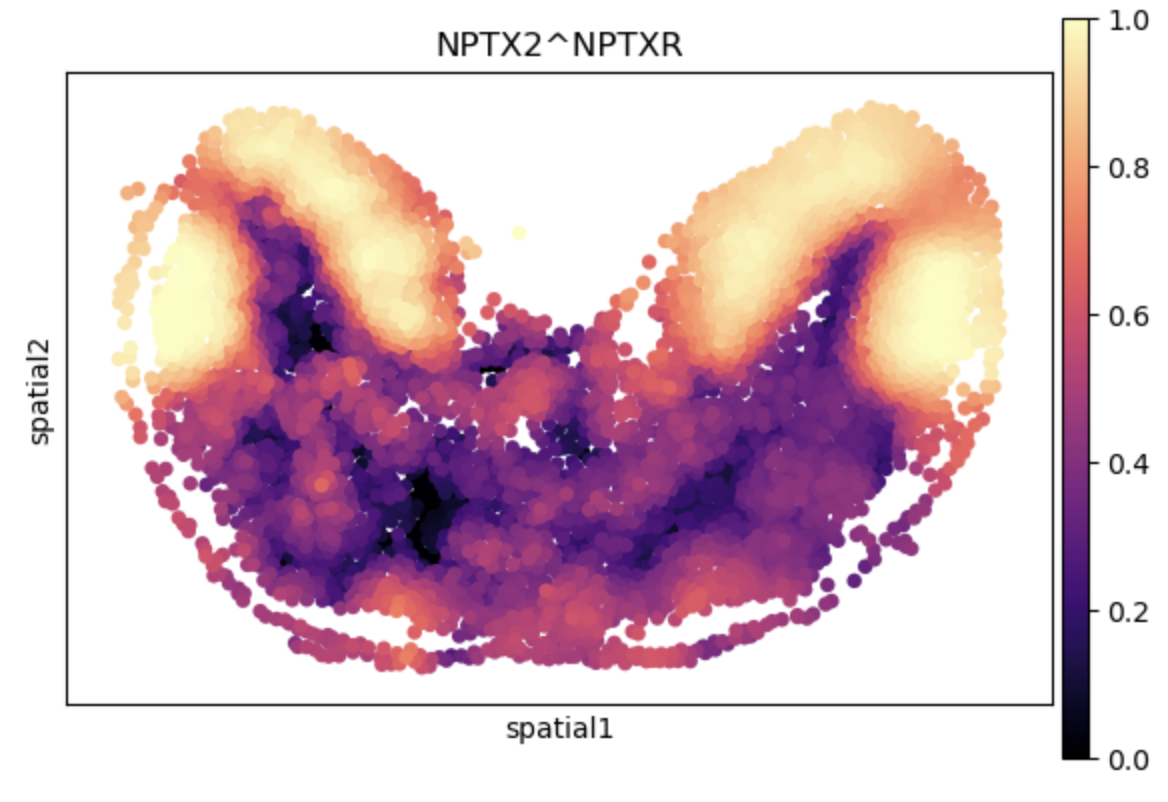} 
    \caption{NPTX2-NPTXR LR pair activation on Stage 3}
    \label{fig:one}
  \end{subfigure}\hfill
  \begin{subfigure}{0.48\textwidth}
    \centering
    \includegraphics[width=\linewidth]{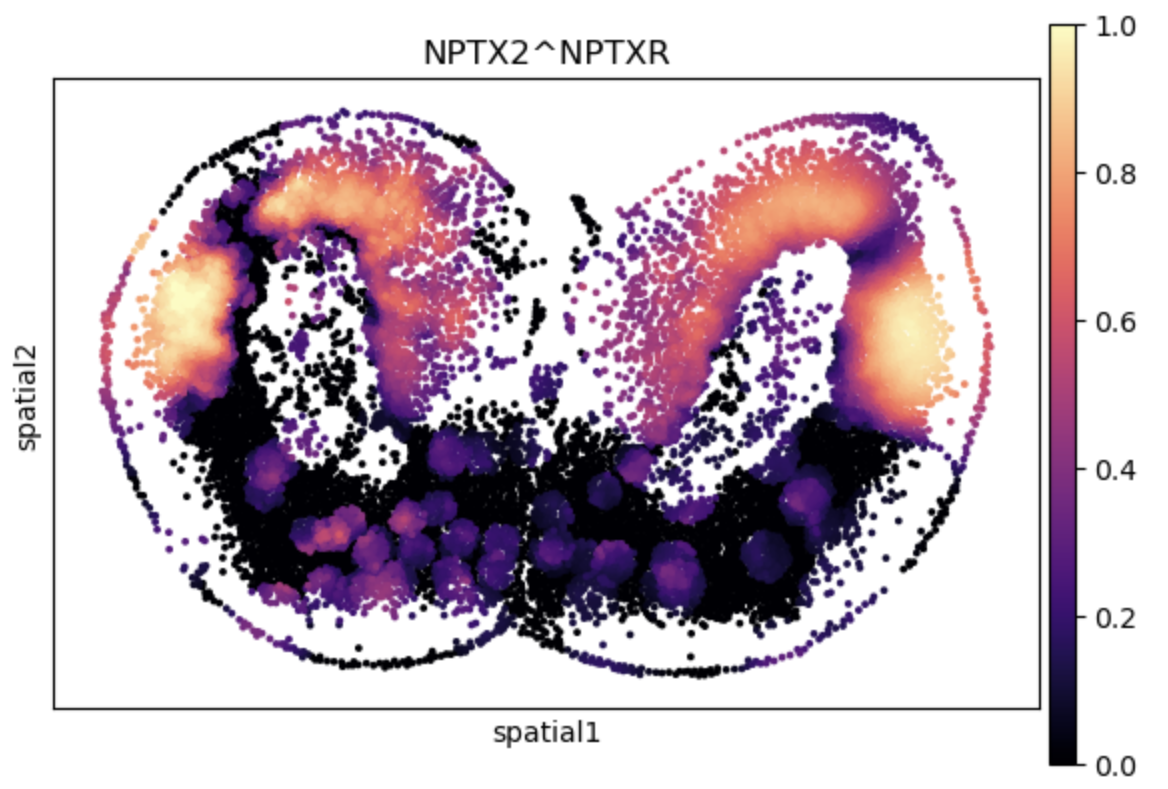} 
    \caption{NPTX2-NPTXR LR pair activation on Stage 4}
    \label{fig:two}
  \end{subfigure}
  \caption{Spatial distributions of LR activation for NPTX2-NPTXR in two consecutive slides from the Brain regeneration dataset. Similar activations are visible at structurally equal positions.}
  \label{fig:LR_viz}
\end{figure}

Figure~\ref{fig:LR_viz} illustrates the ligand-receptor score of the NPTX2-NPTXR pair in two consecutive slides from the Brain regeneration dataset. Similar activities are bilaterally visible in the cerebral cortex, suggesting that ligand–receptor interactions are preserved over time and spatially aligned with underlying tissue structure. This observation provides strong evidence that including LR interactions as contextual priors is biologically meaningful, as they capture functional communication signals between cells that remain stable across short time intervals.



\begin{figure}[t]
  \centering
    \includegraphics[width=0.98\linewidth]{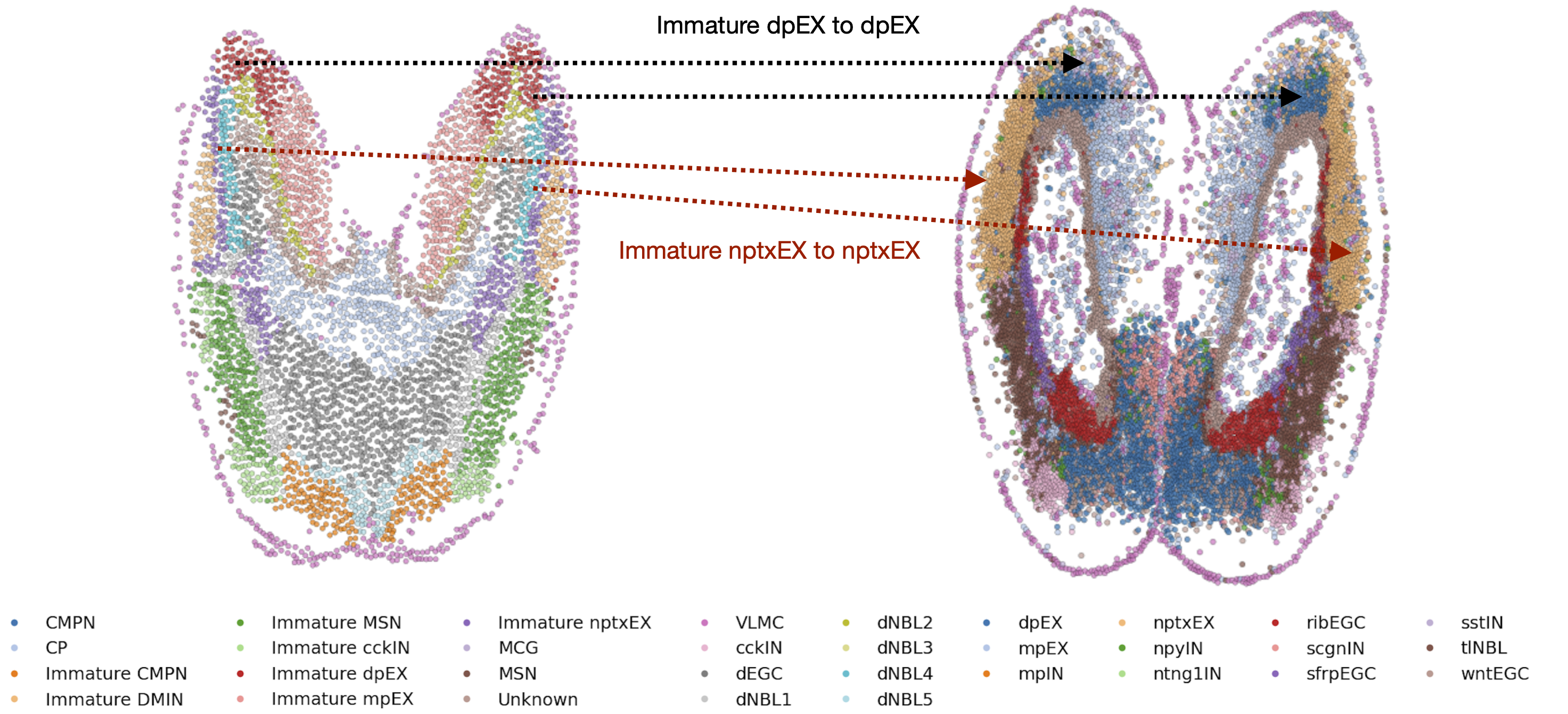} 
  \caption{Visual translation of the bias of NPTX2–NPTXR LR pattern in cell type coupling. }
  \label{fig:consecutive_ref_slices}
\end{figure}

Based on the activation of NPTX2–NPTXR in Figure~\ref{fig:LR_viz}, we observe that the communication pattern naturally biases the optimal couplings towards transitions, such as Immature dpEX $\rightarrow$ dpEX and Immature nptxEX $\rightarrow$ nptxEX (Figure~\ref{fig:consecutive_ref_slices}). These transitions are biologically plausible, as they preserve cell type identity within excitatory neuronal lineages while reflecting maturation within the same functional context. This example highlights the richness of the contextual information captured by our proposed biological prior and demonstrates how incorporating such ligand–receptor–driven cues into the coupling process yields more interpretable, biologically consistent trajectories.

\clearpage
\newpage

\subsection{Cell Type Predictions from ContextFlow}
\label{apdx:IVP cell type progression}

\begin{figure}[H]
    \centering
    \vspace{-0.1in}
    \includegraphics[width=0.8\linewidth]{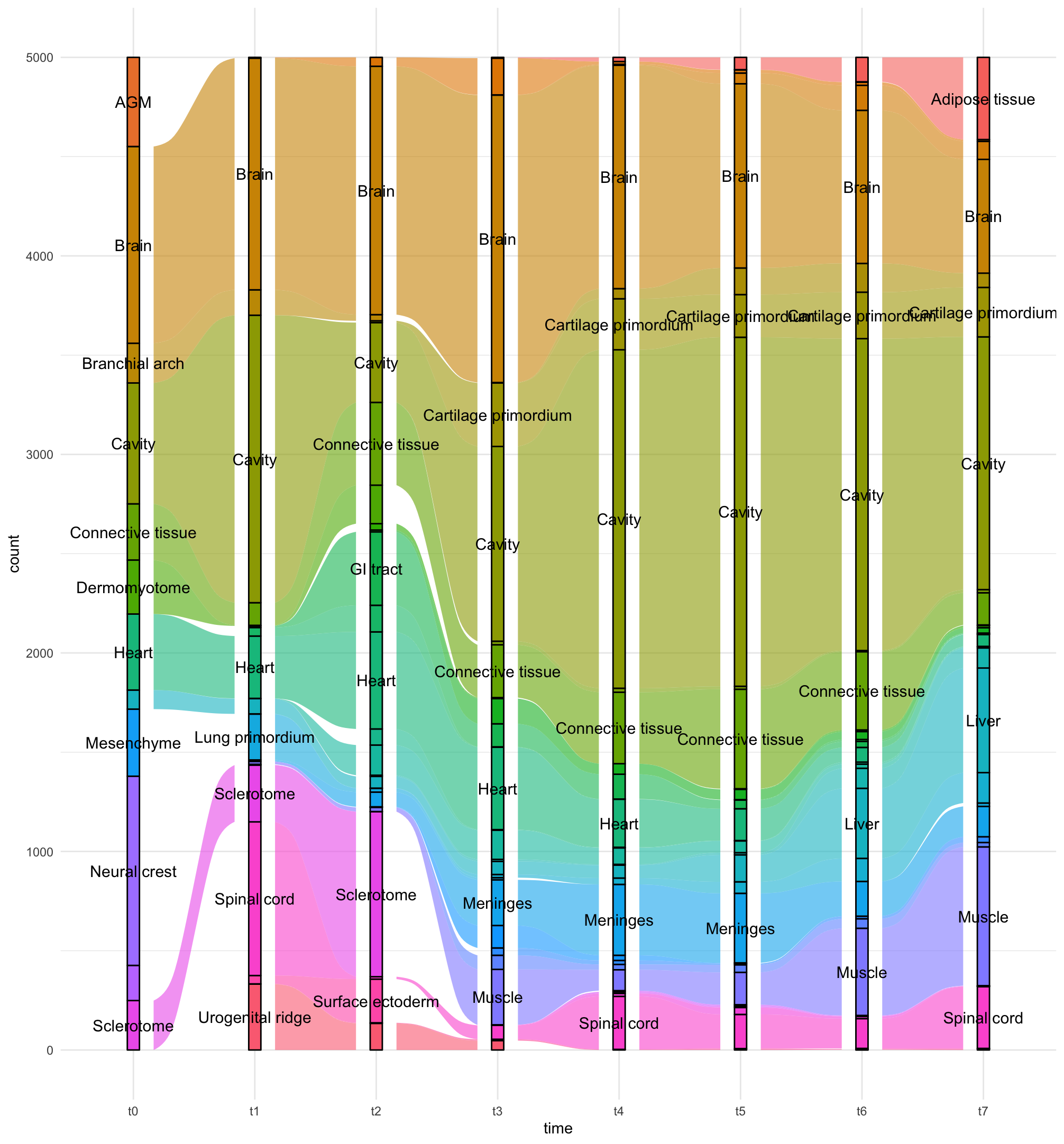}
    \vspace{-0.05in}
    \caption{Temporal predictions from ContextFlow under IVP sampling for major cell types on the Mouse Organogenesis dataset. Early progenitor populations (neural crest and mesenchyme) progressively diminish as development advances, while terminal fates (muscle, cartilage primordium, and liver) emerge at later stages. Major lineages such as the brain, heart, and connective tissue remain continuous throughout. Overall, ContextFlow captures biologically coherent and temporally consistent developmental dynamics.}
    \vspace{-0.1in}
    \label{fig:sankey}
\end{figure}


\section{Statistical Significance Testing}
\label{apdx: pvalues}

We evaluate the statistical significance of ContextFlow's performance using a paired two-sided $t$-test over repeated runs. 
For each metric, we compute paired differences between the best (bolded) setting from the result tables from Section~\ref{sec: experiments} for CTF and the corresponding MOTFM baseline, let $\{d_i\}_{i=1}^n$ be these differences; the test statistic is given by 
$t = \bar{d} / (s_d / \sqrt{n})$, where $\bar{d}$ and $s_d$ are the sample mean and standard deviation of the differences. The $p$-value measures the likelihood of observing such a deviation in metric values by chance, assuming there is no true difference between the methods; the lower the $ p$-value, the better. 
Table \ref{tab:pval_stereoseq_combined} and Table \ref{tab:pval_mosta_combined} report the p-value on the Brain Regeneration and Mouse Organogenesis datasets.
For Liver regeneration, the p-value is computed as $2.74\mathrm{e}{-02}$.

\begin{table}[H]
\caption{P-values for interpolation and extrapolation on  Brain Regeneration from Tables~\ref{tab:stereoseq_interpolation_table} and~\ref{tab:stereoseq_extrapolation_table}.}
\label{tab:pval_stereoseq_combined}
\centering
\small
\renewcommand{\arraystretch}{1.05}
\setlength{\tabcolsep}{5pt}
\resizebox{1.0\textwidth}{!}{
\begin{tabular}{l | cccc | cccc}
\toprule
\multirow{2.4}{*}{\textbf{Sampling}}
& \multicolumn{4}{c|}{\textbf{Interpolation}}
& \multicolumn{4}{c}{\textbf{Extrapolation}} \\
\cmidrule(lr){2-5} \cmidrule(lr){6-9}
& \textbf{Weighted} $\bm{\mathcal{W}_{2}}$ 
& $\bm{\mathcal{W}_{2}}$ 
& \textbf{MMD} 
& \textbf{Energy}
& \textbf{Weighted} $\bm{\mathcal{W}_{2}}$ 
& $\bm{\mathcal{W}_{2}}$ 
& \textbf{MMD} 
& \textbf{Energy} \\
\midrule

Next Step 
& $1.53\mathrm{e}{-02}$ & $3.62\mathrm{e}{-03}$ & $5.64\mathrm{e}{-04}$ & $3.57\mathrm{e}{-04}$
& $1.07\mathrm{e}{-02}$ & $2.58\mathrm{e}{-02}$ & $8.60\mathrm{e}{-02}$ & $1.76\mathrm{e}{-02}$ \\

IVP 
& $1.05\mathrm{e}{-02}$ & $1.57\mathrm{e}{-02}$ & $1.83\mathrm{e}{-05}$ & $6.01\mathrm{e}{-05}$
& $1.53\mathrm{e}{-02}$ & $3.29\mathrm{e}{-02}$ & $5.13\mathrm{e}{-03}$ & $6.33\mathrm{e}{-03}$ \\

\bottomrule
\end{tabular}
}
\end{table}

\begin{table}[H]
\caption{P-values for interpolation and extrapolation on Mouse Organogenesis from Table~\ref{tab:mosta_combined_table}.}
\label{tab:pval_mosta_combined}
\centering
\small
\renewcommand{\arraystretch}{1.05}
\setlength{\tabcolsep}{5pt}
\resizebox{0.8\linewidth}{!}{
\begin{tabular}{cc | cc | cc}
\toprule
\multicolumn{2}{c|}{\textbf{Next Step (Interpolation)}}
& \multicolumn{2}{c|}{\textbf{IVP (Interpolation)}}
& \multicolumn{2}{c}{\textbf{Next Step (Extrapolation)}} \\
\cmidrule(lr){1-2} \cmidrule(lr){3-4} \cmidrule(lr){5-6}
\textbf{Weighted} $\bm{\mathcal{W}_{2}}$ & $\bm{\mathcal{W}_{2}}$
& \textbf{Weighted} $\bm{\mathcal{W}_{2}}$ & $\bm{\mathcal{W}_{2}}$
& \textbf{Weighted} $\bm{\mathcal{W}_{2}}$ & $\bm{\mathcal{W}_{2}}$ \\
\midrule

$5.70\mathrm{e}{-03}$ & $1.06\mathrm{e}{-03}$
& $2.63\mathrm{e}{-02}$ & $2.43\mathrm{e}{-01}$
& $1.96\mathrm{e}{-03}$ & $1.30\mathrm{e}{-03}$ \\

\bottomrule
\end{tabular}
}
\vspace{-0.05in}
\end{table}


\section{Hyperparameter Sensitivity Analysis}
\label{apdx:Hyperparameter_ablations}

\subsection{Ablation on $\lambda$}
\label{apdx:lambda_ablations}

\begin{figure}[H]
\centering

\begin{minipage}{\textwidth}
\centering
\captionof{table}{Extrapolation on the last holdout timestep on the Brain Regeneration dataset.}

\label{tab:salamander_extrap_lambda}
\small
\renewcommand{\arraystretch}{1.05}
\resizebox{0.85\textwidth}{!}{
\begin{tabular}{c cc | cc}
\toprule
\multirow{2.4}{*}{\(\bm{\lambda}\)}
& \multicolumn{2}{c|}{\textbf{Next Step Sampling}}
& \multicolumn{2}{c}{\textbf{IVP Sampling}} \\
\cmidrule(lr){2-3} \cmidrule(lr){4-5}
& \textbf{Weighted} $\bm{\mathcal{W}_{2}}$ & $\bm{\mathcal{W}_{2}}$
& \textbf{Weighted} $\bm{\mathcal{W}_{2}}$ & $\bm{\mathcal{W}_{2}}$ \\
\midrule
0   & $6.968 \pm 0.608$ & $7.198 \pm 0.726$ & $6.243 \pm 0.760$ & $6.220 \pm 0.751$ \\
0.2 & $7.313 \pm 0.384$ & $7.331 \pm 0.467$ & $6.502 \pm 0.634$ & $6.039 \pm 0.733$ \\
0.5 & $7.243 \pm 0.479$ & $7.157 \pm 0.641$ & $6.254 \pm 0.819$ & $5.973 \pm 0.757$ \\
0.8 & $7.333 \pm 0.605$ & $7.334 \pm 0.622$ & $6.598 \pm 0.892$ & $6.402 \pm 1.039$ \\
1   & $7.505 \pm 0.667$ & $7.338 \pm 0.601$ & $5.277 \pm 0.936$ & $6.021 \pm 1.192$ \\
\bottomrule
\end{tabular}
}
\end{minipage}

\vspace{0.2in}

\begin{minipage}{0.44\textwidth}
    \centering
    \includegraphics[width=\linewidth]{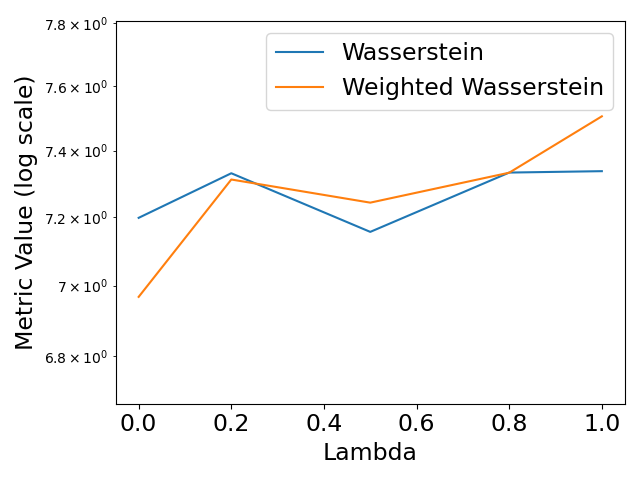}
    \caption*{(a) Next Step Sampling}
\end{minipage}
\hspace{0.1in}
\begin{minipage}{0.44\textwidth}
    \centering
    \includegraphics[width=\linewidth]{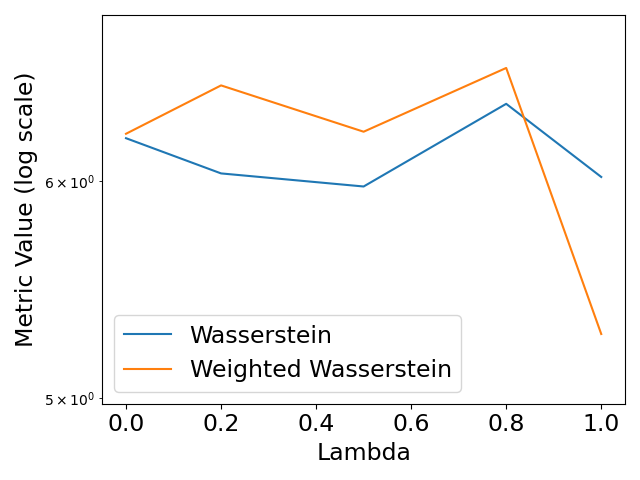}
    \caption*{(b) IVP Sampling}
\end{minipage}
\vspace{-0.05in}
\caption{Performance variation with $\lambda$ for extrapolation on the Brain Regeneration dataset.}
\label{fig:salamander_lambda_full}
\end{figure}

\begin{figure}[H]
\centering

\begin{minipage}{\textwidth}
\centering
\captionof{table}{Interpolation on the middle holdout timestep 3 on the Brain Regeneration dataset.}

\label{tab:salamander_interp_lambda}
\small
\renewcommand{\arraystretch}{1.05}
\resizebox{0.85\textwidth}{!}{
\begin{tabular}{c cc | cc}
\toprule
\multirow{2.4}{*}{\(\bm{\lambda}\)}
& \multicolumn{2}{c|}{\textbf{Next Step Sampling}}
& \multicolumn{2}{c}{\textbf{IVP Sampling}} \\
\cmidrule(lr){2-3} \cmidrule(lr){4-5}
& \textbf{Weighted} $\bm{\mathcal{W}_{2}}$ & $\bm{\mathcal{W}_{2}}$
& \textbf{Weighted} $\bm{\mathcal{W}_{2}}$ & $\bm{\mathcal{W}_{2}}$ \\
\midrule
0   & $2.528 \pm 0.143$ & $2.534 \pm 0.180$ & $3.925 \pm 0.267$ & $4.375 \pm 0.297$ \\
0.2 & $2.544 \pm 0.093$ & $2.389 \pm 0.183$ & $4.153 \pm 0.432$ & $4.393 \pm 0.369$ \\
0.5 & $2.519 \pm 0.167$ & $2.412 \pm 0.158$ & $3.917 \pm 0.343$ & $4.159 \pm 0.455$ \\
0.8 & $2.533 \pm 0.137$ & $2.352 \pm 0.142$ & $4.151 \pm 0.193$ & $4.408 \pm 0.285$ \\
1   & $2.316 \pm 0.141$ & $1.969 \pm 0.221$ & $3.905 \pm 0.395$ & $4.188 \pm 0.685$ \\
\bottomrule
\end{tabular}
}
\end{minipage}

\vspace{0.2in}

\begin{minipage}{0.46\textwidth}
    \centering
    \includegraphics[width=\linewidth]{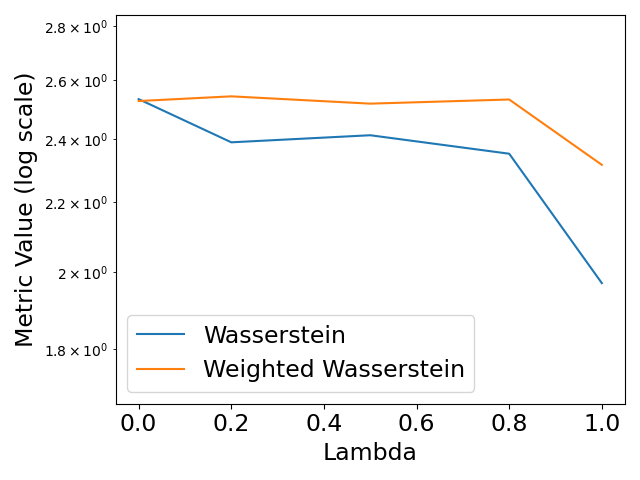}
    \caption*{(a) Next Step Sampling}
\end{minipage}
\hspace{0.1in}
\begin{minipage}{0.46\textwidth}
    \centering
    \includegraphics[width=\linewidth]{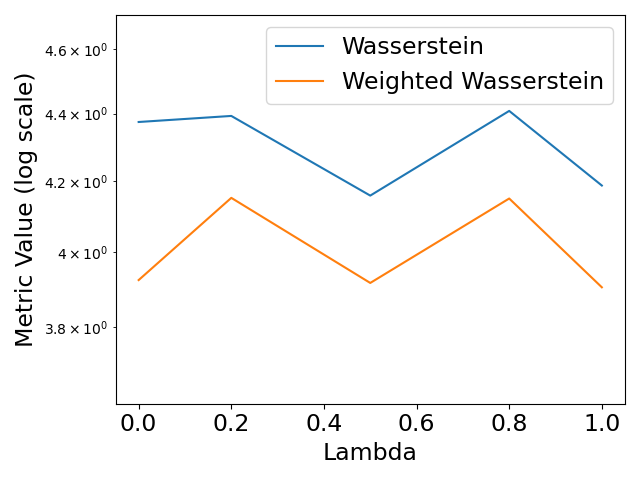}
    \caption*{(b) IVP Sampling}
\end{minipage}
\vspace{-0.05in}
\caption{Performance variation with $\lambda$ for interpolation on the Brain Regeneration dataset.}
\label{fig:salamander_lambda_interp_full}
\end{figure}

\begin{figure}[t]
\centering

\begin{minipage}{\textwidth}
\centering
\captionof{table}{Interpolation for the holdout timestep 5 on the Mouse Organogenesis dataset.}

\label{tab:mosta_interp_lambda}
\small
\renewcommand{\arraystretch}{1.05}
\resizebox{0.85\textwidth}{!}{
\begin{tabular}{c cc | cc}
\toprule
\multirow{2.4}{*}{\(\bm{\lambda}\)}
& \multicolumn{2}{c|}{\textbf{Next Step Sampling}}
& \multicolumn{2}{c}{\textbf{IVP Sampling}} \\
\cmidrule(lr){2-3} \cmidrule(lr){4-5}
& \textbf{Weighted} $\bm{\mathcal{W}_{2}}$ & $\bm{\mathcal{W}_{2}}$
& \textbf{Weighted} $\bm{\mathcal{W}_{2}}$ & $\bm{\mathcal{W}_{2}}$ \\
\midrule
0   & $1.884 \pm 0.027$ & $1.862 \pm 0.123$ & $3.244 \pm 0.713$ & $3.946 \pm 1.671$ \\
0.2 & $1.896 \pm 0.028$ & $1.899 \pm 0.078$ & $2.990 \pm 0.205$ & $3.273 \pm 0.518$ \\
0.5 & $1.871 \pm 0.030$ & $1.919 \pm 0.067$ & $2.814 \pm 0.414$ & $3.233 \pm 0.567$ \\
0.8 & $1.878 \pm 0.031$ & $1.890 \pm 0.064$ & $2.966 \pm 0.411$ & $3.345 \pm 0.508$ \\
1   & $1.898 \pm 0.029$ & $1.866 \pm 0.097$ & $5.200 \pm 0.799$ & $6.306 \pm 1.037$ \\
\bottomrule
\end{tabular}
}
\end{minipage}

\vspace{0.2in}

\begin{minipage}{0.46\textwidth}
    \centering
    \includegraphics[width=\linewidth]{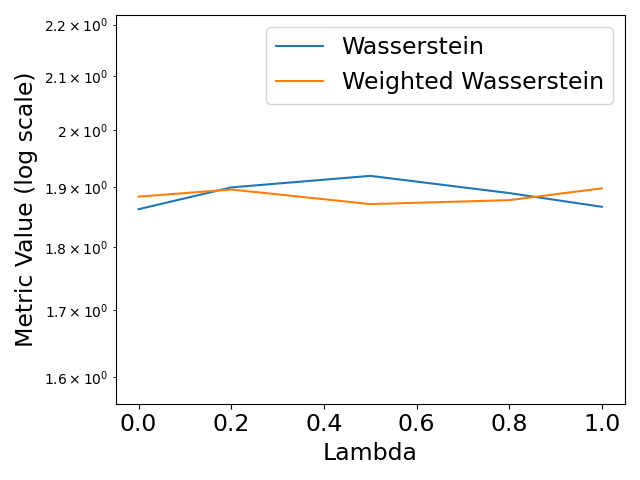}
    \caption*{(a) Next Step Sampling}
\end{minipage}
\hspace{0.1in}
\begin{minipage}{0.46\textwidth}
    \centering
    \includegraphics[width=\linewidth]{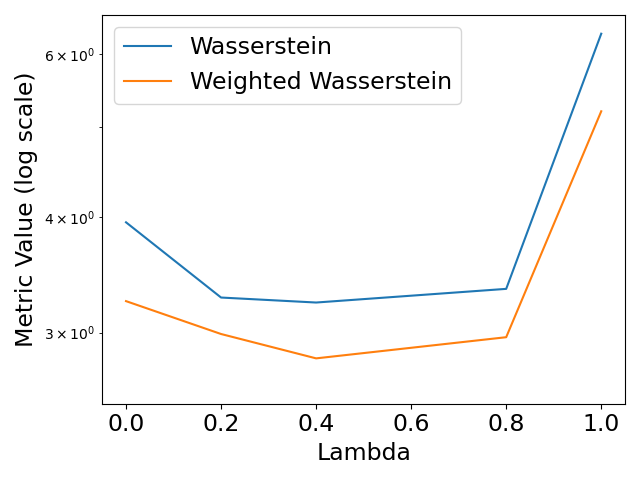}
    \caption*{(b) IVP Sampling}
\end{minipage}
\vspace{-0.05in}
\caption{Performance variation with $\lambda$ for interpolation on the Mouse Organogenesis dataset.}
\label{fig:mosta_lambda_interp_full}
\end{figure}

\begin{figure}[t]
\centering

\begin{minipage}{\textwidth}
\centering
\captionof{table}{Extrapolation for holdout timestep 5 on the Mouse Organogenesis dataset.}

\label{tab:mosta_extrap_lambda}
\small
\renewcommand{\arraystretch}{1.05}
\resizebox{0.90\textwidth}{!}{
\begin{tabular}{c cc | cc}
\toprule
\multirow{2.4}{*}{\(\bm{\lambda}\)}
& \multicolumn{2}{c|}{\textbf{Next Step Sampling}}
& \multicolumn{2}{c}{\textbf{IVP Sampling}} \\
\cmidrule(lr){2-3} \cmidrule(lr){4-5}
& \textbf{Weighted} $\bm{\mathcal{W}_{2}}$ & $\bm{\mathcal{W}_{2}}$
& \textbf{Weighted} $\bm{\mathcal{W}_{2}}$ & $\bm{\mathcal{W}_{2}}$ \\
\midrule

0   & $1.508 \pm 0.047$ & $1.386 \pm 0.088$ 
    & $6.986 \pm 1.296$ & $23.807 \pm 7.516$ \\

0.2 & $1.614 \pm 0.081$ & $1.642 \pm 0.136$ 
    & $6.907 \pm 1.471$ & $24.675 \pm 9.126$ \\

0.5 & $1.638 \pm 0.069$ & $1.676 \pm 0.114$ 
    & $7.749 \pm 1.377$ & $32.480 \pm 8.322$ \\

0.8 & $1.617 \pm 0.042$ & $1.680 \pm 0.094$ 
    & $7.645 \pm 2.376$ & $29.854 \pm 10.015$ \\

1   & $1.906 \pm 0.071$ & $1.892 \pm 0.092$ 
    & $6.378 \pm 1.813$ & $11.845 \pm 6.567$ \\

\bottomrule
\end{tabular}
}
\end{minipage}

\vspace{0.2in}

\begin{minipage}{0.46\textwidth}
    \centering
    \includegraphics[width=\linewidth]{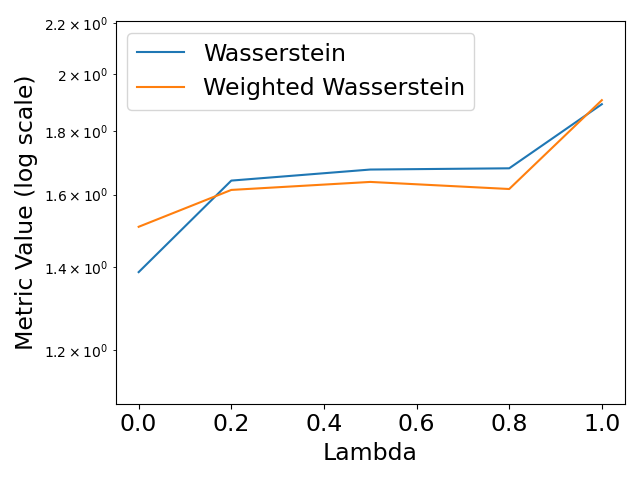}
    \caption*{(a) Next Step Sampling}
\end{minipage}
\hspace{0.1in}
\begin{minipage}{0.46\textwidth}
    \centering
    \includegraphics[width=\linewidth]{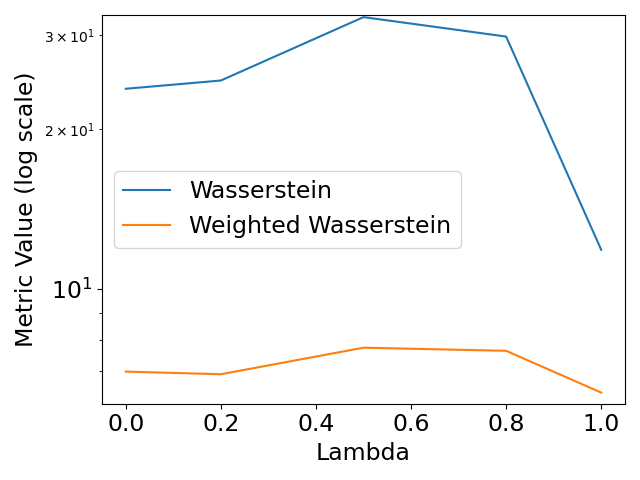}
    \caption*{(b) IVP Sampling}
\end{minipage}
\vspace{-0.05in}
\caption{Performance variation with $\lambda$ for extrapolation on the Mouse Organogenesis dataset.}
\label{fig:mosta_lambda_extrap_full}
\end{figure}

\begin{figure}[H]
    \centering
    \includegraphics[width=0.5\linewidth]{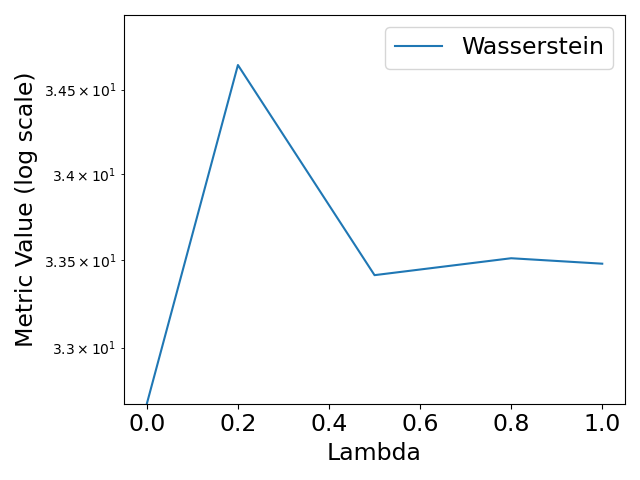}
    
    \caption{Performance variation with $\lambda$ for interpolation on the Liver Regeneration dataset.}
    \label{fig:liver_lambda_interp_full}
\end{figure}

First, we conduct a sensitivity analysis on all three datasets for the parameter $\lambda$, which controls the relative importance of the two priors. All the other hyperparameters are kept constant. From Tables  \ref{tab:salamander_extrap_lambda}-\ref{tab:liver_interp_next_step_lambda} and Figures \ref{fig:salamander_lambda_full}-\ref{fig:liver_lambda_interp_full}, we observe that the best performance is usually achieved towards the extremities, at $\lambda=0$ or $\lambda=1$, with values near the latter dominating more often. We hypothesize that the spatial smoothness prior serves as a proxy for the spatial distance between cells from different slices and always carries relevant information, encoding the structural information present in the data. 
On the other hand, the informativeness of cell–cell communication patterns depends on how distinct the ligand–receptor features are at a given time step compared with those of neighboring ones. When these features remain highly similar across consecutive timesteps, they contribute little to the discriminative signal that the OT objective can leverage. Consequently, the influence of communication priors is strongly dataset- and timestep-dependent. This effect is also reflected in the observation that settings with $\lambda = 0$ tend to perform worse when using Next-Step sampling—where local, immediate effects dominate—than under IVP sampling, which integrates information over the entire preceding trajectory.
We therefore recommend experimenting with different values of $\lambda$ (e.g., $0$, $0.8$, or $1$) depending on the specific use case and context. Due to ContextFlow's scalability, hyperparameter exploration can be performed efficiently, enabling rapid assessment of the effect of $\lambda$.


\subsection{Ablation on $r$}
\label{apdx:radius_ablations}

We examine the effect of varying the neighborhood radius $r$ used to define the boundary for computing the spatial smoothness prior. We evaluate two settings for Brain Regeneration: $\lambda = 1$ and $\lambda = 0.8$, corresponding to using only the spatial prior and a setting with a modest contribution from the cell–cell communication prior. 
From Tables \ref{tab:salamander_extrap_radius}-\ref{tab:salamander_lr_interp_radius} and their corresponding Figures \ref{fig:salamander_radius_extrap_full}-\ref{fig:salamander_lr_radius_interp_full}, we observe that the optimal neighborhood radius tends to lie toward the lower end of the tested range. Radii smaller than this optimum degrade performance by failing to capture sufficient local context, resulting in neighborhood means that are overly similar to individual cellular profiles. Conversely, increasing the radius beyond the optimal range also reduces performance, as the neighborhood begins to include cells from distinct types or spatial regions, thereby diluting the local signal. While certain deviations from this trend occur, likely reflecting underlying biological complexity, this behavior is consistent with the trade-off between spatial specificity and contextual coverage inherent to neighborhood-based priors.
For our case, we set the radius by considering the timestep with the fewest cells, dividing it by $2$ (to account for different hemispheres), and dividing by the number of cell types present in that timestep. For the dataset considered in this study, Stage $44$ had the fewest number of cells, $1400$, with approximately $10$ cell types. We thus set the radius at $50$ in our studies.

\subsection{Ablation on $\epsilon$}
\label{apdx:epsilon_ablations}

Additionally, we conduct an ablation study on the Brain Regeneration dataset by varying the parameter $\epsilon$, which weights the entropic term in the OT objective. The ContextFlow configuration we consider here is CTF-H ($\lambda=1$), which includes only the spatial smoothness prior. As observed from Tables \ref{tab:salamander_extrap_epsilon} and \ref{tab:salamander_interp_epsilon}, as well as their corresponding Figures \ref{fig:salamander_epsilon_extrap_full} and \ref{fig:salamander_epsilon_interp_full}, drastically decreasing $\epsilon$ results in the OT formulation to ignore the relative entropic term containing the prior information and only to consider the transport cost, resulting in higher Wasserstein values. Furthermore, in accordance with results from Theorem \ref{thm: PAER Sinkhorn}, increasing $\epsilon$ too much still does not drastically degrade the performance, as the prior matrix $M$ acts as a soft filter and prohibits uniform couplings.
While setting $\epsilon$, one must look at the Gibbs kernel used in the Sinkhorn Algorithm $\exp(-\mathbf{C}/\epsilon)$, since a small $\epsilon$ can cause potential numerical issues. We thus set $\epsilon$ by examining the order of the cost matrix $\mathbf{C}$, and, for the studies above, we set it to $100$ based on the median of all elements in the cost matrix.

\clearpage
\newpage

\begin{table}[t]
\centering
    \centering
    \caption{Interpolation for holdout timestep 2 with IVP Sampling on the Liver Regeneration dataset.}
    \vspace{0.05in}
    \label{tab:liver_interp_next_step_lambda}
    \small
    \renewcommand{\arraystretch}{1.1}
    \resizebox{0.3\textwidth}{!}{
    \begin{tabular}{c c}
    \toprule
    \(\bm{\lambda}\) & $\bm{\mathcal{W}_{2}}$ \\
    \midrule
    $0$   & $32.682 \pm 1.472$ \\
    $0.2$ & $34.647 \pm 1.461$ \\
    $0.5$ & $33.414 \pm 0.995$ \\
    $0.8$ & $33.512 \pm 0.786$ \\
    $1$   & $33.481 \pm 1.001$ \\
    \bottomrule
    \end{tabular}
    }
\end{table}

%

%


\begin{figure}[H]
\centering

\begin{minipage}{\textwidth}
\centering
\captionof{table}{Extrapolation with CTF-H at $\lambda=1$ (only using the spatial smoothness prior) for the last holdout timestep on the Brain Regeneration dataset.}

\label{tab:salamander_extrap_radius}
\small
\renewcommand{\arraystretch}{1.05}
\resizebox{0.85\textwidth}{!}{
\begin{tabular}{c cc | cc}
\toprule
\multirow{2.4}{*}{\(\bm{\text{Radius}}\)}
& \multicolumn{2}{c|}{\textbf{Next Step Sampling}} 
& \multicolumn{2}{c}{\textbf{IVP Sampling}} \\
\cmidrule(lr){2-3} \cmidrule(lr){4-5}
& \textbf{Weighted} $\bm{\mathcal{W}_{2}}$ & $\bm{\mathcal{W}_{2}}$
& \textbf{Weighted} $\bm{\mathcal{W}_{2}}$ & $\bm{\mathcal{W}_{2}}$ \\
\midrule
12  & $6.228 \pm 1.276$ & $6.163 \pm 1.490$ & $4.415 \pm 0.580$ & $6.843 \pm 4.812$ \\
25  & $6.244 \pm 1.066$ & $6.231 \pm 1.043$ & $6.500 \pm 1.751$ & $5.613 \pm 1.561$ \\
50  & $7.505 \pm 0.667$ & $7.338 \pm 0.601$ & $5.277 \pm 0.936$ & $6.021 \pm 1.192$ \\
100 & $6.892 \pm 0.930$ & $6.702 \pm 0.631$ & $7.061 \pm 1.677$ & $6.860 \pm 1.880$ \\
150 & $7.747 \pm 0.923$ & $7.793 \pm 0.934$ & $9.796 \pm 3.847$ & $10.656 \pm 6.591$ \\
200 & $6.039 \pm 0.282$ & $5.764 \pm 0.272$ & $5.630 \pm 0.793$ & $5.000 \pm 0.735$ \\
250 & $6.804 \pm 1.011$ & $6.834 \pm 1.124$ & $6.578 \pm 1.611$ & $7.379 \pm 2.864$ \\
\bottomrule
\end{tabular}
}
\end{minipage}

\vspace{0.2in}

\begin{minipage}{0.46\textwidth}
    \centering
    \includegraphics[width=\linewidth]{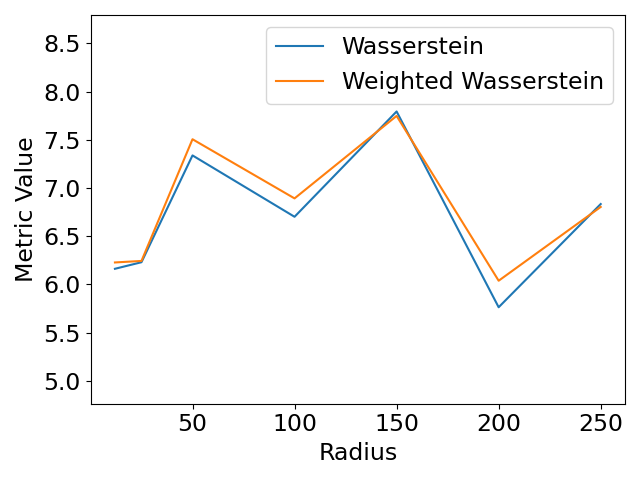}
    \caption*{(a) Next Step Sampling}
\end{minipage}
\hspace{0.1in}
\begin{minipage}{0.46\textwidth}
    \centering
    \includegraphics[width=\linewidth]{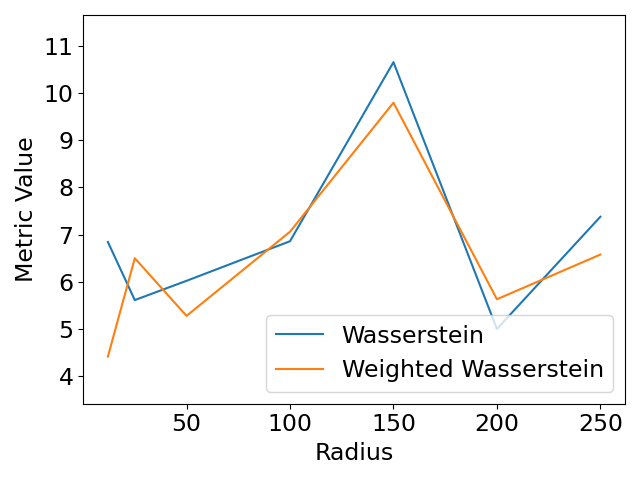}
    \caption*{(b) IVP Sampling}
\end{minipage}
\vspace{-0.05in}
\caption{Performance variation with radius for extrapolation on the Brain Regeneration dataset.}
\label{fig:salamander_radius_extrap_full}
\end{figure}

\begin{figure}[H]
\centering

\begin{minipage}{\textwidth}
\centering
\captionof{table}{Interpolation for the middle holdout timestep 3 for  CTF-H at $\lambda = 1$ (only using the spatial smoothness prior) on the Brain Regeneration dataset.}

\label{tab:salamander_interp_radius}
\small
\renewcommand{\arraystretch}{1.05}
\resizebox{0.85\textwidth}{!}{
\begin{tabular}{c cc | cc}
\toprule
\multirow{2.4}{*}{\(\bm{\text{Radius}}\)}
& \multicolumn{2}{c|}{\textbf{Next Step Sampling}} 
& \multicolumn{2}{c}{\textbf{IVP Sampling}} \\
\cmidrule(lr){2-3} \cmidrule(lr){4-5}
& \textbf{Weighted} $\bm{\mathcal{W}_{2}}$ & $\bm{\mathcal{W}_{2}}$
& \textbf{Weighted} $\bm{\mathcal{W}_{2}}$ & $\bm{\mathcal{W}_{2}}$ \\
\midrule
12  & $4.293 \pm 0.318$ & $3.547 \pm 0.343$ & $2.650 \pm 0.204$ & $2.346 \pm 0.251$ \\
25  & $5.019 \pm 0.270$ & $3.968 \pm 0.274$ & $2.408 \pm 0.239$ & $1.808 \pm 0.257$ \\
50  & $2.316 \pm 0.141$ & $1.969 \pm 0.221$ & $3.905 \pm 0.395$ & $4.188 \pm 0.685$ \\
100 & $4.590 \pm 0.360$ & $3.359 \pm 0.166$ & $2.812 \pm 0.240$ & $2.220 \pm 0.231$ \\
150 & $4.731 \pm 0.424$ & $3.819 \pm 0.239$ & $3.533 \pm 0.220$ & $3.290 \pm 0.778$ \\
200 & $4.548 \pm 0.780$ & $4.249 \pm 1.315$ & $3.751 \pm 0.725$ & $3.677 \pm 1.016$ \\
250 & $4.768 \pm 1.994$ & $4.782 \pm 4.129$ & $4.281 \pm 0.985$ & $4.103 \pm 1.081$ \\
\bottomrule
\end{tabular}
}
\end{minipage}

\vspace{0.2in}

\begin{minipage}{0.46\textwidth}
    \centering
    \includegraphics[width=\linewidth]{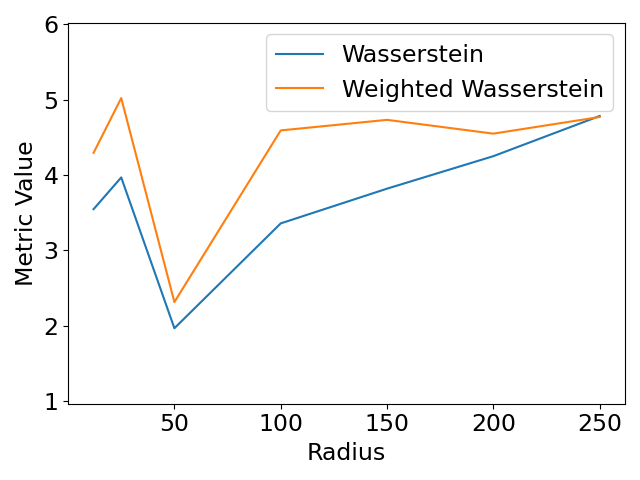}
    \caption*{(a) Next Step Sampling}
\end{minipage}
\hspace{0.1in}
\begin{minipage}{0.46\textwidth}
    \centering
    \includegraphics[width=\linewidth]{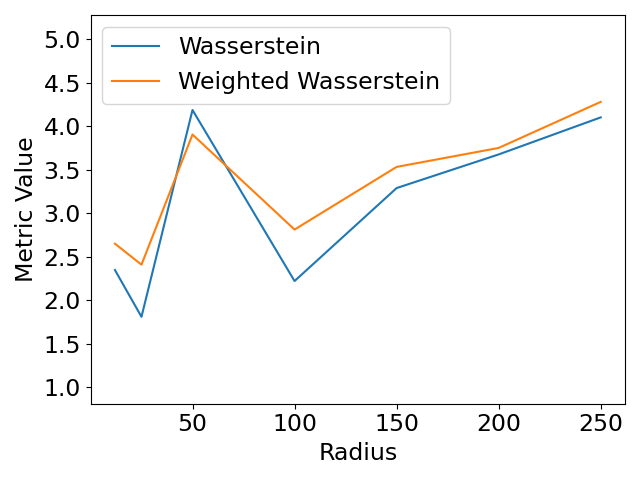}
    \caption*{(b) IVP Sampling}
\end{minipage}
\vspace{-0.05in}
\caption{Performance variation with radius for interpolation on the Brain Regeneration dataset.}
\label{fig:salamander_radius_interp_full}
\end{figure}


\begin{figure}[H]
\centering

\begin{minipage}{\textwidth}
\centering
\captionof{table}{Extrapolation for the last holdout timestep with CTF-H at $\lambda=0.8$ on Brain Regeneration.}

\label{tab:salamander_lr_extrap_radius}
\small
\renewcommand{\arraystretch}{1.05}
\resizebox{0.85\textwidth}{!}{
\begin{tabular}{c cc | cc}
\toprule
\multirow{2.4}{*}{\(\bm{\text{Radius}}\)}
& \multicolumn{2}{c|}{\textbf{Next Step Sampling}}
& \multicolumn{2}{c}{\textbf{IVP Sampling}} \\
\cmidrule(lr){2-3} \cmidrule(lr){4-5}
& \textbf{Weighted} $\bm{\mathcal{W}_{2}}$ & $\bm{\mathcal{W}_{2}}$
& \textbf{Weighted} $\bm{\mathcal{W}_{2}}$ & $\bm{\mathcal{W}_{2}}$ \\
\midrule
12  & $6.924 \pm 1.178$ & $6.831 \pm 1.132$ & $8.049 \pm 2.162$ & $13.223 \pm 12.164$ \\
25  & $6.633 \pm 0.780$ & $6.350 \pm 0.654$ & $6.621 \pm 1.608$ & $9.223 \pm 8.396$ \\
50  & $7.130 \pm 0.389$ & $7.260 \pm 0.632$ & $5.971 \pm 0.461$ & $5.836 \pm 1.181$ \\
100 & $6.411 \pm 0.522$ & $6.351 \pm 0.456$ & $5.932 \pm 0.264$ & $6.434 \pm 0.840$ \\
150 & $6.498 \pm 1.056$ & $6.501 \pm 1.098$ & $6.033 \pm 0.882$ & $7.203 \pm 2.443$ \\
200 & $6.052 \pm 0.873$ & $6.129 \pm 1.052$ & $5.852 \pm 1.085$ & $6.247 \pm 1.731$ \\
250 & $6.449 \pm 0.909$ & $6.278 \pm 0.726$ & $6.151 \pm 0.986$ & $11.261 \pm 7.063$ \\
\bottomrule
\end{tabular}
}
\end{minipage}

\vspace{0.2in}

\begin{minipage}{0.46\textwidth}
    \centering
    \includegraphics[width=\linewidth]{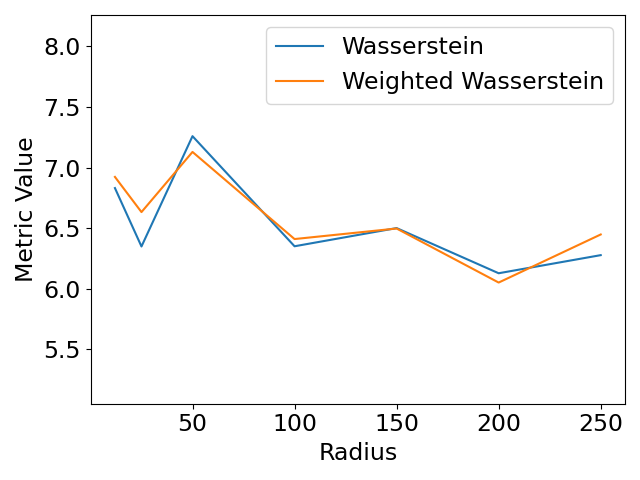}
    \caption*{(a) Next Step Sampling}
\end{minipage}
\hspace{0.1in}
\begin{minipage}{0.46\textwidth}
    \centering
    \includegraphics[width=\linewidth]{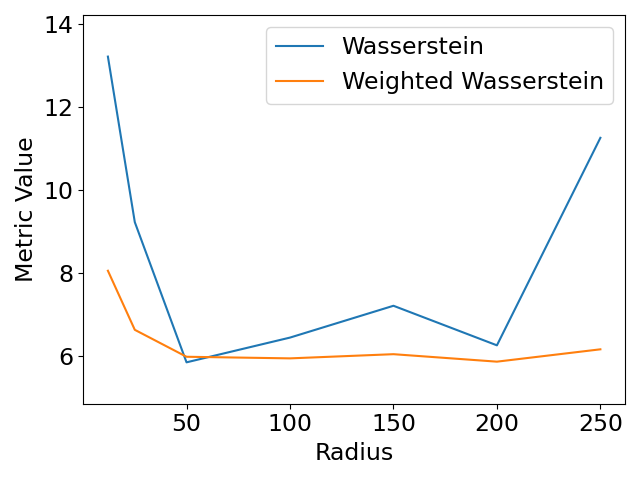}
    \caption*{(b) IVP Sampling}
\end{minipage}
\vspace{-0.05in}
\caption{Performance variation with radius for extrapolation on the Brain Regeneration dataset.}
\label{fig:salamander_lr_radius_extrap_full}
\end{figure}

\begin{figure}[H]
\centering

\begin{minipage}{\textwidth}
\centering
\captionof{table}{Interpolation for the middle holdout timestep with CTF-H at $\lambda=0.8$ on Brain Regeneration.}

\label{tab:salamander_lr_interp_radius}
\small
\renewcommand{\arraystretch}{1.05}
\resizebox{0.85\textwidth}{!}{
\begin{tabular}{c cc | cc}
\toprule
\multirow{2.4}{*}{\(\bm{\text{Radius}}\)}
& \multicolumn{2}{c|}{\textbf{Next Step Sampling}} 
& \multicolumn{2}{c}{\textbf{IVP Sampling}} \\
\cmidrule(lr){2-3} \cmidrule(lr){4-5}
& \textbf{Weighted} $\bm{\mathcal{W}_{2}}$ & $\bm{\mathcal{W}_{2}}$
& \textbf{Weighted} $\bm{\mathcal{W}_{2}}$ & $\bm{\mathcal{W}_{2}}$ \\
\midrule
12  & $5.320 \pm 1.714$ & $4.709 \pm 2.260$ & $3.722 \pm 1.114$ & $3.656 \pm 1.327$ \\
25  & $4.943 \pm 1.384$ & $4.467 \pm 1.821$ & $3.350 \pm 1.548$ & $3.112 \pm 1.418$ \\
50  & $2.440 \pm 0.090$ & $2.302 \pm 0.137$ & $4.181 \pm 0.035$ & $4.238 \pm 0.068$ \\
100 & $4.028 \pm 0.648$ & $3.417 \pm 0.869$ & $2.956 \pm 0.580$ & $2.678 \pm 0.535$ \\
150 & $5.408 \pm 0.889$ & $4.669 \pm 1.364$ & $4.535 \pm 0.823$ & $4.209 \pm 0.884$ \\
200 & $7.110 \pm 2.581$ & $6.490 \pm 3.543$ & $4.043 \pm 1.441$ & $3.754 \pm 1.350$ \\
250 & $4.502 \pm 0.573$ & $3.689 \pm 1.204$ & $3.532 \pm 1.148$ & $3.457 \pm 1.217$ \\
\bottomrule
\end{tabular}
}
\end{minipage}

\vspace{0.2in}

\begin{minipage}{0.46\textwidth}
    \centering
    \includegraphics[width=\linewidth]{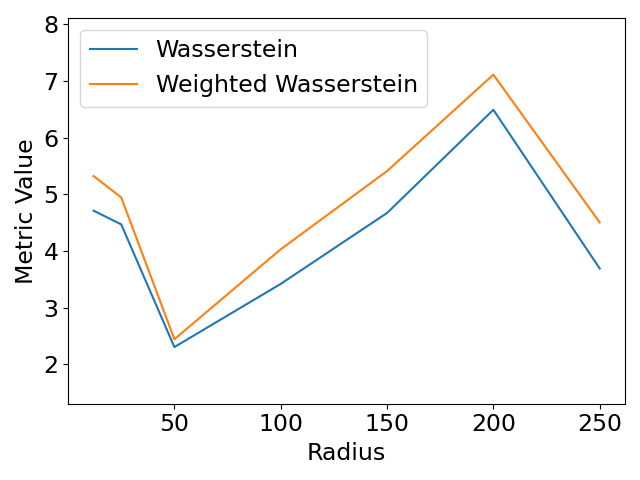}
    \caption*{(a) Next Step Sampling}
\end{minipage}
\hfill
\begin{minipage}{0.46\textwidth}
    \centering
    \includegraphics[width=\linewidth]{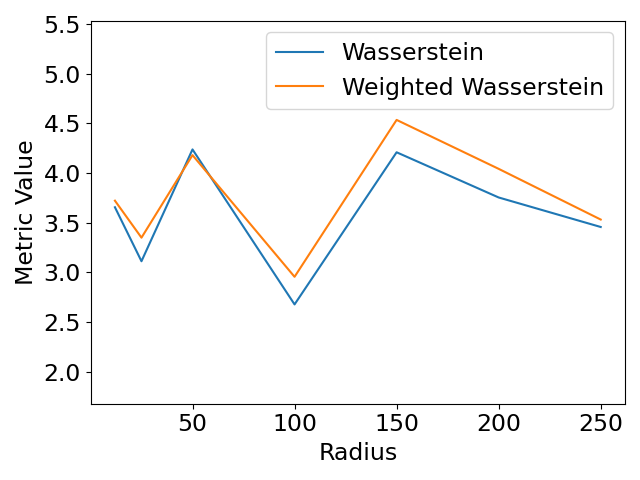}
    \caption*{(b) IVP Sampling}
\end{minipage}
\vspace{-0.05in}
\caption{Performance variation with radius for interpolation on the Brain Regeneration dataset}
\label{fig:salamander_lr_radius_interp_full}
\end{figure}

\begin{figure}[H]
\centering

\begin{minipage}{\textwidth}
\centering
\captionof{table}{Extrapolation for the last holdout timestep on the Brain Regeneration dataset.}

\label{tab:salamander_extrap_epsilon}
\small
\renewcommand{\arraystretch}{1.05}
\resizebox{0.85\textwidth}{!}{
\begin{tabular}{c cc | cc}
\toprule
\multirow{2.4}{*}{\(\bm{\epsilon}\)}
& \multicolumn{2}{c|}{\textbf{Next Step Sampling}} 
& \multicolumn{2}{c}{\textbf{IVP Sampling}} \\
\cmidrule(lr){2-3} \cmidrule(lr){4-5}
& \textbf{Weighted} $\bm{\mathcal{W}_{2}}$ & $\bm{\mathcal{W}_{2}}$
& \textbf{Weighted} $\bm{\mathcal{W}_{2}}$ & $\bm{\mathcal{W}_{2}}$ \\
\midrule
0.001  & $6.007 \pm 0.516$ & $5.939 \pm 0.286$ & $6.260 \pm 1.123$ & $7.301 \pm 2.935$ \\
0.01   & $6.240 \pm 0.870$ & $6.254 \pm 1.111$ & $6.396 \pm 0.236$ & $7.231 \pm 0.968$ \\
0.1    & $6.579 \pm 0.744$ & $6.861 \pm 0.845$ & $6.758 \pm 1.826$ & $7.283 \pm 2.068$ \\
1      & $5.648 \pm 0.471$ & $5.721 \pm 0.595$ & $6.010 \pm 0.674$ & $5.905 \pm 0.737$ \\
10     & $6.841 \pm 0.597$ & $6.940 \pm 0.671$ & $5.532 \pm 1.775$ & $6.646 \pm 1.926$ \\
100    & $7.166 \pm 0.991$ & $7.094 \pm 1.148$ & $6.455 \pm 3.047$ & $5.650 \pm 1.928$ \\
1000   & $6.291 \pm 1.041$ & $6.300 \pm 1.052$ & $7.382 \pm 2.553$ & $7.626 \pm 3.204$ \\
10000  & $6.587 \pm 0.805$ & $6.641 \pm 1.083$ & $5.754 \pm 0.741$ & $7.546 \pm 3.599$ \\
\bottomrule
\end{tabular}
}
\end{minipage}

\vspace{0.2in}

\begin{minipage}{0.46\textwidth}
    \centering
    \includegraphics[width=\linewidth]{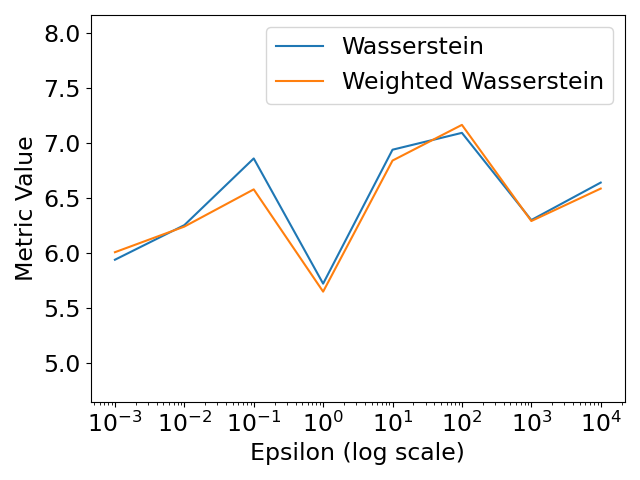}
    \caption*{(a) Next Step Sampling}
\end{minipage}
\hspace{0.1in}
\begin{minipage}{0.46\textwidth}
    \centering
    \includegraphics[width=\linewidth]{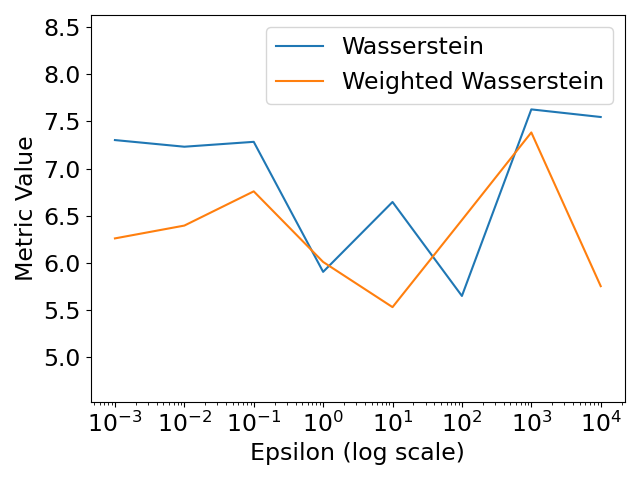}
    \caption*{(b) IVP Sampling}
\end{minipage}
\vspace{-0.05in}
\caption{Performance variation with $\epsilon$ for extrapolation  on the Brain Regeneration Dataset.}
\label{fig:salamander_epsilon_extrap_full}
\end{figure}

\begin{figure}[H]
\centering

\begin{minipage}{\textwidth}
\centering
\captionof{table}{Interpolation for the middle holdout timestep on the Brain Regeneration dataset.}

\label{tab:salamander_interp_epsilon}
\small
\renewcommand{\arraystretch}{1.05}
\resizebox{0.85\textwidth}{!}{
\begin{tabular}{c cc | cc}
\toprule
\multirow{2.4}{*}{\(\bm{\epsilon}\)}
& \multicolumn{2}{c|}{\textbf{Next Step Sampling}} 
& \multicolumn{2}{c}{\textbf{IVP Sampling}} \\
\cmidrule(lr){2-3} \cmidrule(lr){4-5}
& \textbf{Weighted} $\bm{\mathcal{W}_{2}}$ & $\bm{\mathcal{W}_{2}}$
& \textbf{Weighted} $\bm{\mathcal{W}_{2}}$ & $\bm{\mathcal{W}_{2}}$ \\
\midrule
0.001  & $2.899 \pm 0.582$ & $2.715 \pm 0.653$ & $4.056 \pm 0.542$ & $3.286 \pm 0.289$ \\
0.01   & $4.520 \pm 2.066$ & $4.589 \pm 2.298$ & $6.915 \pm 3.573$ & $7.125 \pm 5.289$ \\
0.1    & $2.573 \pm 0.476$ & $2.472 \pm 0.507$ & $3.772 \pm 0.642$ & $3.046 \pm 0.537$ \\
1      & $2.865 \pm 0.612$ & $2.785 \pm 0.576$ & $4.255 \pm 0.679$ & $3.355 \pm 0.584$ \\
10     & $2.899 \pm 0.865$ & $2.833 \pm 0.984$ & $4.908 \pm 1.130$ & $4.159 \pm 1.526$ \\
100    & $2.338 \pm 0.101$ & $1.835 \pm 0.171$ & $5.069 \pm 0.985$ & $4.322 \pm 1.461$ \\
1000   & $3.104 \pm 0.663$ & $2.321 \pm 0.521$ & $5.109 \pm 0.948$ & $3.974 \pm 1.227$ \\
10000  & $2.838 \pm 0.281$ & $2.176 \pm 0.315$ & $4.557 \pm 0.710$ & $3.373 \pm 0.833$ \\
\bottomrule
\end{tabular}
}
\end{minipage}

\vspace{0.25in}

\begin{minipage}{0.46\textwidth}
    \centering
    \includegraphics[width=\linewidth]{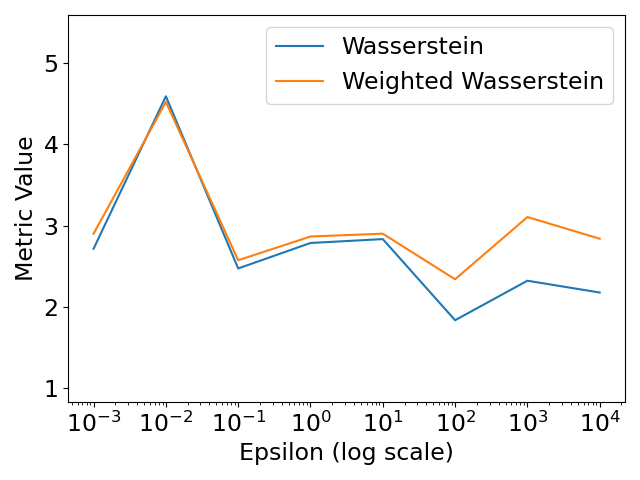}
    \caption*{(a) Next Step Sampling}
\end{minipage}
\hspace{0.1in}
\begin{minipage}{0.46\textwidth}
    \centering
    \includegraphics[width=\linewidth]{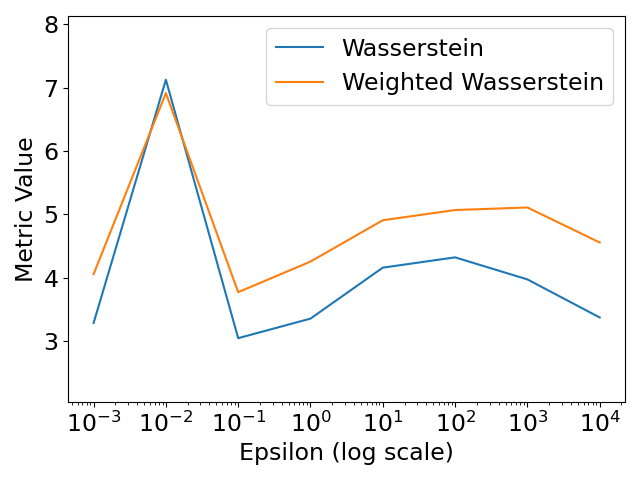}
    \caption*{(b) IVP Sampling}
\end{minipage}
\vspace{-0.05in}
\caption{Performance variation with $\epsilon$ for interpolation on the Brain Regeneration dataset.}
\label{fig:salamander_epsilon_interp_full}
\end{figure}


\section{Full Ablation Results}
\label{apdx:Ablations}

\subsection{Next Step Sampling for Axolotl Brain Regeneration}

\begin{table}[H]
\caption{Interpolation via Next Step Sampling at holdout time 3 for the Brain Regeneration dataset.}

\label{tab:ablation_interp_nextstep_stereoseq_table}
\centering
\small
\setlength{\tabcolsep}{5pt}
\resizebox{1.0\textwidth}{!}{
\begin{tabular}{l l cc | cccc}
\toprule
\textbf{Sampling} & \textbf{Method} & \(\bm{\lambda}\) & \(\bm{\alpha}\) & \textbf{Weighted} $\bm{\mathcal{W}_{2}}$ & $\bm{\mathcal{W}_{2}}$ & \textbf{MMD} & \textbf{Energy} \\
\midrule
\multirow{13}{*}{Next Step} & CFM & -- & -- & $2.618 \pm 0.142$ & $2.579 \pm 0.197$ & $0.043 \pm 0.003$ & $12.505 \pm 1.271$ \\
& MOTFM& -- & -- & $2.567 \pm 0.088$ & $2.476 \pm 0.161$ & $0.040 \pm 0.003$ & $11.269 \pm 1.388$ \\
\cmidrule{2-8}
& \multirow{9}{*}{CTF-C} & $1$ & $0.2$ & $2.503 \pm 0.071$ & $2.425 \pm 0.239$ & $0.037 \pm 0.003$ & $9.868 \pm 1.293$ \\
& & $1$ & $0.5$ & $2.467 \pm 0.107$ & $2.301 \pm 0.163$ & $0.037 \pm 0.002$ & $9.532 \pm 1.093$ \\
& & $1$ & $0.8$ & $2.423 \pm 0.164$ & $2.293 \pm 0.103$ & $0.037 \pm 0.001$ & $9.874 \pm 0.659$ \\
& & $0$ & $0.2$ & $2.396 \pm 0.028$ & $2.100 \pm 0.102$ & $0.033 \pm 0.003$ & $8.577 \pm 0.976$ \\
& & $0$ & $0.5$ & $2.447 \pm 0.142$ & $2.337 \pm 0.216$ & $0.036 \pm 0.005$ & $9.696 \pm 1.882$ \\
& & $0$ & $0.8$ & $2.413 \pm 0.099$ & $2.293 \pm 0.161$ & $0.036 \pm 0.002$ & $9.114 \pm 1.092$ \\
& & $0.5$ & $0.2$ & $2.460 \pm 0.118$ & $2.342 \pm 0.144$ & $0.036 \pm 0.003$ & $9.500 \pm 1.067$ \\
& & $0.5$ & $0.5$ & $2.504 \pm 0.094$ & $2.309 \pm 0.139$ & $0.036 \pm 0.003$ & $9.394 \pm 1.431$ \\
& & $0.5$ & $0.8$ & $2.442 \pm 0.173$ & $2.353 \pm 0.241$ & $0.035 \pm 0.004$ & $9.008 \pm 2.094$ \\
\cmidrule{2-8}
& \multirow{3}{*}{CTF-H} & $0$ & -- & $2.528 \pm 0.143$ & $2.534 \pm 0.180$ & $0.040 \pm 0.004$ & $11.192 \pm 1.304$ \\
& & $1$ & -- & $\bm{2.316 \pm 0.141}$ & $\bm{1.969 \pm 0.221}$ & $\bm{0.030 \pm 0.004}$ & $\bm{6.359 \pm 1.336}$ \\
& & $0.5$ & -- & $2.519 \pm 0.167$ & $2.412 \pm 0.158$ & $0.039 \pm 0.004$ & $10.304 \pm 1.808$ \\
\bottomrule
\end{tabular}
}
\end{table}

\begin{table}[H]
\caption{Extrapolation via Next Step Sampling at holdout time 5 for the Brain Regeneration dataset.}

\label{tab:ablation_extrap_nextstep_stereoseq_table}
\centering
\small
\setlength{\tabcolsep}{5pt}
\resizebox{1.0\textwidth}{!}{
\begin{tabular}{l l cc | cccc}
\toprule
\textbf{Sampling} & \textbf{Method} & \(\bm{\lambda}\) & \(\bm{\alpha}\) & \textbf{Weighted} $\bm{\mathcal{W}_{2}}$ & $\bm{\mathcal{W}_{2}}$ & \textbf{MMD} & \textbf{Energy} \\
\midrule
\multirow{13}{*}{Next Step} & CFM & -- & -- & $7.124 \pm 0.443$ & $7.133 \pm 0.533$ & $0.276 \pm 0.011$ & $76.947 \pm 5.661$ \\
& MOTFM& -- & -- & $7.487 \pm 0.698$ & $7.449 \pm 0.931$ & $0.266 \pm 0.010$ & $81.965 \pm 9.812$ \\
\cmidrule{2-8}
& \multirow{9}{*}{CTF-C} & $1$ & $0.2$ & $7.257 \pm 0.597$ & $7.077 \pm 0.473$ & $0.257 \pm 0.004$ & $79.562 \pm 7.787$ \\
& & $1$ & $0.5$ & $6.968 \pm 0.608$ & $6.969 \pm 0.628$ & $0.265 \pm 0.009$ & $77.025 \pm 6.056$ \\
& & $1$ & $0.8$ & $7.695 \pm 0.443$ & $7.792 \pm 0.463$ & $0.266 \pm 0.007$ & $87.179 \pm 6.690$ \\
& & $0$ & $0.2$ & $8.170 \pm 0.663$ & $8.079 \pm 0.723$ & $0.269 \pm 0.008$ & $91.572 \pm 8.802$ \\
& & $0$ & $0.5$ & $7.244 \pm 0.804$ & $7.146 \pm 0.775$ & $0.265 \pm 0.003$ & $80.424 \pm 10.376$ \\
& & $0$ & $0.8$ & $7.382 \pm 1.068$ & $7.234 \pm 0.852$ & $0.267 \pm 0.009$ & $81.635 \pm 14.135$ \\
& & $0.5$ & $0.2$ & $7.194 \pm 0.239$ & $7.171 \pm 0.422$ & $0.266 \pm 0.001$ & $78.924 \pm 3.715$ \\
& & $0.5$ & $0.5$ & $7.188 \pm 0.391$ & $\bm{6.931 \pm 0.260}$ & $0.267 \pm 0.005$ & $78.992 \pm 6.195$ \\
& & $0.5$ & $0.8$ & $7.242 \pm 0.804$ & $7.166 \pm 0.980$ & $0.267 \pm 0.006$ & $80.509 \pm 10.304$ \\
\cmidrule{2-8}
& \multirow{3}{*}{CTF-H} & $0$ & -- & $\bm{6.914 \pm 0.471}$ & $7.198 \pm 0.726$ & $0.266 \pm 0.009$ & $\bm{76.149 \pm 8.436}$ \\
& & $1$ & -- & $7.505 \pm 0.667$ & $7.338 \pm 0.601$ & $\bm{0.263 \pm 0.006}$ & $83.425 \pm 8.793$ \\
& & $0.5$ & -- & $7.243 \pm 0.479$ & $7.157 \pm 0.641$ & $0.270 \pm 0.007$ & $79.826 \pm 8.067$ \\
\bottomrule
\end{tabular}
}
\end{table}

\subsection{IVP Sampling on Axolotl Brain Regeneration}

\begin{table}[H]
\caption{Interpolation via IVP Sampling at time point 3 for the Brain Regeneration dataset.}

\label{tab:ablation_interp_ivp_stereoseq_table}
\centering
\small
\setlength{\tabcolsep}{5pt}
\resizebox{1.0\textwidth}{!}{
\begin{tabular}{l l cc | cccc}
\toprule
\textbf{Sampling} & \textbf{Method} & \(\bm{\lambda}\) & \(\bm{\alpha}\) & \textbf{Weighted} $\bm{\mathcal{W}_{2}}$ & $\bm{\mathcal{W}_{2}}$ & \textbf{MMD} & \textbf{Energy} \\
\midrule
\multirow{13}{*}{IVP} & CFM & -- & -- & $4.216 \pm 0.463$ & $4.266 \pm 0.308$ & $0.170 \pm 0.029$ & $32.413 \pm 5.122$ \\
& MOTFM& -- & -- & $4.198 \pm 0.319$ & $4.452 \pm 0.243$ & $0.173 \pm 0.017$ & $33.149 \pm 3.321$ \\
\cmidrule{2-8}
& \multirow{9}{*}{CTF-C} & $1$ & $0.2$ & $4.011 \pm 0.276$ & $4.048 \pm 0.321$ & $0.147 \pm 0.021$ & $30.337 \pm 4.713$ \\
& & $1$ & $0.5$ & $3.932 \pm 0.377$ & $4.356 \pm 0.398$ & $0.156 \pm 0.025$ & $31.524 \pm 4.875$ \\
& & $1$ & $0.8$ & $3.603 \pm 0.300$ & $3.816 \pm 0.310$ & $0.127 \pm 0.018$ & $24.271 \pm 3.992$ \\
& & $0$ & $0.2$ & $\bm{3.465 \pm 0.232}$ & $\bm{3.641 \pm 0.320}$ & $0.119 \pm 0.025$ & $23.055 \pm 5.939$ \\
& & $0$ & $0.5$ & $3.943 \pm 0.413$ & $4.241 \pm 0.435$ & $0.150 \pm 0.039$ & $29.221 \pm 5.713$ \\
& & $0$ & $0.8$ & $3.881 \pm 0.368$ & $4.094 \pm 0.551$ & $0.139 \pm 0.026$ & $27.941 \pm 6.676$ \\
& & $0.5$ & $0.2$ & $4.152 \pm 0.341$ & $4.322 \pm 0.291$ & $0.166 \pm 0.014$ & $33.299 \pm 3.629$ \\
& & $0.5$ & $0.5$ & $4.013 \pm 0.187$ & $4.138 \pm 0.297$ & $0.153 \pm 0.020$ & $30.941 \pm 3.685$ \\
& & $0.5$ & $0.8$ & $4.015 \pm 0.351$ & $3.974 \pm 0.442$ & $0.140 \pm 0.038$ & $27.592 \pm 6.669$ \\
\cmidrule{2-8}
& \multirow{3}{*}{CTF-H} & $0$ & -- & $3.925 \pm 0.267$ & $4.375 \pm 0.297$ & $0.164 \pm 0.013$ & $32.034 \pm 3.270$ \\
& & $1$ & -- & $3.905 \pm 0.395$ & $4.188 \pm 0.685$ & $\bm{0.074 \pm 0.014}$ & $\bm{18.728 \pm 2.689}$ \\
& & $0.5$ & -- & $3.917 \pm 0.343$ & $4.159 \pm 0.455$ & $0.147 \pm 0.022$ & $29.613 \pm 4.822$ \\
\bottomrule
\end{tabular}
}
\end{table}

\begin{table}[H]
\caption{Extrapolation via IVP Sampling at holdout time 5 for the Brain Regeneration dataset.}

\label{tab:ablation_extrap_ivp_stereoseq_table}
\centering
\small
\setlength{\tabcolsep}{5pt}
\resizebox{1.0\textwidth}{!}{
\begin{tabular}{l l cc | cccc}
\toprule
\textbf{Sampling} & \textbf{Method} & \(\bm{\lambda}\) & \(\bm{\alpha}\) & \textbf{Weighted} $\bm{\mathcal{W}_{2}}$ & $\bm{\mathcal{W}_{2}}$ & \textbf{MMD} & \textbf{Energy} \\
\midrule
\multirow{13}{*}{IVP} & CFM & -- & -- & $6.633 \pm 1.312$ & $7.116 \pm 1.084$ & $0.143 \pm 0.037$ & $60.573 \pm 21.756$ \\
& MOTFM& -- & -- & $6.503 \pm 0.720$ & $6.352 \pm 0.592$ & $0.162 \pm 0.038$ & $56.452 \pm 15.932$ \\
\cmidrule{2-8}
& \multirow{9}{*}{CTF-C} & $1$ & $0.2$ & $6.403 \pm 0.959$ & $6.558 \pm 1.297$ & $0.160 \pm 0.024$ & $61.051 \pm 16.594$ \\
& & $1$ & $0.5$ & $6.260 \pm 0.616$ & $7.681 \pm 4.003$ & $0.157 \pm 0.039$ & $52.478 \pm 12.010$ \\
& & $1$ & $0.8$ & $6.875 \pm 0.643$ & $6.920 \pm 0.796$ & $0.159 \pm 0.045$ & $62.838 \pm 16.897$ \\
& & $0$ & $0.2$ & $6.722 \pm 0.905$ & $6.782 \pm 1.003$ & $0.154 \pm 0.034$ & $53.996 \pm 15.617$ \\
& & $0$ & $0.5$ & $6.614 \pm 0.710$ & $6.854 \pm 0.740$ & $0.201 \pm 0.023$ & $70.370 \pm 9.099$ \\
& & $0$ & $0.8$ & $6.504 \pm 0.925$ & $6.744 \pm 1.336$ & $0.174 \pm 0.037$ & $56.687 \pm 18.118$ \\
& & $0.5$ & $0.2$ & $6.514 \pm 0.504$ & $5.998 \pm 0.803$ & $0.155 \pm 0.032$ & $51.329 \pm 15.080$ \\
& & $0.5$ & $0.5$ & $6.696 \pm 0.427$ & $6.481 \pm 0.387$ & $0.195 \pm 0.024$ & $66.212 \pm 3.542$ \\
& & $0.5$ & $0.8$ & $6.550 \pm 0.975$ & $6.563 \pm 1.029$ & $0.188 \pm 0.037$ & $63.014 \pm 14.173$ \\
\cmidrule{2-8}
& \multirow{3}{*}{CTF-H} & $0$ & -- & $6.243 \pm 0.760$ & $6.220 \pm 0.751$ & $0.195 \pm 0.020$ & $61.316 \pm 10.288$ \\
& & $1$ & -- & $\bm{5.277 \pm 0.936}$ & $6.021 \pm 1.192$ & $\bm{0.099 \pm 0.007}$ & $\bm{27.777 \pm 8.621}$ \\
& & $0.5$ & -- & $6.254 \pm 0.819$ & $\bm{5.973 \pm 0.757}$ & $0.156 \pm 0.025$ & $54.330 \pm 12.089$ \\
\bottomrule
\end{tabular}
}
\end{table}

\subsection{Next Step Sampling for Mouse Embryo Organogenesis}

\begin{table}[H]
\caption{Interpolation via Next Step Sampling at holdout time 5 for the Organogenesis dataset.}

\label{tab:ablation_interp_nextstep_mosta_table}
\centering
\small
\setlength{\tabcolsep}{5pt}
\resizebox{1.0\textwidth}{!}{
\begin{tabular}{l l cc | cccc}
\toprule
\textbf{Sampling} & \textbf{Method} & \(\bm{\lambda}\) & \(\bm{\alpha}\) & \textbf{Weighted} $\bm{\mathcal{W}_{2}}$ & $\bm{\mathcal{W}_{2}}$ & \textbf{MMD} & \textbf{Energy} \\
\midrule
\multirow{11}{*}{Next Step} & MOTFM& -- & -- & $1.892 \pm 0.028$ & $1.873 \pm 0.086$ & $0.164 \pm 0.002$ & $11.615 \pm 0.092$ \\
\cmidrule{2-8}
& \multirow{9}{*}{CTF-C} & $1$ & $0.2$ & $1.881 \pm 0.020$ & $1.922 \pm 0.078$ & $0.158 \pm 0.003$ & $11.529 \pm 0.197$ \\
& & $1$ & $0.5$ & $\bm{1.865 \pm 0.030}$ & $1.852 \pm 0.093$ & $0.159 \pm 0.001$ & $11.482 \pm 0.108$ \\
& & $1$ & $0.8$ & $1.889 \pm 0.024$ & $1.888 \pm 0.082$ & $0.161 \pm 0.002$ & $11.552 \pm 0.166$ \\
& & $0$ & $0.2$ & $1.893 \pm 0.035$ & $1.912 \pm 0.057$ & $0.159 \pm 0.001$ & $11.462 \pm 0.154$ \\
& & $0$ & $0.5$ & $1.877 \pm 0.039$ & $1.933 \pm 0.088$ & $0.162 \pm 0.002$ & $11.528 \pm 0.110$ \\
& & $0$ & $0.8$ & $1.882 \pm 0.022$ & $1.869 \pm 0.049$ & $0.161 \pm 0.001$ & $\bm{11.399 \pm 0.119}$ \\
& & $0.5$ & $0.2$ & $1.886 \pm 0.022$ & $1.927 \pm 0.111$ & $\bm{0.157 \pm 0.002}$ & $11.430 \pm 0.131$ \\
& & $0.5$ & $0.5$ & $1.899 \pm 0.027$ & $1.899 \pm 0.072$ & $0.160 \pm 0.002$ & $11.517 \pm 0.097$ \\
& & $0.5$ & $0.8$ & $1.888 \pm 0.033$ & $\bm{1.839 \pm 0.134}$ & $0.161 \pm 0.002$ & $11.475 \pm 0.159$ \\
\cmidrule{2-8}
& \multirow{3}{*}{CTF-H} & $0$ & -- & $1.884 \pm 0.027$ & $1.862 \pm 0.123$ & $0.164 \pm 0.001$ & $11.499 \pm 0.123$ \\
& & $1$ & -- & $1.898 \pm 0.029$ & $1.866 \pm 0.097$ & $0.167 \pm 0.002$ & $11.795 \pm 0.170$ \\
& & $0.5$ & -- & $1.871 \pm 0.030$ & $1.919 \pm 0.067$ & $0.164 \pm 0.002$ & $11.639 \pm 0.182$ \\
\bottomrule
\end{tabular}
}
\end{table}

\begin{table}[H]
\caption{Extrapolation via Next Step Sampling at holdout time 8 for the Organogenesis dataset.}

\label{tab:ablation_extrap_nextstep_mosta_table}
\centering
\small
\setlength{\tabcolsep}{5pt}
\resizebox{1.0\textwidth}{!}{
\begin{tabular}{l l cc | cccc}
\toprule
\textbf{Sampling} & \textbf{Method} & \(\bm{\lambda}\) & \(\bm{\alpha}\) & \textbf{Weighted} $\bm{\mathcal{W}_{2}}$ & $\bm{\mathcal{W}_{2}}$ & \textbf{MMD} & \textbf{Energy} \\
\midrule
\multirow{11}{*}{Next Step} & MOTFM& -- & -- & $1.626 \pm 0.066$ & $1.682 \pm 0.096$ & $0.084 \pm 0.007$ & $7.418 \pm 0.749$ \\
\cmidrule{2-8}
& \multirow{9}{*}{CTF-C} & $1$ & $0.2$ & $1.683 \pm 0.058$ & $1.803 \pm 0.117$ & $0.087 \pm 0.006$ & $7.830 \pm 0.551$ \\
& & $1$ & $0.5$ & $1.685 \pm 0.096$ & $1.714 \pm 0.159$ & $0.089 \pm 0.006$ & $8.056 \pm 1.033$ \\
& & $1$ & $0.8$ & $1.703 \pm 0.063$ & $1.830 \pm 0.131$ & $0.095 \pm 0.005$ & $8.928 \pm 0.723$ \\
& & $0$ & $0.2$ & $1.715 \pm 0.123$ & $1.860 \pm 0.267$ & $0.094 \pm 0.009$ & $9.021 \pm 1.740$ \\
& & $0$ & $0.5$ & $1.725 \pm 0.082$ & $1.856 \pm 0.191$ & $0.093 \pm 0.006$ & $8.806 \pm 0.749$ \\
& & $0$ & $0.8$ & $1.774 \pm 0.053$ & $1.897 \pm 0.175$ & $0.094 \pm 0.007$ & $9.466 \pm 0.957$ \\
& & $0.5$ & $0.2$ & $1.818 \pm 0.096$ & $2.089 \pm 0.222$ & $0.084 \pm 0.008$ & $8.875 \pm 0.976$ \\
& & $0.5$ & $0.5$ & $1.774 \pm 0.104$ & $1.899 \pm 0.280$ & $0.093 \pm 0.007$ & $9.139 \pm 1.437$ \\
& & $0.5$ & $0.8$ & $1.768 \pm 0.058$ & $1.858 \pm 0.120$ & $0.101 \pm 0.006$ & $9.303 \pm 0.634$ \\
\cmidrule{2-8}
& \multirow{3}{*}{CTF-H} & $0$ & -- & $\bm{1.505 \pm 0.057}$ & $\bm{1.397 \pm 0.088}$ & $0.087 \pm 0.005$ & $\bm{5.954 \pm 0.492}$ \\
& & $1$ & -- & $1.890 \pm 0.046$ & $1.877 \pm 0.103$ & $0.147 \pm 0.006$ & $10.752 \pm 0.405$ \\
& & $0.5$ & -- & $1.636 \pm 0.060$ & $1.684 \pm 0.099$ & $\bm{0.081 \pm 0.005}$ & $7.088 \pm 0.692$ \\
\bottomrule
\end{tabular}
}
\end{table}

\subsection{IVP Sampling for Mouse Embryo Organogenesis}

Extrapolating to the last holdout time point of the mouse organogenesis dataset~\citep{2_dataset_mosta}, particularly under IVP-Sampling, represents the most challenging setting among all our experiments. This difficulty arises because the target time point lies entirely outside the training horizon, requiring integration from the initial samples through to the end. As a result, the velocity field has more opportunity to drift in incorrect directions, often leading to generations that deviate substantially from the true dynamics.
In our experiments, this instability was evident: across 10 runs, several produced highly unstable trajectories, reflecting the sensitivity of the system to initial conditions and numerical solvers. This variability is also captured in the performance metrics reported in Table~\ref{tab:ablation_extrap_ivp_mosta_table}.

\begin{table}[H]
\caption{Interpolation via IVP Sampling at holdout time 5 for the Mouse Organogenesis dataset.}

\label{tab:ablation_interp_ivp_mosta_table}
\centering
\small
\setlength{\tabcolsep}{5pt}
\resizebox{1.0\textwidth}{!}{
\begin{tabular}{l l cc | cccc}
\toprule
\textbf{Sampling} & \textbf{Method} & \(\bm{\lambda}\) & \(\bm{\alpha}\) & \textbf{Weighted} $\bm{\mathcal{W}_{2}}$ & $\bm{\mathcal{W}_{2}}$ & \textbf{MMD} & \textbf{Energy} \\
\midrule
\multirow{13}{*}{IVP} & MOTFM& -- & -- & $3.251 \pm 0.676$ & $3.418 \pm 0.727$ & $0.090 \pm 0.003$ & $9.226 \pm 0.648$ \\
\cmidrule{2-8}
& \multirow{9}{*}{CTF-C} & $1$ & $0.2$ & $3.261 \pm 0.880$ & $5.264 \pm 3.060$ & $0.089 \pm 0.003$ & $10.724 \pm 1.288$ \\
& & $1$ & $0.5$ & $3.137 \pm 0.407$ & $4.093 \pm 1.187$ & $0.086 \pm 0.004$ & $11.948 \pm 1.393$ \\
& & $1$ & $0.8$ & $3.392 \pm 0.757$ & $4.716 \pm 2.079$ & $0.089 \pm 0.005$ & $9.547 \pm 0.752$ \\
& & $0$ & $0.2$ & $2.953 \pm 0.425$ & $3.816 \pm 0.973$ & $0.083 \pm 0.002$ & $9.816 \pm 0.715$ \\
& & $0$ & $0.5$ & $2.938 \pm 0.476$ & $3.904 \pm 1.120$ & $0.088 \pm 0.005$ & $9.864 \pm 0.764$ \\
& & $0$ & $0.8$ & $3.101 \pm 0.539$ & $3.855 \pm 0.946$ & $0.087 \pm 0.004$ & $9.280 \pm 0.551$ \\
& & $0.5$ & $0.2$ & $3.771 \pm 0.862$ & $5.457 \pm 1.704$ & $\bm{0.079 \pm 0.004}$ & $9.262 \pm 1.134$ \\
& & $0.5$ & $0.5$ & $3.090 \pm 0.635$ & $4.596 \pm 2.357$ & $0.084 \pm 0.005$ & $9.786 \pm 1.067$ \\
& & $0.5$ & $0.8$ & $3.200 \pm 0.403$ & $3.555 \pm 0.637$ & $0.084 \pm 0.004$ & $9.269 \pm 0.541$ \\
\cmidrule{2-8}
& \multirow{3}{*}{CTF-H} & $0$ & -- & $3.244 \pm 0.713$ & $3.946 \pm 1.671$ & $0.089 \pm 0.005$ & $\bm{8.797 \pm 0.612}$ \\
& & $1$ & -- & $5.200 \pm 0.799$ & $6.306 \pm 1.037$ & $0.123 \pm 0.008$ & $45.862 \pm 13.765$ \\
& & $0.5$ & -- & $\bm{2.814 \pm 0.414}$ & $\bm{3.233 \pm 0.567}$ & $0.093 \pm 0.005$ & $10.319 \pm 0.817$ \\
\bottomrule
\end{tabular}
}
\end{table}

\begin{table}[H]
\caption{Extrapolation via IVP Sampling at holdout time 8 for the Mouse Organogenesis dataset.}

\label{tab:ablation_extrap_ivp_mosta_table}
\centering
\small
\setlength{\tabcolsep}{5pt}
\resizebox{1.0\textwidth}{!}{
\begin{tabular}{l l cc | cccc}
\toprule
\textbf{Sampling} & \textbf{Method} & $\bm{\lambda}$ & $\bm{\alpha}$ & \textbf{Weighted} $\bm{\mathcal{W}_{2}}$ & $\bm{\mathcal{W}_{2}}$ & \textbf{MMD} & \textbf{Energy} \\
\midrule
\multirow{13}{*}{IVP} 
& MOTFM & -- & -- & $8.058 \pm 2.415$ & $27.283 \pm 4.350$ & $0.090 \pm 0.003$ & $28.890 \pm 6.948$ \\
\cmidrule{2-8}

& \multirow{9}{*}{CTF-C} 
& $1$ & $0.2$ & $9.988 \pm 2.116$ & $45.117 \pm 15.750$ & $0.084 \pm 0.003$ & $50.952 \pm 12.577$ \\
& & $1$ & $0.5$ & $7.929 \pm 1.464$ & $35.572 \pm 13.551$ & $0.086 \pm 0.002$ & $42.040 \pm 18.811$ \\
& & $0$ & $0.8$ & $7.729 \pm 2.846$ & $26.006 \pm 5.753$ & $0.086 \pm 0.004$ & $24.431 \pm 3.700$ \\

& & $1$ & $0.2$ & $10.324 \pm 1.979$ & $51.492 \pm 16.681$ & $0.084 \pm 0.002$ & $45.002 \pm 24.327$ \\
& & $0$ & $0.5$ & $7.799 \pm 1.136$ & $32.535 \pm 9.496$ & $0.085 \pm 0.002$ & $51.080 \pm 13.594$ \\
& & $0$ & $0.8$ & $9.317 \pm 5.319$ & $32.195 \pm 10.782$ & $0.087 \pm 0.004$ & $28.873 \pm 6.062$ \\

& & $0.5$ & $0.2$ & $10.762 \pm 1.998$ & $54.796 \pm 13.489$ & $0.089 \pm 0.002$ & $60.875 \pm 28.610$ \\
& & $0.5$ & $0.5$ & $7.755 \pm 1.519$ & $31.069 \pm 11.415$ & $0.086 \pm 0.002$ & $39.806 \pm 13.866$ \\
& & $0.5$ & $0.8$ & $8.326 \pm 2.558$ & $24.306 \pm 6.339$ & $0.086 \pm 0.002$ & $29.195 \pm 8.088$ \\
\cmidrule{2-8}

& \multirow{3}{*}{CTF-H} 
& $0$ & -- & $6.986 \pm 1.295$ & $23.806 \pm 7.516$ & $0.093 \pm 0.004$ & $\bm{22.466 \pm 5.099}$ \\
& & $1$ & -- & $\bm{6.378 \pm 1.813}$ & $\bm{11.844 \pm 6.567}$ & $0.103 \pm 0.008$ & $30.752 \pm 9.439$ \\
& & $0.5$ & -- & $7.749 \pm 1.376$ & $32.480 \pm 8.322$ & $\bm{0.084 \pm 0.002}$ & $26.129 \pm 7.315$ \\

\bottomrule
\end{tabular}
}
\end{table}


\subsection{Liver Regeneration}

\begin{table}[H]
\caption{$2$-Wasserstein distances for different methods and configurations.}

\label{tab:wasserstein_distances}
\centering
\setlength{\tabcolsep}{5pt}
\begin{tabular}{l c c | c}
\toprule
\textbf{Variant} & $\bm{\lambda}$ & $\bm{\alpha}$ & $\bm{\mathcal{W}_{2}}$ \\
\midrule
EOT   & --   & --   & $34.303 \pm 1.447$ \\
\midrule
CTF-C & $1$  & $0.2$ & $34.444 \pm 1.193$ \\
CTF-C & $1$  & $0.5$ & $33.956 \pm 1.644$ \\
CTF-C & $1$  & $0.8$ & $34.628 \pm 0.981$ \\
CTF-C & $0$  & $0.2$ & $34.241 \pm 1.169$ \\
CTF-C & $0$  & $0.5$ & $32.741 \pm 1.863$ \\
CTF-C & $0$  & $0.8$ & $33.717 \pm 1.230$ \\
CTF-C & $0.5$ & $0.2$ & $33.566 \pm 1.043$ \\
CTF-C & $0.5$ & $0.5$ & $33.841 \pm 1.714$ \\
CTF-C & $0.5$ & $0.8$ & $33.045 \pm 1.643$ \\
\midrule
CTF-H & $0$  & --   & $\bm{32.682 \pm 1.471}$ \\
CTF-H & $1$  & --   & $33.480 \pm 1.001$ \\
CTF-H & $0.5$ & --   & $33.414 \pm 0.995$ \\
\bottomrule
\end{tabular}
\end{table}

\end{document}